%% file: acl_latex.tex
\pdfoutput=1

\documentclass[11pt]{article}

\usepackage[]{acl}
\usepackage{enumitem}
\usepackage{amssymb}
\usepackage{dashbox}
\usepackage{times}
\usepackage[font=small,labelfont=bf]{caption}
\usepackage{latexsym}
\usepackage{color,graphicx}
\usepackage{subcaption}
\usepackage{trimclip}

\usepackage{tikz}
\usepackage{tikz-dependency}
\usetikzlibrary{%
  shapes,%
  arrows,%
  positioning,%
  calc,%
  automata%
}
\usepackage{pgfplots}
\usepgfplotslibrary{fillbetween}
\usepackage{url}
\definecolor{pf7}{RGB}{166, 118, 29}

\usepackage{nccmath}

\newcommand{\slen}{\ell}
\newcommand\bigfrac[2]{%
  \begin{array}{l}
    #1 \vspace{1mm} \\
    #2
  \end{array}}

\usepackage[T1]{fontenc}
\usepackage[utf8x]{inputenc}

\usepackage{microtype}

\usepackage{graphicx}

\usepackage{amsmath}
\usepackage{amsfonts}
\usepackage{amssymb}

\usepackage{multirow, multicol}
\usepackage{pgfplots}
\usepackage{inconsolata}
\usepackage{xcolor}        
\usepackage{enumitem}
\usepackage{booktabs}       
\usepackage{graphicx}
\usepackage{footmisc}
\usepackage{mwe}
\usepackage{amssymb}
\usepackage{amsfonts}
\usepackage{amsmath}
\usepackage{mathtools}
\usepackage{tikz}
\usepackage{pgfplots}
\usepackage{tikz-qtree}
\usepackage{tikz-dependency}
\usepackage{forest}
\usepackage{multirow}
\usepackage{subcaption}
\usepackage{booktabs}
\usepackage{proof}
\usepackage{hyperref}
\usepackage{csquotes}
\usepackage[linesnumbered,ruled,lined]{algorithm2e}
\usepackage{bm}
\usepackage{arydshln}
\useforestlibrary{linguistics}
\def\blfootnote{\xdef\@thefnmark{}\@footnotetext}
\usepackage{tikz}

\newcommand*\yellowcircled[1]{\tikz[baseline=(char.base)]{%
            \node[shape=circle,fill=yellow!20,draw,inner sep=2pt, minimum size=17.5pt] (char) {#1};}}
\newcommand*\redcircled[1]{\tikz[baseline=(char.base)]{%
            \node[shape=circle,fill=red!20,draw,inner sep=0pt, minimum size=17.5pt] (char) {#1};}}

\usetikzlibrary{bayesnet}

\title{Unsupervised Discontinuous Constituency Parsing with \\ Mildly Context-Sensitive Grammars}

\author{Songlin Yang \\
  ShanghaiTech University   \\
  {\texttt{yangsl66@mit.edu}} \\\And
  Roger P. Levy \\
  MIT BCS \\
   { \texttt{rplevy@mit.edu}}  \\\And 
  Yoon Kim \\
  MIT CSAIL \\ 
  { \texttt{yoonkim@mit.edu}} \\}

\begin{document}
\setlength{\abovedisplayskip}{4pt}
\setlength{\belowdisplayskip}{4pt}
\setlength{\abovedisplayshortskip}{4pt}
\setlength{\belowdisplayshortskip}{4pt}

\maketitle
\begin{abstract}

We study grammar induction with mildly context-sensitive grammars for unsupervised discontinuous parsing. Using the probabilistic linear context-free rewriting system (LCFRS) formalism, our approach fixes the rule structure in advance and focuses on parameter learning with maximum likelihood.  To reduce the computational complexity of both parsing and parameter estimation, we restrict the grammar formalism to binary LCFRS with fan-out two and further discard  rules that require $\mathcal{O}(\slen^6)$ time to parse, reducing inference to  $\mathcal{O}(\slen^5)$. We find that using a large number of nonterminals is beneficial and thus make use of tensor decomposition-based rank-space dynamic programming with an embedding-based parameterization of rule probabilities to scale up the number of nonterminals. Experiments on German and Dutch show that our approach is able to induce  linguistically meaningful trees with continuous and discontinuous structures.
\blfootnote{\noindent \hspace{-6mm} Code: \url{https://github.com/sustcsonglin/TN-LCFRS}.} 

\end{abstract}

\section{Introduction} 
Unsupervised parsing aims to induce hierarchical linguistic structures given only the strings in a language. A classic approach to unsupervised parsing is through probabilistic grammar induction \citep{lari1990estimation}, which learns a probabilistic grammar (i.e., a set of rewrite rules and their probabilities) from raw text.   
Recent work has shown that neural parameterizations of probabilistic context-free grammars (PCFG), wherein the grammar's rule probabilities are given by a neural network over shared symbol embeddings, can achieve promising results on unsupervised constituency parsing~\cite{kim-etal-2019-compound, jin-etal-2019-unsupervised, jin-etal-2021-character-based,  yang-etal-2021-pcfgs,yang-etal-2022-dynamic}. 

However, context-free rules  are not natural for modeling \emph{discontinuous} language phenomena such as extrapositions, cross-serial dependencies, and wh-movements.
\emph{Mildly context-sensitive grammars} \citep{joshi1985much}, which sit between context-free  and context-sensitive grammars in the classic Chomsky–Sch{\"u}tzenberger hierarchy \citep{Chomsky59a,chomsky63},\footnote{This hierarchy does not necessarily extend to probabilistic grammars. For example \citet{Icard2020-ICACGM} show that in a particular probabilistic version of the hierarchy in which a probabilistic grammar over a one-letter alphabet induces a distribution over the integers via its unary representation, the set of distributions that can be expressed by probabilistic mildly context-sensitive grammars (such as linear indexed grammars) is not a proper subset of the set of distributions that can be expressed by probabilistic context-sensitive grammars.}  are powerful enough to model richer aspects of natural language including  discontinuous and non-local phenomena. And despite their expressivity they  enjoy polynomial-time inference algorithms, making them attractive both as cognitively plausible models of human language processing and as targets for unsupervised learning. 
\input{figures/tree.tex}

There are several weakly equivalent  formalisms for generating the mildly context-sensitive languages which might serve as potential targets for grammar induction: tree adjoining grammars \citep{JOSHI1975136}, head grammars \citep{pollard85head}, combinatory categorial grammars~\citep{steedman1987ccg}, and linear indexed grammars \citep{Gazdar1988ApplicabilityOI}. In this paper we work with linear context-free rewriting systems~\citep[LCFGS,][]{vijay-shanker-etal-1987-characterizing}, which generalize the above formalisms and are weakly equivalent to multiple context-free grammars~\citep{Seki1991OnMC}. Derivation trees in an LCFRS directly correspond to discontinuous constituency trees where each node can dominate a non-contiguous sequence of words in the yield, as shown in Fig.~\ref{fig:dis-tree}. 

We focus on the LCFRS formalism as it has previously been successfully  employed for supervised  discontinuous constituency parsing~\cite{levy2005prob, maier-2010-direct,vanCranenburgh2016DataOrientedPW}. The complexity of parsing in a LCFRS is $\mathcal{O}(\slen^{3k}|G|)$, where $\slen$ is the sentence length, $k$ is the fan-out (the maximum number of contiguous blocks of text that can be dominated by a nonterminal), and $|G|$ is the grammar size. While polynomial, this is too high to be practical for unsupervised learning on real-world data. We thus restrict ourselves to LCFRS-2, i.e., binary LCFRS with fan-out two, which has been shown to have high coverage on discontinuous treebanks~\cite{maier-etal-2012-plcfrs}.
Even with this restriction LCFRS-2 remains difficult to induce from raw text due to the $\mathcal{O}(\slen^6|G|)$ dynamic program for parsing and marginalization. However \citet{corro-2020-span} observe that a $\mathcal{O}(\slen^5|G|)$ variant of the grammar that discards certain rules can still recover 98\% of real world treebank constituents. Our  approach  uses with this restricted variant of LCFRS-2 (see Sec~\ref{sec:lcfrs2}). 
Finally, following recent work which finds that that overparameterizing deep latent variable models is beneficial for unsupervised learning~\cite{DBLP:conf/icml/BuhaiHKRS20, yang-etal-2021-pcfgs,chiu-rush-2020-scaling,DBLP:conf/nips/ChiuDR21}, we scale LCFRS-2 to a large number of nonterminals by adapting tensor-decomposition-based  inference techniques---originally developed for PCFGs ~\cite{cohen-etal-2013-approximate,yang-etal-2021-pcfgs, yang-etal-2022-dynamic}---to the LCFRS case.

We conduct experiments German and Dutch---both of which have frequent discontinuous and non-local language phenomena and have available discontinuous treebanks---and observe that our approach is able to induce grammars with nontrivial performance on discontinuous constituents.

\section{Approach}

\subsection{Background: Scaling PCFGs with low-rank neural parameterizations}
\label{sec:pcfg}
Inference in PCFGs is cubic with respect to the number of nonterminals in the general case, which can make it difficult to scale up PCFGs to a large number (e.g., thousands) of nonterminals. However, under certain parameterizations it is possible to exploit low rank factorizations of the rule probability tensor to enable faster inference. For example, given a PCFG with $m$ nonterminals \citet{cohen-etal-2013-approximate} use  canonical-polyadic decomposition \citep[CPD,][]{ DBLP:journals/corr/abs-1711-10781} to decompose the 3D binary rule probability tensor $\mathsf{T} \in \mathbb{R}^{m \times m \times m}$ as,
\begin{align*}
\mathsf{T} = \sum_{q=1}^{r} u_{q} \otimes v_{q} \otimes w_{q},
\end{align*}
where $u_{q}, v_{q}, w_{q} \in \mathbb{R}^{m}$, $r$ is the tensor rank (a hyperparameter), and $\otimes$ is the outer product. Letting $U, V, W \in \mathbb{R}^{r \times m}$ be the matrices resulting from stacking all  $u_{q}, v_{q}, w_{q}$, \citet{cohen-etal-2013-approximate} give the following recursive formula for calculating the inside tensor $\alpha \in \mathbb{R}^{(\slen+1)\times (\slen+1) \times m}$ for a sentence of length $\slen$:
\begin{align*}
\alpha^{L}_{i,j} = V\alpha_{i, k}, \quad& \alpha^{R}_{j,k}=  W\alpha_{k, j},\\
\alpha_{i, j}  
= U^{T} \sum_{k = i+1}^{j - 1} & \alpha^{L}_{i,j}  \circ \alpha^{R}_{j,k}. \end{align*}
 Here $\alpha^{L}, \alpha^{R} \in \mathbb{R}^{(\slen+1)\times (\slen+1) \times r}$ are  auxiliary tensors for storing intermediate values, and $\circ$ is the Hadamard product. The resulting complexity of this version of the inside algorithm is $\mathcal{O}(\slen^3r + \slen^2mr)$, which removes the cubic dependence on $m$.  Based on this formula,  \citet{yang-etal-2021-pcfgs} propose a low-rank \textit{neural} parameterization which uses a neural network over shared symbol embeddings to produce unnormalized score matrices $\bar{U}, \bar{V}, \bar{W}$. Then, $\bar{U}$ is $\operatorname{softmax}$-ed across columns to obtain $U$, while $\bar{V}, \bar{W}$ are $\operatorname{softmax}$-ed  across rows to obtain $V, W$.
The difference between  \citet{cohen-etal-2013-approximate} and \citet{yang-etal-2021-pcfgs} is that the former  performs CPD on an existing probability tensor $\mathsf{T}$ for faster (supervised) parsing, whereas the latter directly parameterizes and learns $U, V, W$ from data without actually instantiating
 $\mathsf{T}$.
 
 \citet{yang-etal-2022-dynamic} build on \citet{yang-etal-2021-pcfgs} 
 and further pre-compute matrices $J= VU^{T}, K=WU^{T}$  to rewrite the above recursive formula as:
\begin{align*}
    \alpha^{L}_{i,j} = J \alpha^{\prime}_{i, j} \quad,& \alpha^{R}_{i, j} = K \alpha^{\prime}_{i,j}   \\
    \alpha^{\prime}_{i, j} = \sum_{k=i+1}^{j-1}& \alpha^{L}_{i, j} \circ \alpha^{R}_{j, k} \
\end{align*}
where $\alpha^{\prime} \in \mathbb{R}^{(n+1)\times (n+1)\times r}$ is an auxiliary inside score tensor.
The resulting complexity of this approach is  $\mathcal{O}(\slen^3r + \slen^2r^2)$, which is smaller than $\mathcal{O}(\slen^3r+\slen^2mr)$ when $r  \ll m$, i.e., in the setting with a large number of nonterminals whose probability tensor is of low rank.  In this paper we adapt this low rank neural parameterization  to the LCFRS case to scale to a large number of nonterminals.

\subsection{Restricted LCFRS}
\label{sec:lcfrs2}
In an LCFRS, a single nonterminal node can dominate a tuple of strings that need not be adjacent in the yield. The tuple size is referred to as the \emph{fan-out}. We mark the fan-out of each non-leaf node in Fig.~\ref{fig:dis-tree}. The fan-out of an LCFRS is defined as the maximal fan-out among all its nonterminals, and influences expressiveness and parsing complexity. For a binary LCFRS (i.e., LCFRS with derivation rules that have at most two nonterminals on the right hand side) with fan-out $k$, the parsing complexity for a sentence of length $\slen$ is $\mathcal{O}(\slen^{3k} )$.\footnote{A binary CFG is thus a special case of a binary LCFRS with fan-out one, and parsing in this case reduces to the classic CKY algorithm.}
In this paper we work with binary  LCFRS with fan-out 2~\cite[LCFRS-2]{stanojevic-steedman-2020-span}, which is expressive enough to model discontinuous constituents but still efficient enough to enable practical grammar induction from natural language data. This choice is also motivated by \citet{maier-etal-2012-plcfrs} who observe that restricting the fan-out to two suffices for capturing a large proportion of  discontinuous constituents in standard treebanks.\footnote{For instance, \citet{stanojevic-steedman-2020-span} report that LCFRS-2 can cover up to 87\% of the gold discontinuous constituents in the NEGRA treebank. We refer readers to Table 1 of \citet{corro-2020-span} for more details.}

However, LCFRS-2's inference complexity of $\mathcal{O}(\slen^6|G|)$ is still too expensive for practical unsupervised learning. 
We thus follow~\citet{corro-2020-span} and discard all rules that  require $\mathcal{O}(\slen^6)$ time to parse, which reduces parsing complexity to $\mathcal{O}(\slen^5|G|$).\footnote{These correspond to rules (d), (i), (j), (k), and (l) in Figure 3 of ~\citet{corro-2020-span}.}  Formally, this restricted LCFRS-2 is a 6-tuple $\mathcal{G} = ({S},\mathcal{N}^1, \mathcal{N}^2,  \mathcal{P},\Sigma, \mathcal{R})$ where: ${S}$ is the start symbol; $\mathcal{N}^1, \mathcal{N}^2 $ are a finite set of nonterminal symbols of fan-out one and two, respectively; $\mathcal{P}$ is a finite set of preterminal symbols;
$\Sigma$ is a finite set of terminal symbols; and $\mathcal{R}$ is a set of rules of the following form (where $\mathcal{M} \triangleq \mathcal{N}^1 \cup \mathcal{P}$):

{\small  
\begin{align*}
&S(x) \rightarrow  A(x)  &A \in \mathcal{N}^1 \\
&A(xy) \rightarrow   B(x) C(y)&A \in \mathcal{N}^1, B, C \in \mathcal{M} \\
&A(yxz) \rightarrow  B(x) C(y, z)& A\in\mathcal{N}^1,B\in\mathcal{M}, C \in  \mathcal{N}^2 \\
&A(x,y) \rightarrow   B(x) C(y) & A \in \mathcal{N}^2, B, C \in \mathcal{M} \\
\end{align*}
}%
{\small  
\begin{align*}
&A(xy, z) \rightarrow   B(x) C(y,z) & A,C \in \mathcal{N}^2, B\in \mathcal{M} \\
&A(yx, z) \rightarrow   B(x) C(y,z) & A,C \in \mathcal{N}^2, B\in \mathcal{M} \\
&A(y, xz) \rightarrow   B(x) C(y,z) & A,C \in \mathcal{N}^2, B\in \mathcal{M} \\
&A(y, zx) \rightarrow   B(x) C(y,z) & A,C \in \mathcal{N}^2, B\in \mathcal{M} \\
&T(w)\rightarrow   w, &  T \in \mathcal{P}, w \in \Sigma.
\end{align*}
}%
Here $A(x)$ indicates that $A$ has a fan-out 1; $A(x,y)$ indicates that $A$ has a fan-out 2 and $x$ and $y$ are non-adjacent contiguous strings in the yield of $A$.
Each nonterminal is annotated with lower-case letters that stand for strings,  and $xy$ denotes the concatenation of $x$ and $y$, which are adjacent, into a single string $s\triangleq xy$.
  \input{tables/deduction.tex}
\vspace{-2mm}
 \paragraph{Illustrative Example.} 
 \input{figures/rule.tex}
 \vspace{-2mm}
As an example of how this LCFRS can model discontinuous spans, we depict the rule $A(xy, z) \rightarrow B(x) C(y,z)$ above. $B$ is a fan-out-1 node whose yield is $x= w_{i}\cdots w_{k-1}$ and $C$ is a fan-out-2 node whose first span is $y= w_{k} \cdots w_{j-1}$ and whose second span is $z= w_{m}\cdots w_{n-1}$. $A$ is the parent node of $B, C$, and inherits the yields of $B$ and $C$, where $x$ is concatenated with $y$ to form a contiguous span and $z$ is a standalone span.

 \paragraph{Parsing.}  Table~\ref{tab:deduction} gives the parsing-as-deduction~\cite{pereira-warren-1983-parsing} description of the CKY-style chart parsing algorithm of our restricted LCFRS-2.

\subsection{Tensor decomposition-based neural parameterization}
\label{sec:model}
We now describe a parameterization of LCFRS-2 that combines a neural parameterization with tensor decomposition, which makes it possible to scale LCFRS-2 to thousands of nonterminals. Let $m_1 = |\mathcal{N}^{1}|, m_2= |\mathcal{N}^{2}|, p= |\mathcal{P}|$, and $m= m_1+p $. The rules involving $A \in \mathcal{N}^{1}$ on the left hand side are $\yellowcircled{1a}$ and $\redcircled{2a}$, whose probabilities can be represented by 3D tensors $C^1 \in \mathbb{R}^{m_1 \times m \times m}$ and $D^1 \in \mathbb{R}^{m_1 \times m \times m_2}$.
For $A \in \mathcal{N}^{2}$, the relevant rules are $\yellowcircled{1b}, \redcircled{2b}, \redcircled{2c}, \redcircled{2d}, \redcircled{2e}$, whose probabilities can be represented by 3D tensors $C^2 \in \mathbb{R}^{m_2 \times m \times m}$ and $D^3,D^4,D^5,D^6 \in \mathbb{R}^{m_2 \times m \times m_2}$.
We stack $D^3, D^4, D^5, D^6$ into a single 4D tensor $D^2 \in \mathbb{R}^{m_2 \times m \times m_2 \times 4}$ to leverage the structural similarity of these rules.
Since these tensors are probabilities, we must have
\begin{align}
    \sum_{j, k} C^1_{i j k} + \sum_{j, k} D^1_{i j k} = 1, \quad {\forall} i \label{eq:1},\\
   \sum_{j,k} C^2_{ijk} + \sum_{j,k,d} D^2_{i jkd} = 1, \quad {\forall} i \label{eq:2}.
\end{align}

\paragraph{Tensor decomposition.} To scale up the LCFRS-2 to a large number of nonterminals, we first apply CPD on all the binary rule probability tensors,
\begin{align*}
    C^1 &= \sum_{q=0}^{r_1-1} U_{:, q}^{1} \otimes V_{:, q}^{1} \otimes W_{:, q}^{1} \\
    C^2 &= \sum_{q=0}^{r_2-1} U_{:, q}^{2} \otimes V_{:,q}^{2} \otimes W_{:, q}^{2} \\
    D^1 &= \sum_{q=0}^{r_3-1}  U_{:, q}^{3} \otimes V_{:, q}^{3} \otimes W_{:, q}^{3} \\
    D^2 &= \sum_{q=0}^{r_4-1}   U_{:, q}^{4} \otimes V_{:,q}^{4} \otimes W_{:, q}^{4} \otimes P_{:, q}
\end{align*}
where $U_{:,q}$ denotes the $q$-th column of $U$. The dimensions of these tensors are $U^1 \in \mathbb{R}^{m_1 \times r_1}$, $ V^1, W^1 \in \mathbb{R}^{m \times r_1}$, $U^2\in \mathbb{R}^{m_1 \times r_2}$, $V^2 \in \mathbb{R}^{m \times r_2}$, $W^2 \in \mathbb{R}^{m_2 \times r_2}$, $U^3, W^3 \in \mathbb{R}^{m_2 \times r_3}$, $U^4, W^4 \in \mathbb{R}^{m_2 \times r_4}$, $V^3 \in \mathbb{R}^{m \times r_3}$, $V^4 \in \mathbb{R}^{m \times r_4}$, and $P \in \mathbb{R}^{ 4\times r_4}$. Here $r_1, r_2, r_3, r_4$ are the ranks of the tensors that control inference complexity.
To ensure these factorizations lead to valid probability tensors, ~\ref{eq:1}),  we additionally impose the following restrictions:
(1) all decomposed matrices are non-negative; (2) $P, V^{i}, W^{i}$ are {column-wise} normalized where $i\in\{1,2,3,4\}$; (3)  $ \forall i, \sum_{j}U^1_{ij} + \sum_{k}U^2_{ik} = 1$; and (4)
$ \forall i,  \sum_{j}U^3_{ij} + \sum_{k}U^4_{ik} = 1$. It is easy to verify that Eq.~\ref{eq:1} and \ref{eq:2} are satisfied if the above requirements are satisfied.

\paragraph{Rank-space dynamic programming.} For unsupervised learning, we need to compute the marginal likelihood of a sentence $p(w_1w_2\cdots w_{\slen})$. We give the \emph{rank-space} dynamic program (i.e., the inside algorithm) for computing $p(w_1w_2\cdots w_{\slen})$ in this tensor decomposition-based LCFRS-2 in App.~\ref{sec:rank-inside}. The resulting complexity is dominated by $\mathcal{O}(\slen^5r_4 + \slen^4(r_3+r_4)(r_2+r_4))$. We thus  set $r_4$ to a very small value, which greatly improves runtime.

\paragraph{Parameterization.} 
Following prior work on neural parameterizations of grammars \cite{jiang-etal-2016-unsupervised,kim-etal-2019-compound},  we parameterize the component matrices to be the output of neural networks over shared embeddings. 

The symbol embeddings are given by: $E^1 \in \mathbb{R}^{m \times d}$ where the first $m_1$ rows correspond to fan-out-1 nonterminal embeddings and  the last $p$ rows are the preterminal embeddings; $E^2 \in \mathbb{R}^{m_2 \times d}$ for the fan-out-2 nonterminal embedding matrix; $r \in \mathbb{R}^{d}$ for the start symbol embedding. We also have four sets of ``rank embeddings'' $R^1 \in \mathbb{R}^{r_1 \times d}, R^2 \in \mathbb{R}^{r_2 \times d}, R^3 \in \mathbb{R}^{r_3 \times d}$,  and $ R^4 \in \mathbb{R}^{r_4 \times d}$. Given this, the entries of the $U,V,W$ matrices are given by,
\begin{align*}
    &U^{o}_{ij} \propto \exp\{ (R^{o}_j)^\top f^o_{U}(E^1_i) \},  &o\in \{1,2&\} \\
    &U^{o}_{ij} \propto \exp\{(R^{o}_j)^\top f^o_{U}(E^2_i)\},  &o\in \{3,4&\} \\
    &V^{o}_{ij} \propto \exp \{ (R^{o}_j)^\top f^{o}_{V}(E^1_i) \}, &o\in \{1,2,3,4&\} \\
    &W^{o}_{ij} \propto \exp\{ (R^{o}_j)^\top f^o_{W}(E^1_i) \},   &o\in \{1,2\} \\
    &W^{o}_{ij} \propto \exp\{(R^{o}_j)^\top f^o_{W}(E^2_i)\},  &o\in \{3,4&\}
    \end{align*}
where $f_{U}^{o}, f_{V}^{o}, f_{W}^{o}$ are one-layer ReLU MLPs with output size $d$. $U^o, V^o, W^o$ are normalized according to the requirements described in the previous subsection. 
We share the parameters of the following MLP pairs: $(f^{1}_{U},f^{2}_{U})$, $(f^{3}_{U}, f^{4}_{U})$, $(f^{1}_{V}, f^{3}_{V})$, $(f^{2}_{V}, f^{4}_{V})$, $(f^{1}_{W}, f^{3}_{W})$, $(f^{2}_{W}, f^{4}_{W})$ as they play similar roles (e.g., $f^{1}_{V}$ and $f^{3}_{V}$ are both applied to left children).  For the $D^2$ tensor we also require the matrix $P \in \mathbb{R}^{4 \times r_4}$, and this is given by $P^\top = f_{P}(R^4)$, where $f_{P}$ is a one-layer residual network with output size $4$ that is normalized via a $\operatorname{softmax}$ across the last dimension.

Finally, for the starting and the terminal distributions we have
\begin{align*}
s = f_{s}(r), \quad Q = f_{Q}(E^1_{m_1:}),
\end{align*}
which results in $s\in \mathbb{R}^{m_1}$ (i.e., the probability vector for rules of the form $S \to A$) and  $Q \in \mathbb{R}^{p \times v}$ (i.e., probability matrix for rules of the form $T(w) \to w$). Here $E^1_{m_1:}$ is the last $p$ rows of $E^1$, and  $f_{s}$ and $f_Q$ are residual MLPs with $\operatorname{softmax}$ applied in the last layer to ensure that $s$ and $Q$ are valid probabilities.

\paragraph{Decoding.} 
While the rank-space inside algorithm enables efficient computation of sentence likelihoods, direct CKY-style argmax decoding in this grammar requires instantiating the full probability tensors and is thus computationally intractable. We follow~\citet{yang-etal-2021-pcfgs} and use Minimal Bayes Risk (MBR) decoding~\cite{goodman-1996-parsing}. This procedure first  obtains the posterior probability of each span's being a constituent via the inside-outside algorithm (which has the same complexity as the inside algorithm). Then, these posterior probabilities are used as input into CKY in a grammar that only has a single nonterminal. The complexity of this approach is thus independent of the number of nonterminals in the original grammar, and takes $O(\slen^5)$. This strategy can be seen as finding the tree that has the largest number of expected constituents \cite{smith-eisner-2006-minimum}. See App.~\ref{sec:rank-inside} for details.

\section{Empirical Study}

\paragraph{Data.} We conduct experiments with our \textbf{T}ensor decomposition-based \textbf{N}eural \textbf{LCFRS} (TN-LCFRS) on German and Dutch, where discontinuous phenomena are more common (than in English). For German we concatenate TIGER~\cite{Brants2001TheTT} and NEGRA~\cite{skut-etal-1997-annotation} as our training set, while for Dutch we use the LASSY Small Corpus treebank~\cite{DBLP:series/tanlp/NoordBEKLSSV13}. The data split can be found in App.~\ref{sec:data}. For processing we use \texttt{disco-dop}\footnote{\url{https://github.com/andreasvc/disco-dop}}~\cite{vanCranenburgh2016DataOrientedPW} and discard all punctuation marks. We further take the most frequent 10,000 words for each language as the vocabulary, similar to the standard setup in unsupervised constituency parsing on PTB \cite{shen2018nlm,shen2019ordered,kim-etal-2019-compound}.

\paragraph{Grammar size.} To investigate the importance of using a large number of latent variables (which has previously been shown to be helpful for structure induction \cite{DBLP:conf/icml/BuhaiHKRS20,yang-etal-2021-pcfgs}), we train TN-LCFRSs of varying sizes. We first choose the number of preterminals $|\mathcal{P}| \in \{45, 450, 4500\}$ and set the number of fan-out one and fan-out two nonterminals to be $|\mathcal{N}^1| = |\mathcal{N}^2| = \frac{1}{3} |\mathcal{P}|$. The rank of the probability tensors are set to $r_1=r_3=400, r_2=r_4=4$, and the dimensionality of the embedding space is $d=512$. 
Model parameters are initialized with Xavier uniform initialization. More training details and hyperparameters can be found in App.~\ref{sec:training} and App.~\ref{sec:hyper}.

\paragraph{Baselines.}  
Our baselines include: the  neural PCFG (N-PCFG) and the compound PCFG  (C-PCFG) \cite{kim-etal-2019-compound}, which cannot directly predict discontinuous constituents\footnote{But these models could implicitly model discontinuous constituents with a large number of nonterminals (in the neural PCFG case) and/or with a sentence-level random vector (in the compound PCFG case).} but still serve as strong baselines for overall F1 since the majority of spans in these treebanks are continuous; and their direct extensions, neural LCFRS (N-LCFRS) and compound LCFRS (C-LCFRS), which do not employ the tensor-based low-rank factorization. These non-low-rank models have high computational complexity and hence we set $|\mathcal{P}|=45$ for these models. When $|\mathcal{P}|=4500$, we also compare against the tensor decompositional-based neural PCFG (TN-PCFG) from~\citet{yang-etal-2021-pcfgs}.

\paragraph{Evaluation.} We use unlabeled corpus-level F1 to evaluate unsupervised parsing performance, reporting both overall F1 and discontinuous F1 (DF1). For all experiments, we report the mean results and standard deviations  over four  runs with different random seeds.  See App.~\ref{sec:eval} for further details.

\input{tables/main_result}

\subsection{Main results}
Table~\ref{tab:main_result} shows the main results. With smaller grammars ($|\mathcal{P}| = 45$), we find that both neural/compound LCFRSs have lower F1  than their PCFG counterparts, despite being able to predict discontinuous constituent spans.  On the other hand, TN-LCFRS  achieves better F1 than N-LCFRS even though it is a more restricted model (since it assumes that the rule probability tensors are of low rank), showing the benefits of parameter sharing through low rank factorizations. As we scale up TN-LCFRSs with $|\mathcal{P}| \in \{45, 450, 4500\}$ we observe continuous improvements in performance, with TN-LCFRS$_{4500}$ achieving the best F1 and DF1 on all three datasets. 
These results all outperform trivial (left branching, right branching, and random tree) baselines.

As an upper bound we also train a supervised model with TN-LCFRS$_{4500}$.\footnote{For supervised training we  use the optimal binarization from \citet{gildea-2010-optimal} to binarize treebanks and remove all trees that are unrecognizable by our restricted LCFRS.
We fixed the tree topology (provided by gold binarized tree) and used dynamic programming to sum out all possible nonterminals for each node, resulting in the joint log probability of unlabeled binarized tree and sentence. This was then maximized during training.  As for the oracle bound, we emphasize that the gold trees are nonbinary while our model can only predict binary trees. 
} We also show the maximum possible performance with oracle binary trees with this optimal binarization. 

\input{tables/recall_main.tex}
While the discontinuous F1 of our unsupervised parsers are nontrivial, there is still a large gap between the unsupervised and supervised scores (and also between the supervised and the oracle scores), indicating opportunities for further work in this area.

\subsection{Analysis}
\paragraph{Recall by constituent label.} Table~\ref{tab:recall_all} shows the recall by constituent tag for the different models averaged over four independent runs. Overall the unsupervised methods do well on noun phrases (NP), prepositional phrases (PP) and proper nouns (PN), with some of the models approach the supervised baselines. Verb phrases (VP) and adjective phrases (AP) remain challenging. Table~\ref{tab:recall_disco} has recall by label for discontinuous constituents only, where we observe that most discontinuous constituents are VPs. In App.~\ref{appd:f1}
, we also show F1/DF1 broken down by sentence length.

\input{tables/recall_disco.tex}

\input{figures/training_curve.tex}

\paragraph{Approximation error.}
Approximation error in the context of unsupervised learning arises due to the mismatch between the EM objective (i.e., log marginal likelihood) and structure recovery (i.e., F1), and is related to model misspecification \cite{liang-klein-2008-analyzing}. 
Figure~\ref{fig:curve} (left column) plots the training/dev perplexity as well as the dev F1/DF1 as a function of the number of epochs. We find that larger grammars result in better performance in terms of both perplexity and structure recovery, which ostensibly indicates that the unsupervised objective is positively correlated with structure induction performance. 

However, when we first perform supervised learning on the log joint likelihood  and then switch to  unsupervised learning with log marginal likelihood (Figure~\ref{fig:curve}, right), we find that while perplexity improves when we switch to the unsupervised objective, structure induction performance deteriorates.\footnote{
It is worth noting that the phenomenon of mismatch between log marginal likelihood objective and parsing accuracy is quite common in unsupervised grammar induction (and latent variable modeling approaches to structured induction more generally). Many previous works have observed this phenomenon, e.g.,  \citet{merialdo:1994} in the context of HMMs, and \citet{johnson-etal-2007-bayesian} and \citet{liang-klein-2008-analyzing} in the context of PCFGs. This is partially attributed to the fact that generative grammars often make some unreasonable independence assumptions to make the training process tractable, which does not fully comply with the true generative process of human languages and their underlying structures.  
} Still, the difference in F1 before and after switching to the unsupervised objective is less for larger models, confirming the benefits of using  larger grammars.

\paragraph{Even more restricted LCFRS formalisms.} There are even more restricted versions of LCFRSs which have faster parsing (e.g. $\mathcal{O}(\slen^3), \mathcal{O}(\slen^4)$) but can still model discontinuous constituents. In the supervised case, these restricted variants have been shown to perform almost as well as the more expressive  $\mathcal{O}(\slen^5)$ and $\mathcal{O}(\slen^6)$ variants \cite{corro-2020-span}. In the unsupervised case however, we observe in Table~\ref{tab:ablation} that 
disallowing $\mathcal{O}(\slen^5)$   rules ($\redcircled{2b}$, $\redcircled{2c}$, $\redcircled{2d}$, $\redcircled{2e}$) significantly degrades discontinuous F1 scores. We posit that this phenomena is again related to empirical benefits of latent variable overparameterization---while in \emph{theory} it is possible to model most discontinuous phenomena with more restricted rules, making the generative model more expressive via ``overparameterizing'' in rule expressivity space (i.e., using more flexible rules than is necessariy) empirically leads to better performance.

\vspace{-1mm}
\paragraph{Parameter sharing.} As shown in Table~\ref{tab:ablation}, it was  important to share the symbol embeddings across the different rules. Sharing the parameters of the MLPs as described in Sec.~\ref{sec:model} was also found to be helpful. This highlights the benefits of working with neural parameterizations of grammars which enable easy parameter sharing across rules that share symbols and/or have similar shapes.
\input{tables/ablation.tex}

\input{figures/example_tree2.tex}
\vspace{-1mm}
\paragraph{Qualitative analysis.} In Fig.~\ref{fig:tree-comparison}, we show some examples trees in German. For each sentence, we show the gold, TN-LCFRS$_{4500}$, and TN-PCFG$_{4500}$ trees. In the first sentence, the crossing dependency occurs due to the initial adverb (``So'')'s being analyzed as a dependent of the non-finite verb phrase at the end of the sentence which occurs due to German V2 word order. Our parser correctly predicts this dependency, although the subject NP (which itself is correctly identified) has the wrong internal structure.
The second sentence highlights a case of partial success with right-extraposed relative clauses. While our model is able to correctly predict the top-level discontinuous constituent ``[Für 15 200 Mark]$-$[Lampen einbauen  lassen die  mutwilligen Zerstörungen standhalten]'',  the parser does not adopt a discontinuous-constituency analysis of the right-extraposed relative clause itself (``[Lampen]–[die mutwilligen Zerstörungen standhalten]''). Instead it makes the relative clause a part of the non-finite verb complex, which does not conform to the annotation guidelines but resembles an alternative analysis that has been proposed for extraposed relative clauses \citep{baltin-1983:extraposition}. 

Sentence initial adverbs  in the context of auxiliary verb constructions  and right-extraposed relative clauses describe two common instances of discontinuous phenomena in German. Wh- questions constitute another potential class of discontinuous phenomena; however, these are not treated as discontinuous in TIGER/NEGRA. See App.~\ref{appd:example-trees} for more examples trees (including on Dutch).

\section{Related work}

\paragraph{Mildly context-sensitive grammars.} Given the evidence against the context-freeness of natural language  \cite{shieber:cfl}, mildly context-sensitive grammars such as tree adjoining grammars were thought to be just flexible (but still constrained) enough to model natural language \cite{joshi1985much}. Prior work on inducing mildly context-sensitive grammars has generally  focused on combinatory categorial grammars \cite{bisk2012ccg,bisk-hockenmaier-2013-hdp}, and we are unaware of any work on inducing LCFRSs from observed yields alone.  Our work is also related to the rich line of work on supervised discontinuous parsing \cite{kallmeyer-maier-2010-data,maier-etal-2012-plcfrs,maier-2015-discontinuous,corro-2020-span,vilares-gomez-rodriguez-2020-discontinuous,fernandez2020,fernandez-gonzalez-gomez-rodriguez-2021-reducing,FERNANDEZGONZALEZ202343}, though we are unaware of any prior work on unsupervised discontinuous parsing.

 \paragraph{Neural grammars.} Early work on probabilistic approaches to grammar induction was largely negative \cite{lari1990estimation,Carroll1992TwoEO}. However, recent work has shown  that neural parameterizations of classic grammars can greatly improve structure induction. Our work adds to the line of work on neural parameterizations of dependency models \cite{jiang-etal-2016-unsupervised, han-etal-2017-dependency, he-etal-2018-unsupervised, yang-etal-2020-second}, context-free grammars \cite{kim-etal-2019-compound,jin-etal-2019-unsupervised,zhu-etal-2020-return, yang-etal-2021-neural}, and synchronous grammars  \cite{DBLP:conf/nips/Kim21a,wang2022hier,friedman2022finding}.  Neural parameterizations make it easy to share parameters and  condition on additional side information (images/audio/video) which has shown to be particularly useful for multimodal grammar induction \cite{zhao-titov-2020-visually, jin-schuler-2020-grounded,su-etal-2021-dependency,hong2021grammar,zhang-etal-2021-video}.

 \paragraph{Scaling latent variable models.} \citet{DBLP:conf/icml/BuhaiHKRS20} study the empirical benefits of overparameterization in learning latent variable models.    Other works have explored parameterizations of latent variable models that make it especially amenable to scaling \cite{chiu-rush-2020-scaling,DBLP:conf/nips/ChiuDR21,yang-etal-2021-pcfgs,yang-etal-2022-dynamic}. Relatedly, \citet{DBLP:conf/icml/PeharzLVS00BKG20} and \citet{DBLP:journals/corr/abs-2210-04398} show the benefits of scaling probabilistic circuits \cite{Choi2020ProbabilisticCA}.

\section{Conclusion}

This work studied  unsupervised discontinuous constituency parsing with mildly context-sensitive grammars, focusing on the formalism of linear context-free rewriting systems. By using a tensor decomposition-based neural parameterization of linear context-free rewriting systems, our approach was able to induce grammars that had nontrivial discontinuous parsing performance on German and Dutch. Whether  even more expressive grammars will eventually lead to  models learn linguistically meaningful structures and are at the same time competitive with pure neural language models (as a language model) remains an open question.

\section*{Limitations}

There are several limitations of our work.  We tried training the TN-LCFRS on the discontinuous version of the English Penn Treebank \citep[DPTB,][]{evang-kallmeyer-2011-plcfrs} but failed to induce any meaningful discontinuous structures. This is possibly because discontinuous phenomena in English are much less common than in German and Dutch. For example,  while 5.67\% of the gold constituents are discontinuous in NEGRA,  only 1.84\% gold constituents are discontinuous in DPTB \cite{corro-2020-span}.

The neural LCFRS was also quite sensitive to hyperparameters and parameterization. The instability of unsupervised structure induction is widely acknowledged and could  potentially be mitigated by a large amount of training data, as suggested by~\citet{liang-klein-2008-analyzing} and \citet{pate-johnson-2016-grammar}.  Due to this sensitivity, we rely on dev sets for some modeling choices (e.g., rank of the probability tensors). Hence, our approach is arguably not fully unsupervised in the strictest sense of the term, although this is a common setup in unsupervised parsing due to the mismatch between the unsupervised learning objective and structure recovery. (However see \citet{shi-etal-2020-role} for a critical discussion of this approach.)

Finally, while we observed significant increases in performance as we scaled up the number of nonterminals, we also observed diminishing returns. Further scaling up the grammar  is thus unlikely to close the (large) gap that still exists between the unsupervised and supervised parsing results.

\section*{Ethics Statement}

We foresee no ethical concerns with this work.

\section*{Acknowledgment}
SY was supported by the National Natural
Science Foundation of China (61976139). This study was also  supported by funds from an MIT-IBM Watson AI Lab grant.

\vspace{-4mm}
\bibliography{anthology,custom}
\bibliographystyle{acl_natbib}

\clearpage
\appendix

\section{Fast LCFRS Inference with CPD}
\label{sec:rank-inside}
\citet{yang-etal-2022-dynamic} propose a family of CPD-based algorithms  for fast inference in B-FGGs which combine B-graphs~\cite{DBLP:conf/iwpt/KleinM01} and factor graph grammars~\cite[FGG,][]{DBLP:conf/nips/0001R20}. Inference in LCFRS is subsumed by B-FGG because for each rule, the number of variables in the left-hand side is always one. As such, we can adopt the method of~\citet{yang-etal-2022-dynamic} to perform fast dynamic programming inference in ``rank space'' for our restricted LCFRS-2.

Concretely, for a length-$\slen$ sentence $x_0, \dots x_{n-1}$ ($x_j$ is the index in the terminal vocabulary), let $N=n+1$. The inside scores defined in the rank-space (similar to Sec.~\ref{sec:pcfg}) are, 
\begin{itemize}
    \item $\alpha^{A_1},  \alpha^{B_1}, \alpha^{C_1} \in \mathbb{R}^{N \times N \times r_1}$: corresponding to $A, B, C$ in rule $\yellowcircled{1a}$.
    \item $\alpha^{A_2} \in \mathbb{R}^{N \times N \times r_2}$, $\alpha^{B_2} \in \mathbb{R}^{N \times N \times N \times N \times r_2}$, $\alpha^{C_2} \in \mathbb{R}^{N \times N \times r_2}$: corresponding to $A,B,C$ in  rule $\redcircled{2a}$.
        \item $\alpha^{A_3} \in \mathbb{R}^{N \times N \times N \times N \times r_3}$, $\alpha^{B_3}, \alpha^{C_3} \in \mathbb{R}^{N \times N \times r_3} $: corresponding to $A,B,C$ in  rule $\yellowcircled{1b}$.
    \item $\alpha^{A_4} \in \mathbb{R}^{N \times N \times N \times N \times r_4}$, $\alpha^{B_4} \in \mathbb{R}^{N \times N \times r_4}$, $\alpha^{C_4} \in \mathbb{R}^{N \times N \times N \times N \times r_4}$: corresponding to $A,B,C$ in rule $\redcircled{2b}, \redcircled{2c}, \redcircled{2d}, \redcircled{2e}$.
\end{itemize}
The base cases are,
\begin{align*}
\alpha^{B_{o}}_{i, i+1} &=  (Q_{:, x_i})^{T} V^{o}_{m_1:}  &o\in \{1, 2, 3, 4\}\\
\alpha^{C_{o}}_{i, i+1} &=  (Q_{:, x_i})^{T} W^{o}_{m_1:} &o \in \{1, 3\} 
\end{align*}
where $Q_{:, x_i}$ is the $x_i$-th column of $Q$. The recursive DP computation formulas are,
\begin{align}
\alpha^{A_{1}}_{ij} &= \sum_{i<k<j} \alpha^{B_1}_{ik} \circ \alpha^{C_1}_{kj} \nonumber \\
\alpha^{A_{2}}_{ij} &= \sum_{i<m<n<j} \alpha^{B_2}_{mn} \circ \alpha^{C_2}_{imnj} \nonumber \\
\alpha^{A_{3}}_{imnj} &= \alpha^{B_3}_{im} \circ \alpha^{C_3}_{nj} \label{eq:inside1}  \\
\alpha^{A_{4}}_{imnj} &=  \sum_{i<k<m}  \alpha^{B_{4}}_{ik} \circ \alpha^{C_{4}}_{kmnj} \circ P_{0} \nonumber \\ &+  \sum_{i<k<m} \alpha^{B_{4}}_{km}\circ \alpha^{C_4}_{iknj} \circ P_{1} \nonumber \\
&+ \sum_{n<k<j} \alpha^{B_4}_{nk}\circ \alpha^{C_4}_{imkj} \circ P_2 \nonumber \\ &+\sum_{n<k<j} \alpha^{B_4}_{kj}\circ \alpha^{C_4}_{imnk} \circ P_3 \label{eq:inside2}
\end{align}
\begin{align}
\alpha^{B_{o}}_{ij} &= F^{o} \alpha^{A_{1}}_{ij}+ G^{o} \alpha^{A_{2}}_{ij}  &o\in \{1,2,3,4\} \nonumber \\
\alpha^{C_{o}}_{ij} &= H^{o} \alpha^{A_{1}}_{ij}  + I^{o} \alpha^{A_{2}}_{ij}   &o\in \{1,3\} \nonumber \\
\alpha^{C_{o}}_{imnj} &= J^{o} \alpha^{A_3}_{imnj} + K^{o} \alpha^{A_{4}}_{imnj} & o\in \{2,4\}  
\label{eq:inside3}
\end{align}
where
\begin{align*}
    &F^{o} = V^{o}_{:m_1} (U^{1})^{T} & o \in \{1,2,3,4\}\\
    &G^{o} = V^{o}_{:m_1} (U^{2})^{T} & o \in \{1,2,3,4\} \\
    &H^{o} = W^{o}_{:m_1} (U^{1})^{T} & o \in \{1,3\} \\
    &I^{o} = W^{o}_{:m_1} (U^{2})^{T} & o \in \{1,3\} \\
    &J^{o} = W^{o} (U^3)^{T} &  o \in \{2, 4\} \\
    &K^{o} =  W^{o} (U^4)^{T} & o \in \{2, 4\}
\end{align*}
can pre-computed before inference. The partition function $Z$ (i.e., the sentence  likelihood) is then given by,
\[
Z = R_1 \alpha^{A_{1}}_{0n}  + R_2 \alpha^{A_2}_{0n}
\]
where $R_1 =  s^{T} U^{1}$ and $R_2 = s^{T} U^2$. 
\input{tables/mbr.tex}
\input{tables/max_result.tex}

\paragraph{Time complexity.} From the above we can see that Eq.~\ref{eq:inside1} takes $\mathcal{O}(\slen^4r_3)$,    Eq.~\ref{eq:inside2} takes $\mathcal{O}(\slen^5r_4)$,  and Eq.~\ref{eq:inside3} takes  $\mathcal{O}(\slen^4(r_2+r_4)(r_3+r_4))$. Therefore the total time complexity is dominated by $\mathcal{O}(\slen^5r_4+\slen^4(r_2+r_4)(r_3+r_4))$.
\paragraph{MBR decoding.} MBR decoding aims to find the best parse with maximum expected number of constituent spans, which can be decomposed into two steps: i) span marginal estimation, and ii) CKY-style parsing with marginals. Denote continuous and discontinuous span marginals  as $X \in \mathbb{R}^{N \times N}$ and $Y\in \mathbb{R}^{N\times N\times N\times N}$ with $\sum_{ij} X_{ij} + \sum_{ijmn} Y_{ijmn} = 2\slen-1$. Span marginals can be estimated via inside-outside, or equivalently, backprogation on the inside algorithm \cite[][Sec. 6.2]{eisner-2016-inside}, i.e., 
\begin{align*}
    X_{ij} &= \sum_r \sum_{o \in \{1,2\}} \frac{\partial \log Z}{\partial \log \alpha^{A_{o}}_{ijr}}, \\
    Y_{imnj} &= \sum_r \sum_{o \in \{3,4\}}  \frac{\partial \log Z}{ \partial \log\alpha^{A_{o}}_{imnjr}}.
\end{align*}
The second-stage CKY-style parsing is similar to the description in Table~\ref{tab:deduction}, except that the grammar rule probabilities are replaced with span marginals, as described in Table~\ref{tab:mbr}.  The total time complexity is dominated by the first stage of marginal estimation, whose complexity is the same as that of the inside algorithm~\cite{eisner-2016-inside}.

\input{tables/f1_bylength.tex}

\section{Experimental Details}
\subsection{Data split}
\label{sec:data}
For German, we follow \citet{corro-2020-span} and use the NEGRA treebank~\cite{skut-etal-1997-annotation} with the split proposed by~\citet{dubey-keller-2003-probabilistic}, 
and the TIGER treebank~\cite{Brants2001TheTT} with the split provided by the SPRML 2014 shared task~\cite{seddah-etal-2014-introducing}. For Dutch, there is no standard split in the discontinuous parsing literature. We follow \texttt{UD-Dutch-Alpino}~\cite{bouma-van-noord-2017-increasing} and use a hybrid training dataset that comprises the whole Alpino treebank~\cite{Beek2001TheAD} and a subset of LASSY Small Corpus ~\cite{DBLP:series/tanlp/NoordBEKLSSV13}. We further use the whole \texttt{WR-P-P-H} section and \texttt{WR-P-P-L} section as the development and test sets, respectively.

\subsection{Evaluation metric details}
\label{sec:eval}
  Following standard practice in unsupervised parsing evaluation, we ignore all trivial continuous spans, i.e., whole-sentence spans and single-word spans. In addition, we ignore all discontinuous spans of fan-out greater than two. 
Finally, we evaluate only on sentences of length up to 40 due to  computational considerations.

\subsection{Training details}
\label{sec:training}
 For training, we use a curriculum training strategy~\cite{Bengio2009CurriculumL} where we train only on sentences of length up to 30 in the first epoch, and increase the maximum length by five for each epoch until  we reach the maximum sentence length (60  for Dutch and 40 for German). We use the Adam optimizer~\cite{Kingma2015AdamAM} with $\beta_1 = 0.75, \beta_2 = 0.999$, learning rate $0.002$,  batch size $20$, and a maximum gradient norm limit of 3. We train for 20 epochs and perform early stopping strategy based on the performance of  development set with maximum patience 5.

\subsection{Choice of hyperparameters}
\label{sec:hyper}
We assumed a 1:1 ratio between the numbers of fan-out one and fan-out two nonterminals and tuned the ratio of the number of fan-out one nonterminals to preterminals from $\{\frac{1}{2}, \frac{1}{3}, \frac{1}{4}\}$. Since C-LCFRS and N-LCFRS are computationally expensive without tensor decomposition, we could only use up to 45 preterminals and 15 fan-out one/two nonterminals. We then scaled up our approach by a factor of 10 and 1000 to study the benefits of overparameterization, which resulted in our final choice. 

Regarding rank size, we used as much as we could while keeping the ratio of $\frac{r_1}{r_3}=\frac{r_2}{r_4}=100$. To save tuning time, we assumed $r_1=r_2$ and $r_3=r_4$. Due to the high computational complexity, we used $r_1$ up to 400. It is important to note that we cannot use a ratio of $\frac{r_1}{r_3}$ or $\frac{r_2}{r_4}$ arbitrarily, such as 80:20 or 50:50. We observed much lower total F1 scores (much more discontinuous spans would be predicted) when using such ratios in our experiments. This is because $\frac{r_3}{r_1+r_3}$ can be regarded as the prior probability (when the network is randomly initialized) of having a discontinuous child for a fan-out-1 parent node.  If the ratio of $\frac{r_3}{r_1+r_3}$ is too high, the model will predict many discontinuous spans from the beginning. Unsupervised learning will use the expected counts from the start for feedback self-supervised learning, resulting in the grammar learned at the end predicting many more discontinuous spans.

\section{Additional results}
\label{appd:f1} 
\label{appd:max}
Table~\ref{tab:max_result} shows the maximum performance across four seeds, while Table~\ref{tab:length_result} gives the F1 broken down by sentence length on TIGER.

\section{Additional example trees}
We show some additional trees on German in Fig.~\ref{fig:appd-example-german} and on Dutch in Fig.~\ref{fig:appd-example-dutch}.
\label{appd:example-trees}
\input{figures/appd_example_trees.tex}
\input{figures/appd_tree_dutch.tex}

\end{document}

%% file: figures/tree.tex
\begin{figure}

    \centering
\begin{tikzpicture}[scale=0.8, align=center,
                           ]
\footnotesize
\path
        (4.25, 4) node (n1) { \textcolor{black}{S(1)} }
        (5.5, 3) node (n2) { \textcolor{black}{VP(2)} }
        (0.625, 2) node (n3) { \textcolor{black}{PP(1)} }
        (6, 2) node (n8) { \textcolor{black}{PP(1)} }
        (0, 1) node (n4) { \textcolor{black}{APPR} }
        (1.25, 1) node (n6) { \textcolor{black}{NN} }
        (2.75, 1) node (n15) { \textcolor{black}{VAFIN} }
        (4.25, 1) node (n17) { \textcolor{black}{NE} }
        (5.5, 1) node (n9) { \textcolor{black}{APPR} }
        (6.5, 1) node (n11) { \textcolor{black}{NE} }
        (8, 1) node (n13) { \textcolor{black}{VVPP} }
        (0, 0) node (n5) { \textcolor{black}{Im} }
        (1.25, 0) node (n7) { \textcolor{black}{Oktober} }
        (2.75, 0) node (n16) { \textcolor{black}{wurde} }
        (4.25, 0) node (n18) { \textcolor{black}{Passent} }
        (5.5, 0) node (n10) { \textcolor{black}{in} }
        (6.5, 0) node (n12) { \textcolor{black}{Berlin} }
        (8, 0) node (n14) { \textcolor{black}{operiert} }
        (0, -0.5) node (m5) { \textcolor{black}{\textit{In}} }
        (1.25, -0.5) node (m7) { \textcolor{black}{\textit{Octobor} }}
        (2.75, -0.55) node (m16) { \textcolor{black}{\textit{surgery}}}
        (4.25, -0.5) node (m18) { \textcolor{black}{\textit{Passent}}}
        (5.5, -0.5) node (m10) { \textcolor{black}{\textit{in}} }
        (6.5, -0.5) node (m12) { \textcolor{black}{\textit{Berlin}} }
        (8, -0.5) node (14) { \textcolor{black}{\textit{underwent}}}
;
\draw (n17) -- +(0, -0.5) -| (n18);
\draw (n15) -- +(0, -0.5) -| (n16);
\draw (n13) -- +(0, -0.5) -| (n14);
\draw (n11) -- +(0, -0.5) -| (n12);
\draw (n9) -- +(0, -0.5) -| (n10);
\draw (n6) -- +(0, -0.5) -| (n7);
\draw (n4) -- +(0, -0.5) -| (n5);
\draw (n8) -- +(0, -0.5) -| (n9);
\draw (n8) -- +(0, -0.5) -| (n11);
\draw (n3) -- +(0, -0.5) -| (n4);
\draw (n3) -- +(0, -0.5) -| (n6);
\draw [white, -, line width=6pt] (n2)  +(0, -0.5) -| (n3);
\draw (n2) -- +(0, -0.5) -| (n3);
\draw [white, -, line width=6pt] (n2)  +(0, -0.5) -| (n8);
\draw (n2) -- +(0, -0.5) -| (n8);
\draw [white, -, line width=6pt] (n2)  +(0, -0.5) -| (n13);
\draw (n2) -- +(0, -0.5) -| (n13);
\draw (n1) -- +(0, -0.5) -| (n2);
\draw (n1) -- +(0, -0.5) -| (n15);
\draw (n1) -- +(0, -0.5) -| (n17);
\end{tikzpicture}

    \caption{An example of a discontinuous parse tree in German. Each non-leaf node's fan-out is marked in brackets.
    \label{fig:dis-tree}
    }

\end{figure}
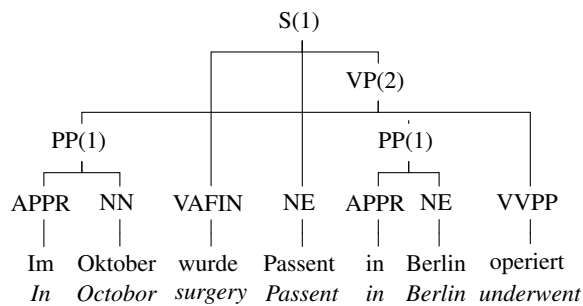

%% file: tables/deduction.tex
\begin{table}[tb]
	\centering
	\scriptsize
	{\setlength{\tabcolsep}{.0em}
		\begin{tabular}{r}
			\toprule 
			\begin{minipage}{\linewidth}
    \textit{Item form}: 
$\bigfrac{\text{$[A, i, j]$: fan-out-1 node $A$ spanning $[i, j)$}}{\text{$[A, i, j, k, n]$: fan-out-2 node $A$ spanning $[i, j), [k,n)$}}$
    \textit{Axioms}: \hspace{4mm} $[A, i, i+1]$, $0 \le i < \slen + 1, A \in \mathcal{N}^1$ \\ 
    \textit{Goals}: \hspace{5mm} $[S, 0, n]$ \\ 
    \textit{Deductive rules:} 
    \vspace{-1mm}
 \\
 \\
          \infer[\bigfrac{A(xy) \rightarrow B(x)C(y)}{i < k < j } \phantom{jj} \yellowcircled{1a}]{[A, i, j]}{ [B, i, k]& [C, k,j]}  \vspace{0.2mm} \\ 
         \infer[\bigfrac{A(x, y) \rightarrow  B(x)C(y)}{i < j < m < n }\phantom{jj}\yellowcircled{1b}]{[A, i, j, m, n]}{ [B, i, j]& [C, m, n]} 
 \vspace{0.2mm} \\
        \phantom{} \infer[\bigfrac{A(yxz) \rightarrow  B(x)C(y,z)}{i < m < n < j}\phantom{j}\redcircled{2a}]{[A, i, j]}{ [B, m, n]& [C, i, m, n, j]} \\
       \phantom{} \infer[\bigfrac{A(xy,z) \rightarrow  B(x)C(y,z)}{i < k < j < m < n}\phantom{j}\redcircled{2b}]{[A, i, j, m, n]}{ [B, i, k]& [C, k, j, m, n]} \\
            \phantom{} \infer[\bigfrac{A(yx, z) \rightarrow  B(x)C(y,z)}{i < k < j < m < n}\phantom{j}\redcircled{2c}]{[A, i, j, m, n]}{ [B, k, j]& [C, i, k, m, n]}  \\
              \phantom{} \infer[\bigfrac{A(y, xz) \rightarrow  B(x)C(y,z)}{i < j < m < k < n}\phantom{j}\redcircled{2d}]{[A, i, j, m, n]}{ [B, m, k]& [C, i, j, k, n]}  \\
              \phantom{} \infer[\bigfrac{A(y, zx) \rightarrow  B(x)C(y,z)}{ i < j < m < k < n} \phantom{j} \redcircled{2e}]{[A, i, j, m, n]}{ [B, m, k]& [C, i, j, k, n]} 
			\end{minipage}\\
			\bottomrule
	\end{tabular}}
 \vspace{-2mm}
	\caption{Chart parsing algorithm described in the parsing-as-deduction framework. Here $\slen$ is the sentence length and we use interstice indices (not word indices) as in \citet{corro-2020-span}.}
	\label{tab:deduction}
 \vspace{-2mm}
\end{table}

%% file: figures/rule.tex
	\begin{figure}[!h]
\centering
	\begin{tikzpicture}[scale=.8,  cst/.style={circle,fill=black,inner sep=2pt}, every label/.style={label distance=0.1cm,inner sep=0pt}]
\node[label={[label distance=-0.2cm]below:\strut\scriptsize $i$}] (i) at (0, 0) {};
\node[label={[label distance=-0.2cm]below:\strut\scriptsize $k$}] (m) at (1,0) {};
\node[cst,label=left:{\scriptsize $B(x)$}] (t1) at (0.5,1) {};
\node[cst,label=right:{\scriptsize $C(y, z)$}] (t2) at (1.5,1) {};
\node[cst,label=above:{\scriptsize $A(xy, z)$}] (t3) at (1,1.5) {};
\draw[->] (t3) -- (t1);
\draw[->] (t3) -- (t2);

\node[label={[label distance=-0.2cm]below:\strut\scriptsize $j$}] (k) at (2,0) {};
\node[label={[label distance=-0.2cm]below:\strut\scriptsize $m$}] (l) at (3,0) {};
\node[label={[label distance=-0.2cm]below:\strut\scriptsize $n$}] (j) at (4,0) {};
\draw[fill=yellow!10] (t1.center) -- (i.center) -- (m.center) -- (t1.center);
\draw[fill=blue!10] (t2.center) -- (j.center) -- (l.center) -- (t2.center) -- (k.center) --(m.center) -- (t2.center);
\node[label={[label distance=-0.2cm]below:\strut\scriptsize $x$}] (x) at (0.5, 0.5) {};
\node[label={[label distance=-0.2cm]below:\strut\scriptsize $y$}] (y) at (1.5, 0.5) {};
\node[label={[label distance=-0.2cm]below:\strut\scriptsize $z$}] (z) at (3.05, 0.5) {};

\end{tikzpicture}
\end{figure}
\vspace{-2mm}

%% file: tables/main_result.tex
\begin{table*}[tb!]
\footnotesize

    \centering
    \begin{tabular}{llllllll}
        \toprule 
   & 
    & \multicolumn{2}{c}{{\bf NEGRA}} & \multicolumn{2}{c}{{\bf TIGER}}& \multicolumn{2}{c}{\bf LASSY}\\
    {\bf Model}   & $|\mathcal{P}|$   & F1 &   DF1 & F1  & DF1 & F1 &  DF1 \\
        \midrule 
    N-PCFG & 45 & 40.8$_{\pm 0.5}$ &  $-$  & 39.5$_{\pm 0.4}$ & $-$  & 40.1$_{\pm 3.9}$ & $-$ \\ 
    C-PCFG & 45 & 39.1$_{\pm 1.9}$ & $-$  & 38.8$_{\pm 1.3}$ & $-$  & 37.9$_{\pm 3.4}$ & $-$ \\ 
    N-LCFRS & 45 & 33.7$_{\pm 2.8}$ & 2.0$_{\pm 0.8}$ & 32.7$_{\pm 1.8}$ & 1.2$_{\pm 0.8}$ & 36.9$_{\pm 1.5}$ &    0.9$_{\pm 0.8}$ \\
    C-LCFRS & 45 & 36.7$_{\pm 1.5}$ & 2.7$_{\pm 1.4}$ & 35.2$_{\pm 1.2}$ & 1.7$_{\pm 1.1}$ & 36.9$_{\pm 3.7}$ &  2.2$_{\pm 1.0}$\\
    TN-LCFRS & 45 & 41.1$_{\pm 1.2}$ & 3.1$_{\pm 2.8}$ & 40.2$_{\pm 1.1}$ & 2.3$_{\pm 2.3}$ & 41.6$_{\pm 3.0}$ & 2.3$_{\pm 2.3}$ \\
    TN-LCFRS & 450 & 45.0$_{\pm 1.8}$ & 5.6$_{\pm 2.7}$ & 44.1$_{\pm 1.7}$ &  4.4$_{\pm 2.3}$ & 42.9$_{\pm 3.8}$ & 2.8$_{\pm 3.3} $ \\
    TN-PCFG & 4500 & 45.4$_{\pm 0.5}$ & $-$ & 44.7$_{\pm 0.6}$ & $-$  & 44.3$_{\pm 6.4}$ & $-$ \\
        TN-LCFRS & 4500 & \textbf{46.1}$_{\pm 1.1}$ & \textbf{8.0}$_{\pm 1.1}$ & \textbf{45.4}$_{\pm 0.9}$ & \textbf{6.1}$_{\pm 0.8}$  & \textbf{45.6}$_{\pm 2.3}$ & \textbf{8.9}$_{\pm 1.5}$ \\
    \midrule
    Supervised & 4500 & 54.4$_{\pm 0.3}$ & 38.1$_{\pm 1.1}$ & 50.7$_{\pm 0.2}$ & 32.1$_{\pm 1.0}$ & $-$  &  $-$  \\ 
    Left branching & $-$  & 7.8 & $-$ & 7.9 & $-$ &  7.2 & $-$ \\
    Right branching & $-$  & 12.9 & $-$ & 14.5 & $-$ & 24.1 & $-$ \\
    Random trees & $-$  & 7.0$_{\pm 0.1}$ &$-$ & 7.1$_{\pm 0.2}$ & $-$ &  9.1$_{\pm 0.4}$ & $-$\\
    Oracle bound &  $-$  & 64.3 & 88.5 & 65.0 & 86.2 & 73.7 & 68.0\\ 
    \bottomrule 
    \end{tabular}
    \vspace{-2mm}
    \caption{Results on test sets of German (NEGRA, TIGER) and Dutch (LASSY) treebanks for the various models. $|\mathcal{P}|$ indicates the number of preterminals, which also determines the number of nonterminals ($|\mathcal{N}^1| = |\mathcal{N}^1| = \frac{1}{3} |\mathcal{P}|$), and thus grammar size. F1 is the overall F1 for both continuous and discontinuous spans, while DF1 is the F1 on discontinuous spans only. These results are averaged across four seeds, and $_\pm$ indicates standard deviation. Oracle bound shows the upper bound obtainable from binarized trees.}
    \label{tab:main_result}
    \vspace{-2mm}
\end{table*}

%% file: tables/recall_main.tex
\begin{table}[t!] 	 

	{\setlength{\tabcolsep}{.9em}
		\makebox[\linewidth]{\resizebox{\linewidth}{!}{%
	   \begin{tabular}{llllll}
					\toprule 
				  &  NP & PP & VP &
				  AP &  PN  \\	
				  \toprule
	 count & 10236 & 8471 & 3312 & 1375 & 1249 \\
	 \midrule
	 				  N-PCFG$_{45}$ &   71.5 & 78.4 & 37.5 & 31.5 & 44.1\\
	 		      C-PCFG$_{45}$ & 67.3 & 79.4 & 31.1 & 29.0 & 51.2\\
   				  N-LCFRS$_{45}$ & 60.9 & 70.5 & 25.8 & 29.9 & 40.8\\
				  C-LCFRS$_{45}$ & 58.6 & 72.6 & 28.6 & 33.0 & 24.0\\
 				  TN-LCFRS$_{45}$ & 73.3 & 76.1 & 34.1 & 27.7 & 69.7 \\
 				  TN-LCFRS$_{450}$ & 77.6 & \textbf{84.2} & 30.6 & 42.8 & 72.1 \\
				  TN-PCFG$_{4500}$ & 76.5 & 81.8 & \textbf{51.4} & 41.3 & 67.9\\
				  TN-LCFRS$_{4500}$ & \textbf{78.7} & 83.7 & 46.1 & \textbf{55.8} & \textbf{73.6}\\
				  \midrule
				  Supervised &  78.8 & 86.1 & 60.9 & 74.3 & 79.0 \\
			     \bottomrule
	\end{tabular}}}}
 \vspace{-2mm}
	\caption{\label{tab:recall_all}
	Recall ($\%$) of the most five frequent constituent labels on the TIGER test set. }
		\vspace{-2mm}

\end{table}

%% file: tables/recall_disco.tex
\begin{table}[t!] 	 

	{\setlength{\tabcolsep}{.9em}
		\makebox[\linewidth]{\resizebox{\linewidth}{!}{%
	   \begin{tabular}{llllll}
					\toprule 
				  &  VP & NP & PP &
				  AP &  AVP  \\	
				  \toprule
	 count & 1195 & 395 & 172 & 84 & 71 \\
	 \midrule
   				  N-LCFRS$_{45}$ & 10.3 & \textbf{4.8} & \textbf{1.9} & \textbf{2.4} & \textbf{2.1}\\
				  C-LCFRS$_{45}$ & 11.8 & 2.2 & 1.0 & 2.7 & 0.4\\
 				  TN-LCFRS$_{45}$ & 6.0 & 3.0 & 1.2 & 0.3 & 1.1 \\
 				  TN-LCFRS$_{450}$ & 11.9 & 2.2 & 0.3 & 1.2 & 0.4 \\
				  TN-LCFRS$_{4500}$ & \textbf{19.9} & 2.5 & 0.0 & 0.9 & 0.4 \\
				  \midrule 
				  Supervised & 23.7 & 14.1 & 31.7 & 18.5 & 25.4 \\
			     \bottomrule
	\end{tabular}}}}
 \vspace{-2mm}
	\caption{\label{tab:recall_disco}
	Recall ($\%$) of the most five frequent discontinuou constituent labels on the TIGER test set. }
	\vspace{-2mm}
	
\end{table}

%% file: figures/training_curve.tex
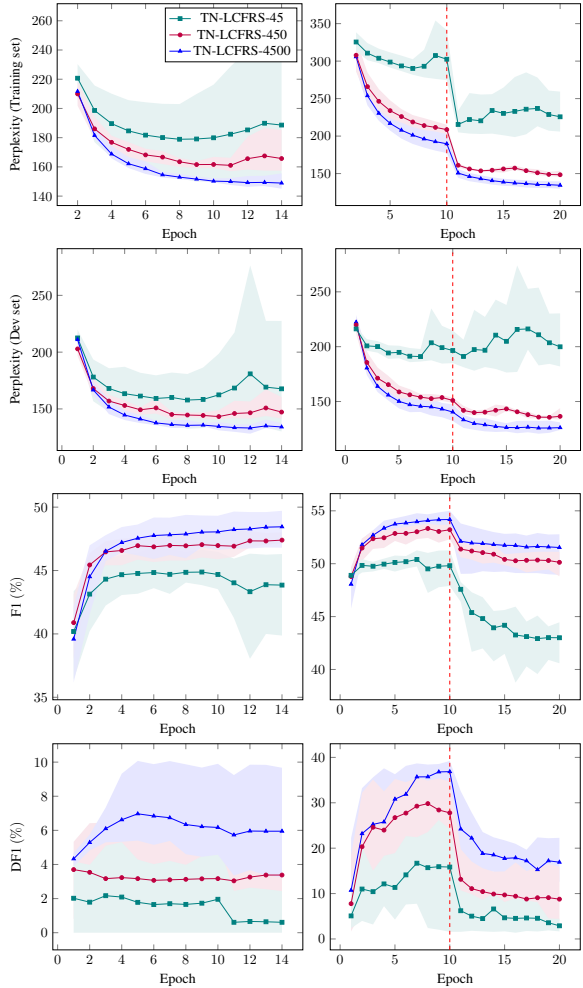
\begin{figure}[tb!]
    \vspace{-2mm}
\centering
\begin{subfigure}{\linewidth}
    \scalebox{0.47}{
\begin{tikzpicture}
\begin{axis}[ ylabel={Perplexity (Training set)}, xlabel={Epoch},
ylabel near ticks,
]
\addplot [teal, mark=square*,mark size=1.5pt] table[x=x,y=y] {ppl_train_45_unsup.dat};
\addlegendentry{TN-LCFRS-45};
\addplot[purple, mark=*,mark size=1.5pt] table[x=x,y=y] {ppl_train_450_unsup.dat};
\addlegendentry{TN-LCFRS-450};
\addplot [blue, mark=triangle*,mark size=1.5pt] table[x=x,y=y] {ppl_train_4500_unsup.dat};
\addlegendentry{TN-LCFRS-4500};

\addplot [name path=upper,draw=none] table[x=x,y expr=\thisrow{max}] {ppl_train_45_unsup.dat};
\addplot [name path=lower,draw=none] table[x=x,y expr=\thisrow{min}] {ppl_train_45_unsup.dat};
\addplot [fill=teal!10] fill between[of=upper and lower];

\addplot [name path=upper,draw=none] table[x=x,y expr=\thisrow{max}] {ppl_train_450_unsup.dat};
\addplot [name path=lower,draw=none] table[x=x,y expr=\thisrow{min}] {ppl_train_450_unsup.dat};
\addplot [fill=purple!10] fill between[of=upper and lower];

\addplot [name path=upper,draw=none] table[x=x,y expr=\thisrow{max}] {ppl_train_4500_unsup.dat};
\addplot [name path=lower,draw=none] table[x=x,y expr=\thisrow{min}] {ppl_train_4500_unsup.dat};
\addplot [fill=blue!10] fill between[of=upper and lower];
\end{axis}
\end{tikzpicture}

\begin{tikzpicture}
\begin{axis}[ ylabel={}, xlabel={Epoch}]
\addplot [blue, mark=triangle*,mark size=1.5pt] table[x=x,y=y] {ppl_train_4500_sup.dat};
\addplot [name path=upper,draw=none] table[x=x,y expr=\thisrow{max}] {ppl_train_4500_sup.dat};
\addplot [name path=lower,draw=none] table[x=x,y expr=\thisrow{min}] {ppl_train_4500_sup.dat};
\addplot [fill=blue!10] fill between[of=upper and lower];

\addplot[purple, mark=*,mark size=1.5pt] table[x=x,y=y] {ppl_train_450_sup.dat};
\addplot [name path=upper,draw=none] table[x=x,y expr=\thisrow{max}] {ppl_train_450_sup.dat};
\addplot [name path=lower,draw=none] table[x=x,y expr=\thisrow{min}] {ppl_train_450_sup.dat};
\addplot [fill=purple!10] fill between[of=upper and lower];

\addplot [teal, mark=square*,mark size=1.5pt] table[x=x,y=y] {ppl_train_45_sup.dat};
\addplot [name path=upper,draw=none] table[x=x,y expr=\thisrow{max}] {ppl_train_45_sup.dat};
\addplot [name path=lower,draw=none] table[x=x,y expr=\thisrow{min}] {ppl_train_45_sup.dat};
\addplot [fill=teal!10] fill between[of=upper and lower];
 \draw [red, dashed] (axis cs:10,\pgfkeysvalueof{/pgfplots/ymin})--(axis cs:10,\pgfkeysvalueof{/pgfplots/ymax});

\end{axis}
\end{tikzpicture}

}
\end{subfigure}

\begin{subfigure}{\linewidth}
    \scalebox{0.47}{
\begin{tikzpicture}
\begin{axis}[ ylabel={Perplexity (Dev set)}, xlabel={Epoch},
ylabel near ticks,
]
\addplot [teal, mark=square*,mark size=1.5pt] table[x=x,y=y] {ppl_dev_45_unsup.dat};
\addplot [name path=upper,draw=none] table[x=x,y expr=\thisrow{max}] {ppl_dev_45_unsup.dat};
\addplot [name path=lower,draw=none] table[x=x,y expr=\thisrow{min}] {ppl_dev_45_unsup.dat};
\addplot [fill=teal!10] fill between[of=upper and lower];

\addplot[purple, mark=*,mark size=1.5pt] table[x=x,y=y] {ppl_dev_450_unsup.dat};
\addplot [name path=upper,draw=none] table[x=x,y expr=\thisrow{max}] {ppl_dev_450_unsup.dat};
\addplot [name path=lower,draw=none] table[x=x,y expr=\thisrow{min}] {ppl_dev_450_unsup.dat};
\addplot [fill=purple!10] fill between[of=upper and lower];
\addplot [blue, mark=triangle*,mark size=1.5pt] table[x=x,y=y] {ppl_dev_4500_unsup.dat};
\addplot [name path=upper,draw=none] table[x=x,y expr=\thisrow{max}] {ppl_dev_4500_unsup.dat};
\addplot [name path=lower,draw=none] table[x=x,y expr=\thisrow{min}] {ppl_dev_4500_unsup.dat};
\addplot [fill=blue!10] fill between[of=upper and lower];

\end{axis}
\end{tikzpicture}

\begin{tikzpicture}
\begin{axis}[ ylabel={}, xlabel={Epoch}]
\addplot [blue, mark=triangle*,mark size=1.5pt] table[x=x,y=y] {ppl_dev_4500_sup.dat};
\addplot [name path=upper,draw=none] table[x=x,y expr=\thisrow{max}] {ppl_dev_4500_sup.dat};
\addplot [name path=lower,draw=none] table[x=x,y expr=\thisrow{min}] {ppl_dev_4500_sup.dat};
\addplot [fill=blue!10] fill between[of=upper and lower];

\addplot[purple, mark=*,mark size=1.5pt] table[x=x,y=y] {ppl_dev_450_sup.dat};
\addplot [name path=upper,draw=none] table[x=x,y expr=\thisrow{max}] {ppl_dev_450_sup.dat};
\addplot [name path=lower,draw=none] table[x=x,y expr=\thisrow{min}] {ppl_dev_450_sup.dat};
\addplot [fill=purple!10] fill between[of=upper and lower];

\addplot [teal, mark=square*,mark size=1.5pt] table[x=x,y=y] {ppl_dev_45_sup.dat};
\addplot [name path=upper,draw=none] table[x=x,y expr=\thisrow{max}] {ppl_dev_45_sup.dat};
\addplot [name path=lower,draw=none] table[x=x,y expr=\thisrow{min}] {ppl_dev_45_sup.dat};
\addplot [fill=teal!10] fill between[of=upper and lower];
 \draw [red, dashed] (axis cs:10,\pgfkeysvalueof{/pgfplots/ymin})--(axis cs:10,\pgfkeysvalueof{/pgfplots/ymax});

\end{axis}
\end{tikzpicture}

}
\end{subfigure}

\begin{subfigure}{\linewidth}
    \scalebox{0.48}{
\begin{tikzpicture}
\begin{axis}[ ylabel={F1  $(\/ \%)$}, xlabel={Epoch},
ylabel near ticks,
]
\addplot [teal, mark=square*,mark size=1.5pt] table[x=x,y=y] {f_45_unsup.dat};
\addplot [name path=upper,draw=none] table[x=x,y expr=\thisrow{max}]{f_45_unsup.dat};
\addplot [name path=lower,draw=none] table[x=x,y expr=\thisrow{min}]{f_45_unsup.dat};
\addplot [fill=teal!10] fill between[of=upper and lower];
\addplot[purple, mark=*,mark size=1.5pt] table[x=x,y=y] {f_450_unsup.dat};
\addplot [name path=upper,draw=none] table[x=x,y expr=\thisrow{max}] {f_450_unsup.dat};
\addplot [name path=lower,draw=none] table[x=x,y expr=\thisrow{min}] {f_450_unsup.dat};
\addplot [fill=purple!10,opacity=0.8] fill between[of=upper and lower];
\addplot [blue, mark=triangle*,mark size=1.5pt] table[x=x,y=y] {f_4500_unsup.dat};
\addplot [name path=upper,draw=none] table[x=x,y expr=\thisrow{max}] {f_4500_unsup.dat};
\addplot [name path=lower,draw=none] table[x=x,y expr=\thisrow{min}] {f_4500_unsup.dat};
\addplot [fill=blue!10,opacity=0.8] fill between[of=upper and lower];

\end{axis}
\end{tikzpicture}

\begin{tikzpicture}
\begin{axis}[ ylabel={}, xlabel={Epoch},
ylabel near ticks,
]
\addplot [blue, mark=triangle*,mark size=1.5pt] table[x=x,y=y] {f_4500_sup.dat};
\addplot [name path=upper,draw=none] table[x=x,y expr=\thisrow{max}] {f_4500_sup.dat};
\addplot [name path=lower,draw=none] table[x=x,y expr=\thisrow{min}] {f_4500_sup.dat};
\addplot [fill=blue!10] fill between[of=upper and lower];

\addplot[purple, mark=*,mark size=1.5pt] table[x=x,y=y] {f_450_sup.dat};
\addplot [name path=upper,draw=none] table[x=x,y expr=\thisrow{max}] {f_450_sup.dat};
\addplot [name path=lower,draw=none] table[x=x,y expr=\thisrow{min}] {f_450_sup.dat};
\addplot [fill=purple!10] fill between[of=upper and lower];

\addplot [teal, mark=square*,mark size=1.5pt] table[x=x,y=y] {f_45_sup.dat};
\addplot [name path=upper,draw=none] table[x=x,y expr=\thisrow{max}] {f_45_sup.dat};
\addplot [name path=lower,draw=none] table[x=x,y expr=\thisrow{min}] {f_45_sup.dat};
\addplot [fill=teal!10] fill between[of=upper and lower];
 \draw [red, dashed] (axis cs:10,\pgfkeysvalueof{/pgfplots/ymin})--(axis cs:10,\pgfkeysvalueof{/pgfplots/ymax});

\end{axis}
\end{tikzpicture}

}
\end{subfigure}

\begin{subfigure}{\linewidth}
    \scalebox{0.48}{
\begin{tikzpicture}
\begin{axis}[ ylabel={DF1  $(\/ \%)$}, xlabel={Epoch},
ylabel near ticks,
]

\addplot[purple, mark=*,mark size=1.5pt] table[x=x,y=y] {df_450_unsup.dat};
\addplot [name path=upper,draw=none] table[x=x,y expr=\thisrow{max}] {df_450_unsup.dat};
\addplot [name path=lower,draw=none] table[x=x,y expr=\thisrow{min}] {df_450_unsup.dat};
\addplot [fill=purple!10] fill between[of=upper and lower];

\addplot [teal, mark=square*,mark size=1.5pt] table[x=x,y=y] {df_45_unsup.dat};
\addplot [name path=upper,draw=none] table[x=x,y expr=\thisrow{max}] {df_45_unsup.dat};
\addplot [name path=lower,draw=none] table[x=x,y expr=\thisrow{min}] {df_45_unsup.dat};
\addplot [fill=teal!10] fill between[of=upper and lower];

\addplot [blue, mark=triangle*,mark size=1.5pt] table[x=x,y=y] {df_4500_unsup.dat};
\addplot [name path=upper,draw=none] table[x=x,y expr=\thisrow{max}] {df_4500_unsup.dat};
\addplot [name path=lower,draw=none] table[x=x,y expr=\thisrow{min}] {df_4500_unsup.dat};
\addplot [fill=blue!10] fill between[of=upper and lower];

\end{axis}
\end{tikzpicture}

\begin{tikzpicture}
\begin{axis}[ ylabel={}, xlabel={Epoch},
ylabel near ticks,
]
\addplot [blue, mark=triangle*,mark size=1.5pt] table[x=x,y=y] {df_4500_sup.dat};
\addplot [name path=upper,draw=none] table[x=x,y expr=\thisrow{max}] {df_4500_sup.dat};
\addplot [name path=lower,draw=none] table[x=x,y expr=\thisrow{min}] {df_4500_sup.dat};
\addplot [fill=blue!10] fill between[of=upper and lower];

\addplot[purple, mark=*,mark size=1.5pt] table[x=x,y=y] {df_450_sup.dat};
\addplot [name path=upper,draw=none] table[x=x,y expr=\thisrow{max}] {df_450_sup.dat};
\addplot [name path=lower,draw=none] table[x=x,y expr=\thisrow{min}] {df_450_sup.dat};
\addplot [fill=purple!10] fill between[of=upper and lower];

\addplot [teal, mark=square*,mark size=1.5pt] table[x=x,y=y] {df_45_sup.dat};
\addplot [name path=upper,draw=none] table[x=x,y expr=\thisrow{max}] {df_45_sup.dat};
\addplot [name path=lower,draw=none] table[x=x,y expr=\thisrow{min}] {df_45_sup.dat};
\addplot [fill=teal!10] fill between[of=upper and lower];
 \draw [red, dashed] (axis cs:10,\pgfkeysvalueof{/pgfplots/ymin})--(axis cs:10,\pgfkeysvalueof{/pgfplots/ymax});

\end{axis}
\end{tikzpicture}

}
\end{subfigure}
\vspace{-3mm}
    \caption{On the rows we have the German (TIGER) training set perplexities, dev set perplexities, overall F1, and discontinuous F1 (DF1) for TN-LCFRSs of various sizes as a function of training epochs. Colored regions indicate min/max values across four runs.
    Left column shows pure unsupervsed learning, while right column shows case where we train with supervised learning for 10 epochs and then switch to unsupervised learning (indicated by dashed lines).}
    \label{fig:curve}
\vspace{-6mm}
\end{figure}

%% file: tables/ablation.tex
\begin{table}[tb!]
    \vspace{-2mm}
    \centering 
    \small
    \begin{tabular}{lllll}
        \toprule 
    {\bf Model} & 
     \multicolumn{2}{c}{{\bf NEGRA}} & \multicolumn{2}{c}{{\bf TIGER}}\\
      & F1 &   DF1 & F1  & DF1\\
        \midrule 
    TN-LCFRS$_{4500}$ & 46.1 & 8.0 & 45.4 & 6.1 \\
 \small{w/o $\mathcal{O}(n^5)$ rules} & 46.4 & 4.0 & 45.3 & 3.0 \\
    \small{w/o shared MLPs }  & 44.4 & 6.7 & 43.6 & 5.3 \\
    \small{w/o shared emb.} & 45.4 & 0.9 & 44.5 & 0.5\\
     \bottomrule 
    \end{tabular}
    \vspace{-2mm}
    \caption{Ablation studies on the German (TIGER) treebank.}
    \vspace{-5mm}
    \label{tab:ablation}
\end{table}

%% file: figures/example_tree2.tex
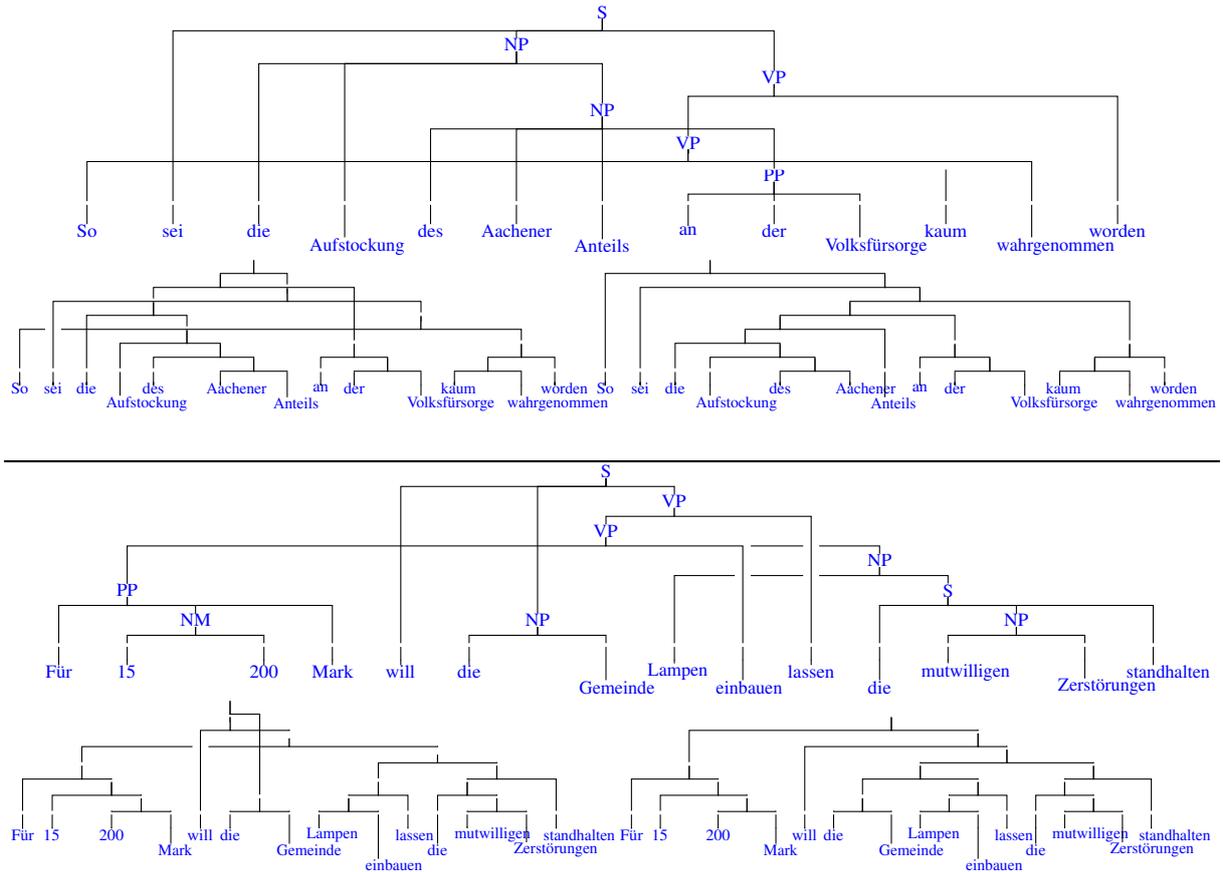
\begin{figure*}[t!]

\centering
\begin{subfigure}{.99\linewidth}
\begin{subfigure}{.99\linewidth}
\centering
\begin{tikzpicture}[scale=0.5, align=center,
                                text width=0.9cm, inner sep=0mm, node distance=1mm]
\scriptsize 
\matrix[row sep=0.27cm,column sep=0.23cm] {
& & & & & & \node (n1) { \textcolor{blue}{S} }; & & & & & & \\
& & & & & \node (n14) { \textcolor{blue}{NP} }; & & & & & & & \\
& & & & & & & & \node (n2) { \textcolor{blue}{VP} }; & & & & \\
& & & & & & \node (n19) { \textcolor{blue}{NP} }; & & & & & & \\
& & & & & & & \node (n3) { \textcolor{blue}{VP} }; & & & & & \\
& & & & & & & & \node (n26) { \textcolor{blue}{PP} }; & & & & \\
\node (n4) {  }; & \node (n12) {   }; & \node (n15) {   }; & \node (n17) {  }; & \node (n20) {  }; & \node (n22) {  }; & \node (n24) {  }; & \node (n27) {  }; & \node (n29) { }; & \node (n31) {  }; & \node (n6) {  }; & \node (n8) {  }; & \node (n10) {}; \\
\node (n5) { \textcolor{blue}{So} }; & \node (n13) { \textcolor{blue}{sei} }; & \node (n16) { \textcolor{blue}{die} }; & \node[yshift=-2mm] (n18) { \textcolor{blue}{Aufstockung} }; & \node (n21) { \textcolor{blue}{des} }; & \node (n23) { \textcolor{blue}{Aachener} }; & \node[yshift=-2mm] (n25) { \textcolor{blue}{Anteils} }; & \node (n28) { \textcolor{blue}{an} }; & \node (n30) { \textcolor{blue}{der} }; & \node[yshift=-2mm] (n32) { \textcolor{blue}{Volksfürsorge} }; & \node (n7) { \textcolor{blue}{kaum} }; & \node[yshift=-2mm] (n9) { \textcolor{blue}{wahrgenommen} }; & \node (n11) { \textcolor{blue}{worden} }; \\
};
\draw (n31) -- +(0, -0.5) -| (n32);
\draw (n29) -- +(0, -0.5) -| (n30);
\draw (n27) -- +(0, -0.5) -| (n28);
\draw (n24) -- +(0, -0.5) -| (n25);
\draw (n22) -- +(0, -0.5) -| (n23);
\draw (n20) -- +(0, -0.5) -| (n21);
\draw (n17) -- +(0, -0.5) -| (n18);
\draw (n15) -- +(0, -0.5) -| (n16);
\draw (n12) -- +(0, -0.5) -| (n13);
\draw (n10) -- +(0, -0.5) -| (n11);
\draw (n8) -- +(0, -0.5) -| (n9);
\draw (n6) -- +(0, -0.5) -| (n7);
\draw (n4) -- +(0, -0.5) -| (n5);
\draw [white, -, line width=6pt] (n3)  +(0, -0.5) -| (n4);
\draw (n3) -- +(0, -0.5) -| (n4);
\draw [white, -, line width=6pt] (n3)  +(0, -0.5) -| (n6);
\draw (n3) -- +(0, -0.5) -| (n6);
\draw [white, -, line width=6pt] (n3)  +(0, -0.5) -| (n8);
\draw (n3) -- +(0, -0.5) -| (n8);
\draw (n26) -- +(0, -0.5) -| (n27);
\draw (n26) -- +(0, -0.5) -| (n29);
\draw (n26) -- +(0, -0.5) -| (n31);
\draw [white, -, line width=6pt] (n2)  +(0, -0.5) -| (n3);
\draw (n2) -- +(0, -0.5) -| (n3);
\draw [white, -, line width=6pt] (n2)  +(0, -0.5) -| (n10);
\draw (n2) -- +(0, -0.5) -| (n10);
\draw (n19) -- +(0, -0.5) -| (n20);
\draw (n19) -- +(0, -0.5) -| (n22);
\draw (n19) -- +(0, -0.5) -| (n24);
\draw (n19) -- +(0, -0.5) -| (n26);
\draw (n14) -- +(0, -0.5) -| (n15);
\draw (n14) -- +(0, -0.5) -| (n17);
\draw (n14) -- +(0, -0.5) -| (n19);
\draw (n1) -- +(0, -0.5) -| (n2);
\draw (n1) -- +(0, -0.5) -| (n12);
\draw (n1) -- +(0, -0.5) -| (n14);
\end{tikzpicture}    
\end{subfigure}
\begin{subfigure}{.48\linewidth}

\begin{tikzpicture}[scale=0.4, align=center,
                                text width=0.35cm, inner sep=0mm, node distance=1mm]
\tiny
\matrix[row sep=0.15cm,column sep=0.09cm] {
& & & & & & & \node (n0) {}; & & & & & & & & & \\
& & & & & & \node (n15) {  }; & & & & & & & & & & \\
& & & & & & & & \node (n1) {  }; & & & & & & & & \\
& & & & \node (n16) {  }; & & & & & & & & & & & & \\
& & & & & & & & & & & & \node (n2) {  }; & & & & \\
& & & & & \node (n19) {  }; & & & & & & & & & & & \\
& & & & & & \node (n22) {  }; & & & & \node (n30) {  }; & & & & & \node (n5) {  }; & \\
& & & & & & & \node (n25) {  }; & & & & \node (n33) {  }; & & & \node (n6) {  }; & & \\
\node (n3) {  }; & \node (n13) {  }; & \node (n17) {  }; & \node (n20) {  }; & \node (n23) {  }; & & \node (n26) {  }; & & \node (n28) {  }; & \node (n31) {  }; & \node (n34) {  }; & & \node (n36) {  }; & \node (n7) {  }; & & \node (n9) {  }; & \node (n11) {  }; \\
\node (n4) { \textcolor{blue}{So} }; & \node (n14) { \textcolor{blue}{sei} }; & \node (n18) { \textcolor{blue}{die} }; & \node[yshift=-2mm] (n21) { \textcolor{blue}{Aufstockung} }; & \node (n24) { \textcolor{blue}{des} }; & & \node (n27) { \textcolor{blue}{Aachener} }; & & \node[yshift=-2mm] (n29) { \textcolor{blue}{Anteils} }; & \node (n32) { \textcolor{blue}{an} }; & \node (n35) { \textcolor{blue}{der} }; & & \node[yshift=-2mm] (n37) { \textcolor{blue}{Volksfürsorge} }; & \node (n8) { \textcolor{blue}{kaum} }; & & \node[yshift=-2mm] (n10) { \textcolor{blue}{wahrgenommen} }; & \node (n12) { \textcolor{blue}{worden} }; \\
};
\draw (n36) -- +(0, -0.5) -| (n37);
\draw (n34) -- +(0, -0.5) -| (n35);
\draw (n31) -- +(0, -0.5) -| (n32);
\draw (n28) -- +(0, -0.5) -| (n29);
\draw (n26) -- +(0, -0.5) -| (n27);
\draw (n23) -- +(0, -0.5) -| (n24);
\draw (n20) -- +(0, -0.5) -| (n21);
\draw (n17) -- +(0, -0.5) -| (n18);
\draw (n13) -- +(0, -0.5) -| (n14);
\draw (n11) -- +(0, -0.5) -| (n12);
\draw (n9) -- +(0, -0.5) -| (n10);
\draw (n7) -- +(0, -0.5) -| (n8);
\draw (n3) -- +(0, -0.5) -| (n4);
\draw (n33) -- +(0, -0.5) -| (n34);
\draw (n33) -- +(0, -0.5) -| (n36);
\draw (n25) -- +(0, -0.5) -| (n26);
\draw (n25) -- +(0, -0.5) -| (n28);
\draw (n6) -- +(0, -0.5) -| (n7);
\draw (n6) -- +(0, -0.5) -| (n9);
\draw (n30) -- +(0, -0.5) -| (n31);
\draw (n30) -- +(0, -0.5) -| (n33);
\draw (n22) -- +(0, -0.5) -| (n23);
\draw (n22) -- +(0, -0.5) -| (n25);
\draw (n5) -- +(0, -0.5) -| (n6);
\draw (n5) -- +(0, -0.5) -| (n11);
\draw [white, -, line width=6pt] (n2)  +(0, -0.5) -| (n3);
\draw (n2) -- +(0, -0.5) -| (n3);
\draw [white, -, line width=6pt] (n2)  +(0, -0.5) -| (n5);
\draw (n2) -- +(0, -0.5) -| (n5);
\draw (n19) -- +(0, -0.5) -| (n20);
\draw (n19) -- +(0, -0.5) -| (n22);
\draw [white, -, line width=6pt] (n1)  +(0, -0.5) -| (n2);
\draw (n1) -- +(0, -0.5) -| (n2);
\draw [white, -, line width=6pt] (n1)  +(0, -0.5) -| (n13);
\draw (n1) -- +(0, -0.5) -| (n13);
\draw (n16) -- +(0, -0.5) -| (n17);
\draw (n16) -- +(0, -0.5) -| (n19);
\draw (n15) -- +(0, -0.5) -| (n16);
\draw (n15) -- +(0, -0.5) -| (n30);
\draw (n0) -- +(0, -0.5) -| (n1);
\draw (n0) -- +(0, -0.5) -| (n15);
\end{tikzpicture}
\end{subfigure}
\begin{subfigure}{.49\linewidth}
\begin{tikzpicture}[scale=0.4, align=center,
                                text width=0.36cm, inner sep=0mm, node distance=1mm]
\tiny
\matrix[row sep=0.15cm,column sep=0.10cm] {
& & & \node (n0) { }; & & & & & & & & & & & & & \\
& & & & & & & & \node (n3) { }; & & & & & & & & \\
& & & & & & & & & \node (n6) { }; & & & & & & & \\
& & & & & & & \node (n7) { }; & & & & & & & & & \\
& & & & & \node (n8) { }; & & & & & & & & & & & \\
& & & & \node (n9) { }; & & & & & & & & & & & & \\
& & & & & \node (n12) { }; & & & & & \node (n22) { }; & & & & & \node (n30) { }; & \\
& & & & & & \node (n15) { }; & & & & & \node (n25) { }; & & & \node (n31) { }; & & \\
\node (n1) {  }; & \node (n4) {  }; & \node (n10) {  }; & \node (n13) {  }; & & \node (n16) {  }; & & \node (n18) {  }; & \node (n20) {  }; & \node (n23) {  }; & \node (n26) {  }; & & \node (n28) {  }; & \node (n32) {  }; & & \node (n34) {  }; & \node (n36) {  }; \\
\node (n2) { \textcolor{blue}{So} }; & \node (n5) { \textcolor{blue}{sei} }; & \node (n11) { \textcolor{blue}{die} }; & \node[yshift=-2mm] (n14) { \textcolor{blue}{Aufstockung} }; & & \node (n17) { \textcolor{blue}{des} }; & & \node (n19) { \textcolor{blue}{Aachener} }; & \node[yshift=-2mm] (n21) { \textcolor{blue}{Anteils} }; & \node (n24) { \textcolor{blue}{an} }; & \node (n27) { \textcolor{blue}{der} }; & & \node[yshift=-2mm] (n29) { \textcolor{blue}{Volksfürsorge} }; & \node (n33) { \textcolor{blue}{kaum} }; & & \node[yshift=-2mm] (n35) { \textcolor{blue}{wahrgenommen} }; & \node (n37) { \textcolor{blue}{worden} }; \\ 
};
\draw (n36) -- +(0, -0.5) -| (n37);
\draw (n34) -- +(0, -0.5) -| (n35);
\draw (n32) -- +(0, -0.5) -| (n33);
\draw (n28) -- +(0, -0.5) -| (n29);
\draw (n26) -- +(0, -0.5) -| (n27);
\draw (n23) -- +(0, -0.5) -| (n24);
\draw (n20) -- +(0, -0.5) -| (n21);
\draw (n18) -- +(0, -0.5) -| (n19);
\draw (n16) -- +(0, -0.5) -| (n17);
\draw (n13) -- +(0, -0.5) -| (n14);
\draw (n10) -- +(0, -0.5) -| (n11);
\draw (n4) -- +(0, -0.5) -| (n5);
\draw (n1) -- +(0, -0.5) -| (n2);
\draw (n31) -- +(0, -0.5) -| (n32);
\draw (n31) -- +(0, -0.5) -| (n34);
\draw (n25) -- +(0, -0.5) -| (n26);
\draw (n25) -- +(0, -0.5) -| (n28);
\draw (n15) -- +(0, -0.5) -| (n16);
\draw (n15) -- +(0, -0.5) -| (n18);
\draw (n30) -- +(0, -0.5) -| (n31);
\draw (n30) -- +(0, -0.5) -| (n36);
\draw (n22) -- +(0, -0.5) -| (n23);
\draw (n22) -- +(0, -0.5) -| (n25);
\draw (n12) -- +(0, -0.5) -| (n13);
\draw (n12) -- +(0, -0.5) -| (n15);
\draw (n9) -- +(0, -0.5) -| (n10);
\draw (n9) -- +(0, -0.5) -| (n12);
\draw (n8) -- +(0, -0.5) -| (n9);
\draw (n8) -- +(0, -0.5) -| (n20);
\draw (n7) -- +(0, -0.5) -| (n8);
\draw (n7) -- +(0, -0.5) -| (n22);
\draw (n6) -- +(0, -0.5) -| (n7);
\draw (n6) -- +(0, -0.5) -| (n30);
\draw (n3) -- +(0, -0.5) -| (n4);
\draw (n3) -- +(0, -0.5) -| (n6);
\draw (n0) -- +(0, -0.5) -| (n1);
\draw (n0) -- +(0, -0.5) -| (n3);
\end{tikzpicture}
\end{subfigure}
\end{subfigure}
\medskip
\par\noindent\rule{\textwidth}{0.8pt}
\begin{subfigure}{.99\linewidth}
\begin{subfigure}{.99\linewidth}

\centering
\begin{tikzpicture}[scale=0.4, align=center,
                                text width=0.7cm, inner sep=0mm, node distance=1mm]
\scriptsize 
\matrix[row sep=0.23cm,column sep=0.2cm] {
& & & & & & & & \node (n1) { \textcolor{blue}{S} }; & & & & & & & & \\
& & & & & & & & & \node (n2) { \textcolor{blue}{VP} }; & & & & & & & \\
& & & & & & & & \node (n3) { \textcolor{blue}{VP} }; & & & & & & & & \\
& & & & & & & & & & & & \node (n14) { \textcolor{blue}{NP} }; & & & & \\
& \node (n4) { \textcolor{blue}{PP} }; & & & & & & & & & & & & \node (n17) { \textcolor{blue}{S} }; & & & \\
& & \node (n7) { \textcolor{blue}{NM} }; & & & & & \node (n33) { \textcolor{blue}{NP} }; & & & & & & & \node (n20) { \textcolor{blue}{NP} }; & & \\
\node (n5) {  }; & \node (n8) { }; & & \node (n10) {  }; & \node (n12) { }; & \node (n31) {  }; & \node (n34) { }; & & \node (n36) { }; & \node (n15) {   }; & \node (n27) {  }; & \node (n29) {  }; & \node (n18) {  }; & \node (n21) { }; & & \node (n23) {  }; & \node (n25) {}; \\
\node (n6) { \textcolor{blue}{Für} }; & \node (n9) { \textcolor{blue}{15} }; & & \node (n11) { \textcolor{blue}{200} }; & \node (n13) { \textcolor{blue}{Mark} }; & \node (n32) { \textcolor{blue}{will} }; & \node (n35) { \textcolor{blue}{die} }; & & \node[yshift=-2mm] (n37) { \textcolor{blue}{Gemeinde} }; & \node (n16) { \textcolor{blue}{Lampen} }; & \node[yshift=-2mm] (n28) { \textcolor{blue}{einbauen} }; & \node (n30) { \textcolor{blue}{lassen} }; & \node[yshift=-2mm] (n19) { \textcolor{blue}{die} }; & \node (n22) { \textcolor{blue}{mutwilligen} }; & & \node[yshift=-2mm] (n24) { \textcolor{blue}{Zerstörungen} }; & \node (n26) { \textcolor{blue}{standhalten} }; \\
};
\draw (n36) -- +(0, -0.5) -| (n37);
\draw (n34) -- +(0, -0.5) -| (n35);
\draw (n31) -- +(0, -0.5) -| (n32);
\draw (n29) -- +(0, -0.5) -| (n30);
\draw (n27) -- +(0, -0.5) -| (n28);
\draw (n25) -- +(0, -0.5) -| (n26);
\draw (n23) -- +(0, -0.5) -| (n24);
\draw (n21) -- +(0, -0.5) -| (n22);
\draw (n18) -- +(0, -0.5) -| (n19);
\draw (n15) -- +(0, -0.5) -| (n16);
\draw (n12) -- +(0, -0.5) -| (n13);
\draw (n10) -- +(0, -0.5) -| (n11);
\draw (n8) -- +(0, -0.5) -| (n9);
\draw (n5) -- +(0, -0.5) -| (n6);
\draw (n33) -- +(0, -0.5) -| (n34);
\draw (n33) -- +(0, -0.5) -| (n36);
\draw (n20) -- +(0, -0.5) -| (n21);
\draw (n20) -- +(0, -0.5) -| (n23);
\draw (n7) -- +(0, -0.5) -| (n8);
\draw (n7) -- +(0, -0.5) -| (n10);
\draw (n17) -- +(0, -0.5) -| (n18);
\draw (n17) -- +(0, -0.5) -| (n20);
\draw (n17) -- +(0, -0.5) -| (n25);
\draw (n4) -- +(0, -0.5) -| (n5);
\draw (n4) -- +(0, -0.5) -| (n7);
\draw (n4) -- +(0, -0.5) -| (n12);
\draw [white, -, line width=6pt] (n14)  +(0, -0.5) -| (n15);
\draw (n14) -- +(0, -0.5) -| (n15);
\draw [white, -, line width=6pt] (n14)  +(0, -0.5) -| (n17);
\draw (n14) -- +(0, -0.5) -| (n17);
\draw [white, -, line width=6pt] (n3)  +(0, -0.5) -| (n4);
\draw (n3) -- +(0, -0.5) -| (n4);
\draw [white, -, line width=6pt] (n3)  +(0, -0.5) -| (n14);
\draw (n3) -- +(0, -0.5) -| (n14);
\draw [white, -, line width=6pt] (n3)  +(0, -0.5) -| (n27);
\draw (n3) -- +(0, -0.5) -| (n27);
\draw [white, -, line width=6pt] (n2)  +(0, -0.5) -| (n3);
\draw (n2) -- +(0, -0.5) -| (n3);
\draw [white, -, line width=6pt] (n2)  +(0, -0.5) -| (n29);
\draw (n2) -- +(0, -0.5) -| (n29);
\draw (n1) -- +(0, -0.5) -| (n2);
\draw (n1) -- +(0, -0.5) -| (n31);
\draw (n1) -- +(0, -0.5) -| (n33);
\end{tikzpicture}
\end{subfigure}
\begin{subfigure}{.50\linewidth}
\begin{tikzpicture}[scale=0.4, align=center,
                                text width=0.33cm, inner sep=0mm, node distance=1mm]
\tiny
\matrix[row sep=0.18cm,column sep=0.06cm] {
& & & & & & & \node (n0) {  }; & & & & & & & & & & & \\
& & & & & & & \node (n1) { }; & & & & & & & & & & & \\
& & & & & & & & & \node (n2) { }; & & & & & & & & & \\
& & & & & & & & & & & & & & \node (n14) { }; & & & & \\
& & \node (n3) { }; & & & & & & & & & & & & & & \node (n23) { }; & & \\
& & & \node (n6) { }; & & & & & & & & & \node (n15) { }; & & & \node (n24) { }; & & & \\
& & & & \node (n9) { }; & & & & \node (n36) { }; & & & \node (n16) { }; & & & & & \node (n27) { }; & & \\
\node (n4) { }; & \node (n7) { }; & & \node (n10) { }; & & \node (n12) { }; & \node (n34) { }; & \node (n37) { }; & & \node (n39) { }; & \node (n17) { }; & & \node (n19) { }; & \node (n21) { }; & \node (n25) { }; & \node (n28) { }; & & \node (n30) { }; & \node (n32) { }; \\
\node (n5) { \textcolor{blue}{Für} }; & \node (n8) { \textcolor{blue}{15} }; & & \node (n11) { \textcolor{blue}{200} }; & & \node[yshift=-2mm] (n13) { \textcolor{blue}{Mark} }; & \node (n35) { \textcolor{blue}{will} }; & \node (n38) { \textcolor{blue}{die} }; & & \node[yshift=-2mm](n40) { \textcolor{blue}{Gemeinde} }; & \node (n18) { \textcolor{blue}{Lampen} }; & & \node[yshift=-4mm] (n20) { \textcolor{blue}{einbauen} }; & \node (n22) { \textcolor{blue}{lassen} }; & \node[yshift=-2mm] (n26) { \textcolor{blue}{die} }; & \node (n29) { \textcolor{blue}{mutwilligen} }; & & \node[yshift=-2mm](n31) { \textcolor{blue}{Zerstörungen} }; & \node (n33) { \textcolor{blue}{standhalten} }; \\
};
\draw (n39) -- +(0, -0.5) -| (n40);
\draw (n37) -- +(0, -0.5) -| (n38);
\draw (n34) -- +(0, -0.5) -| (n35);
\draw (n32) -- +(0, -0.5) -| (n33);
\draw (n30) -- +(0, -0.5) -| (n31);
\draw (n28) -- +(0, -0.5) -| (n29);
\draw (n25) -- +(0, -0.5) -| (n26);
\draw (n21) -- +(0, -0.5) -| (n22);
\draw (n19) -- +(0, -0.5) -| (n20);
\draw (n17) -- +(0, -0.5) -| (n18);
\draw (n12) -- +(0, -0.5) -| (n13);
\draw (n10) -- +(0, -0.5) -| (n11);
\draw (n7) -- +(0, -0.5) -| (n8);
\draw (n4) -- +(0, -0.5) -| (n5);
\draw (n36) -- +(0, -0.5) -| (n37);
\draw (n36) -- +(0, -0.5) -| (n39);
\draw (n27) -- +(0, -0.5) -| (n28);
\draw (n27) -- +(0, -0.5) -| (n30);
\draw (n16) -- +(0, -0.5) -| (n17);
\draw (n16) -- +(0, -0.5) -| (n19);
\draw (n9) -- +(0, -0.5) -| (n10);
\draw (n9) -- +(0, -0.5) -| (n12);
\draw (n24) -- +(0, -0.5) -| (n25);
\draw (n24) -- +(0, -0.5) -| (n27);
\draw (n15) -- +(0, -0.5) -| (n16);
\draw (n15) -- +(0, -0.5) -| (n21);
\draw (n6) -- +(0, -0.5) -| (n7);
\draw (n6) -- +(0, -0.5) -| (n9);
\draw (n23) -- +(0, -0.5) -| (n24);
\draw (n23) -- +(0, -0.5) -| (n32);
\draw (n3) -- +(0, -0.5) -| (n4);
\draw (n3) -- +(0, -0.5) -| (n6);
\draw (n14) -- +(0, -0.5) -| (n15);
\draw (n14) -- +(0, -0.5) -| (n23);
\draw [white, -, line width=6pt] (n2)  +(0, -0.5) -| (n3);
\draw (n2) -- +(0, -0.5) -| (n3);
\draw [white, -, line width=6pt] (n2)  +(0, -0.5) -| (n14);
\draw (n2) -- +(0, -0.5) -| (n14);
\draw [white, -, line width=6pt] (n1)  +(0, -0.5) -| (n2);
\draw (n1) -- +(0, -0.5) -| (n2);
\draw [white, -, line width=6pt] (n1)  +(0, -0.5) -| (n34);
\draw (n1) -- +(0, -0.5) -| (n34);
\draw (n0) -- +(0, -0.5) -| (n1);
\draw (n0) -- +(0, -0.5) -| (n36);
\end{tikzpicture}
\end{subfigure}
\begin{subfigure}{.49\linewidth}
\begin{tikzpicture}[scale=0.4, align=center,
                                text width=0.32cm, inner sep=0mm, node distance=1mm]
\tiny
\matrix[row sep=0.18cm,column sep=0.06cm] {
& & & & & & & & & \node (n0) { }; & & & & & & & & & \\
& & & & & & & & & & & & \node (n12) { }; & & & & & & \\
& & & & & & & & & & & & & \node (n15) { }; & & & & & \\
& & \node (n1) { }; & & & & & & & & \node (n16) { }; & & & & & & \node (n30) { }; & & \\
& & & \node (n4) { }; & & & & & & & & & \node (n22) { }; & & & \node (n31) { }; & & & \\
& & & & \node (n7) { }; & & & & \node (n17) { }; & & & \node (n23) { }; & & & & & \node (n34) { }; & & \\
\node (n2) { }; & \node (n5) { }; & & \node (n8) { }; & & \node (n10) { }; & \node (n13) { }; & \node (n18) { }; & & \node (n20) { }; & \node (n24) { }; & & \node (n26) { }; & \node (n28) { }; & \node (n32) { }; & \node (n35) { }; & & \node (n37) { }; & \node (n39) { }; \\
\node (n3) { \textcolor{blue}{Für} }; & \node (n6) { \textcolor{blue}{15} }; & & \node (n9) { \textcolor{blue}{200} }; & & \node[yshift=-2mm] (n11) { \textcolor{blue}{Mark} }; & \node (n14) { \textcolor{blue}{will} }; & \node (n19) { \textcolor{blue}{die} }; & & \node[yshift=-2mm] (n21) { \textcolor{blue}{Gemeinde} }; & \node (n25) { \textcolor{blue}{Lampen} }; & & \node[yshift=-4mm] (n27) { \textcolor{blue}{einbauen} }; & \node (n29) { \textcolor{blue}{lassen} }; & \node[yshift=-2mm] (n33) { \textcolor{blue}{die} }; & \node (n36) { \textcolor{blue}{mutwilligen} }; & & \node[yshift=-2mm] (n38) { \textcolor{blue}{Zerstörungen} }; & \node (n40) { \textcolor{blue}{standhalten} }; \\
};
\draw (n39) -- +(0, -0.5) -| (n40);
\draw (n37) -- +(0, -0.5) -| (n38);
\draw (n35) -- +(0, -0.5) -| (n36);
\draw (n32) -- +(0, -0.5) -| (n33);
\draw (n28) -- +(0, -0.5) -| (n29);
\draw (n26) -- +(0, -0.5) -| (n27);
\draw (n24) -- +(0, -0.5) -| (n25);
\draw (n20) -- +(0, -0.5) -| (n21);
\draw (n18) -- +(0, -0.5) -| (n19);
\draw (n13) -- +(0, -0.5) -| (n14);
\draw (n10) -- +(0, -0.5) -| (n11);
\draw (n8) -- +(0, -0.5) -| (n9);
\draw (n5) -- +(0, -0.5) -| (n6);
\draw (n2) -- +(0, -0.5) -| (n3);
\draw (n34) -- +(0, -0.5) -| (n35);
\draw (n34) -- +(0, -0.5) -| (n37);
\draw (n23) -- +(0, -0.5) -| (n24);
\draw (n23) -- +(0, -0.5) -| (n26);
\draw (n17) -- +(0, -0.5) -| (n18);
\draw (n17) -- +(0, -0.5) -| (n20);
\draw (n7) -- +(0, -0.5) -| (n8);
\draw (n7) -- +(0, -0.5) -| (n10);
\draw (n31) -- +(0, -0.5) -| (n32);
\draw (n31) -- +(0, -0.5) -| (n34);
\draw (n22) -- +(0, -0.5) -| (n23);
\draw (n22) -- +(0, -0.5) -| (n28);
\draw (n4) -- +(0, -0.5) -| (n5);
\draw (n4) -- +(0, -0.5) -| (n7);
\draw (n16) -- +(0, -0.5) -| (n17);
\draw (n16) -- +(0, -0.5) -| (n22);
\draw (n30) -- +(0, -0.5) -| (n31);
\draw (n30) -- +(0, -0.5) -| (n39);
\draw (n1) -- +(0, -0.5) -| (n2);
\draw (n1) -- +(0, -0.5) -| (n4);
\draw (n15) -- +(0, -0.5) -| (n16);
\draw (n15) -- +(0, -0.5) -| (n30);
\draw (n12) -- +(0, -0.5) -| (n13);
\draw (n12) -- +(0, -0.5) -| (n15);
\draw (n0) -- +(0, -0.5) -| (n1);
\draw (n0) -- +(0, -0.5) -| (n12);
\end{tikzpicture}

\end{subfigure}
\end{subfigure}

\vspace{-1mm}
\caption{Examples of two sentences from the German NEGRA tree bank. In each example, the gold tree is shown at the top, the TN-LCFRS$_{4500}$ is shown bottom left, and the  TN-PCFG$_{4500}$ tree is shown bottom right.}
\vspace{-2mm}
\label{fig:tree-comparison}
\end{figure*}

%% file: tables/mbr.tex
\begin{table}[tb]
	\centering
	\small
	{\setlength{\tabcolsep}{.0em}
		\begin{tabular}{r}
			\toprule 
			\begin{minipage}{\linewidth}
    \textit{Items}: 
    \renewcommand{\labelenumi}{\Roman{enumi}}
    \begin{enumerate}
          \item $[i, j]$: accumulated scores for continuous spans.
      \item  $[i, j, k, n]$: accumulated scores for discontinuous spans. 
    \end{enumerate}
    \textit{Deductive rules:} 
 \\
 \\
          \phantom{} \infer[\text{$X_{ij}$}\phantom{jj}]{[i, j]}{ [  i, k]& [ k,j]} \\ 
         \phantom{} \infer[\text{$ Y_{ijmn} $}\phantom{jj}]{[i, j, m, n]}{ [ i, j]& m, n]} \\
        \phantom{} \infer[\text{$ X_{ij} $}\phantom{j}]{[ i, j]}{ [  m, n]& [ i, m, n, j]} \\
       \phantom{} \infer[\text{$Y_{ijmn}$}\phantom{j}]{[ i, j, m, n]}{ [  i, k]& [ k, j, n, j]} \\
              \phantom{} \infer[\text{$Y_{ijmn}$}\phantom{j}]{[ i, j, m, n]}{ [  k, j]& [ i, k, m, n]} \\
              \phantom{} \infer[\text{$Y_{ijmn}$}\phantom{j}]{[ i, j, m, n]}{ [  m, k]& [ i, j, k, n]} \\
              \phantom{} \infer[\text{$Y_{ijmn}$} \phantom{j}]{[ i, j, m, n]}{ [  m, k]& [ i, j, k, n]} \\
			\end{minipage}\\
			\bottomrule
	\end{tabular}}
	\caption{CKY-style parsing with span marginals. }
	\label{tab:mbr}
\end{table}

%% file: tables/max_result.tex
\begin{table*}[t!]
    \centering 
    \begin{tabular}{llllllll}
        \toprule 
    {\bf Model} & $|\mathcal{P}|$ 
    & \multicolumn{2}{c}{{\bf NEGRA}} & \multicolumn{2}{c}{{\bf TIGER}}& \multicolumn{2}{c}{\bf LASSY}\\
    &  & F1 &   DF1 & F1  & DF1 & F1 &  DF1 \\
        \midrule 
    N-PCFG & 45 &  41.3   & $-$  & 40.0 & $-$ & 45.5 & $-$ \\ 
    C-PCFG & 45 & 40.2 &  $-$ &  39.8 & $-$ & 40.9 & $-$ \\ 
    N-LCFRS & 45 & 37.0 & 3.4 & 35.6 & 2.0 & 39.4 & 1.7  \\ 
    C-LCFRS & 45 & 38.2 & 4.3 & 36.4 & 3.0 & 42.4 & 3.7  \\
    TN-LCFRS & 45 & 42.5 & 5.5 & 41.3 & 4.4 & 44.4 & 4.6 \\
    TN-LCFRS & 450 & 47.1 & 8.4 & 45.9 & 6.4 & 47.0 & 8.1 \\
    TN-LCFRS & 4500 & \textbf{47.2} &  \textbf{9.7} &  \textbf{46.6} & \textbf{7.3} & 48.0 & \textbf{10.2} \\
    TN-PCFG & 4500 & 46.2 & $-$  & 45.5 & $-$ &  \textbf{50.0} &  $-$ \\
    \midrule 
    Supervised & 4500 & 54.8 & 39.2 & 50.9 & 33.3 & $-$ & $-$ \\
     \bottomrule 
    \end{tabular}
    \vspace{-2mm}
    \caption{Maximum F1 results across four random seeds on the  German (NEGRA, TIGER) and Dutch (LASSY) test sets.}
    \label{tab:max_result}
    \vspace{-4mm}
\end{table*}

%% file: tables/f1_bylength.tex
\begin{table*}[tb!]
    \centering 
    \begin{tabular}{llllllll}
        \toprule 
    {\bf Model} & $|\mathcal{P}|$ 
    & \multicolumn{2}{c}{{\bf TIGER-10}} & \multicolumn{2}{c}{{\bf TIGER-20}}& \multicolumn{2}{c}{\bf TIGER-30}\\
    &  & F1 &   DF1 & F1  & DF1 & F1 &  DF1 \\
        \midrule 
    N-PCFG & 45 & 47.7$_{\pm 0.9}$ & $-$ & 42.5$_{\pm 0.2}$ & $-$ & 40.5$_{\pm 0.2 }$& $-$ \\ 
    C-PCFG & 45 & 48.1$_{\pm 1.1}$ & $-$ & 41.7$_{\pm 1.3}$ & $-$ & 39.7$_{\pm 1.2}$ & $-$ \\ 
    N-LCFRS & 45 & 41.7$_{\pm 2.4}$ & 3.2$_{\pm 1.4}$ & 36.3$_{\pm 2.4}$ & 2.7$_{\pm 1.0}$ & 34.5$_{\pm 2.5}$ & 2.2$_{\pm 0.8}$ \\
    C-LCFRS & 45 & 42.5$_{\pm 1.6}$ & 2.7$_{\pm 1.6}$ & 37.7$_{\pm 1.2}$ & 2.3$_{\pm 1.3}$ & 36.0$_{\pm 1.1}$ & 1.9$_{\pm 1.0}$ \\
    TN-LCFRS & 45 & 48.3 $_{\pm 1.4}$ & 1.9$_{\pm 2.3}$ & 42.8$_{\pm 0.9}$ & 1.6$_{\pm 1.9}$ & 41.0$_{\pm 1.0 }$ & 1.4$_{\pm 1.6}$ \\
    TN-LCFRS & 450 & 51.4$_{\pm 1.8}$ & 6.1$_{\pm 1.7}$ & 46.1$_{\pm 1.7}$ &  5.5$_{\pm 1.9}$ & 44.5$_{\pm 1.7}$ & 4.8$_{\pm 1.8}$  \\
    TN-PCFG & 4500 & 52.4$_{\pm 0.4 }$ & 0.0$_{\pm 0.0 }$ & 47.6$_{\pm 0.5}$ & $-$ & 45.8$_{\pm 0.5}$ & $-$ \\
        TN-LCFRS & 4500 & \textbf{52.9}$_{\pm 1.3}$ & \textbf{8.2}$_{\pm 2.0}$ & \textbf{47.9}$_{\pm 1.1}$ & \textbf{7.4}$_{\pm 1.1}$  & \textbf{46.3}$_{\pm 0.9}$ & \textbf{6.4}$_{\pm 1.0}$ \\
    \midrule
    Oracle bound &  & 64.3 & 88.5 & 65.0 & 86.2 & 73.7 & 68.0\\ 
     \bottomrule 
    \end{tabular}
    \vspace{-2mm}
    \caption{Results on TIGER test set by broken down by sentence length.}
    \label{tab:length_result}
    \vspace{-4mm}
\end{table*}

%% file: figures/appd_example_trees.tex
\begin{figure*}[t!]
\begin{subfigure}{.99\linewidth}
{
\begin{tikzpicture}[scale=0.5, align=center,
                                text width=1cm, inner sep=0mm, node distance=1mm]
\scriptsize
\matrix[row sep=0.3cm,column sep=0.1cm] {
& & & & & & \node (n1) { \textcolor{blue}{S} }; & & & & & & \\
& & & & \node (n2) { \textcolor{blue}{AP} }; & & & & & & & & \\
& & \node (n3) { \textcolor{blue}{S} }; & & & & & & & & & & \\
& & & \node (n4) { \textcolor{blue}{VP} }; & & & & & & & & & \\
& & & & \node (n7) { \textcolor{blue}{PP} }; & & & & & & \node (n26) { \textcolor{blue}{CNP} }; & & \\
\node (n5) { \textcolor{blue}{PROAV} }; & \node (n16) { \textcolor{blue}{VAFIN} }; & \node (n18) { \textcolor{blue}{PIS} }; & \node (n8) { \textcolor{blue}{APPR} }; & \node (n10) { \textcolor{blue}{ART} }; & \node (n12) { \textcolor{blue}{NN} }; & \node (n14) { \textcolor{blue}{VVINF} }; & \node (n24) { \textcolor{blue}{VAFIN} }; & \node (n20) { \textcolor{blue}{PRF} }; & \node (n27) { \textcolor{blue}{NE} }; & \node (n29) { \textcolor{blue}{KON} }; & \node (n31) { \textcolor{blue}{NE} }; & \node (n22) { \textcolor{blue}{ADJD} }; \\
\node (n6) { \textcolor{red}{Dabei} }; & \node (n17) { \textcolor{red}{werden} }; & \node (n19) { \textcolor{red}{einige} }; & \node (n9) { \textcolor{red}{durch} }; & \node (n11) { \textcolor{red}{den} }; & \node (n13) { \textcolor{red}{Rost} }; & \node (n15) { \textcolor{red}{fallen} }; & \node (n25) { \textcolor{red}{sind} }; & \node (n21) { \textcolor{red}{sich} }; & \node (n28) { \textcolor{red}{Minassian} }; & \node (n30) { \textcolor{red}{und} }; & \node (n32) { \textcolor{red}{Richter} }; & \node (n23) { \textcolor{red}{einig} }; \\
};
\draw (n31) -- +(0, -0.5) -| (n32);
\draw (n29) -- +(0, -0.5) -| (n30);
\draw (n27) -- +(0, -0.5) -| (n28);
\draw (n24) -- +(0, -0.5) -| (n25);
\draw (n22) -- +(0, -0.5) -| (n23);
\draw (n20) -- +(0, -0.5) -| (n21);
\draw (n18) -- +(0, -0.5) -| (n19);
\draw (n16) -- +(0, -0.5) -| (n17);
\draw (n14) -- +(0, -0.5) -| (n15);
\draw (n12) -- +(0, -0.5) -| (n13);
\draw (n10) -- +(0, -0.5) -| (n11);
\draw (n8) -- +(0, -0.5) -| (n9);
\draw (n5) -- +(0, -0.5) -| (n6);
\draw (n26) -- +(0, -0.5) -| (n27);
\draw (n26) -- +(0, -0.5) -| (n29);
\draw (n26) -- +(0, -0.5) -| (n31);
\draw (n7) -- +(0, -0.5) -| (n8);
\draw (n7) -- +(0, -0.5) -| (n10);
\draw (n7) -- +(0, -0.5) -| (n12);
\draw [white, -, line width=6pt] (n4)  +(0, -0.5) -| (n5);
\draw (n4) -- +(0, -0.5) -| (n5);
\draw [white, -, line width=6pt] (n4)  +(0, -0.5) -| (n7);
\draw (n4) -- +(0, -0.5) -| (n7);
\draw [white, -, line width=6pt] (n4)  +(0, -0.5) -| (n14);
\draw (n4) -- +(0, -0.5) -| (n14);
\draw (n3) -- +(0, -0.5) -| (n4);
\draw (n3) -- +(0, -0.5) -| (n16);
\draw (n3) -- +(0, -0.5) -| (n18);
\draw [white, -, line width=6pt] (n2)  +(0, -0.5) -| (n3);
\draw (n2) -- +(0, -0.5) -| (n3);
\draw [white, -, line width=6pt] (n2)  +(0, -0.5) -| (n20);
\draw (n2) -- +(0, -0.5) -| (n20);
\draw [white, -, line width=6pt] (n2)  +(0, -0.5) -| (n22);
\draw (n2) -- +(0, -0.5) -| (n22);
\draw (n1) -- +(0, -0.5) -| (n2);
\draw (n1) -- +(0, -0.5) -| (n24);
\draw (n1) -- +(0, -0.5) -| (n26);
\end{tikzpicture}

\begin{tikzpicture}[scale=0.5, align=center,
                                text width=0.7cm, text height=0.15cm, inner sep=0mm, node distance=1mm]
\tiny
\matrix[row sep=0.4cm,column sep=0.1cm] {
& & & & & & & & \node (n0) { \textcolor{blue}{NT} }; & & & & & & & \\
& & & & & & & \node (n1) { \textcolor{blue}{NT} }; & & & & & & & & \\
& & & & & \node (n2) { \textcolor{blue}{NT} }; & & & & & & & & & & \\
& & \node (n3) { \textcolor{blue}{NT} }; & & & & & & & & & & & & & \\
& & & \node (n4) { \textcolor{blue}{NT} }; & & & & & & & & & & & & \\
& & & & \node (n5) { \textcolor{blue}{NT} }; & & & & & & & & & & & \\
& & & & & & \node (n8) { \textcolor{blue}{NT} }; & & & & & & \node (n27) { \textcolor{blue}{NT} }; & & & \\
& & & & & \node (n9) { \textcolor{blue}{NT} }; & & & & & & & & \node (n30) { \textcolor{blue}{NT} }; & & \\
& & & & & & \node (n12) { \textcolor{blue}{NT} }; & & & & & & \node (n31) { \textcolor{blue}{NT} }; & & & \\
\node (n6) { \textcolor{blue}{P} }; & \node (n19) { \textcolor{blue}{P} }; & \node (n21) { \textcolor{blue}{P} }; & & \node (n10) { \textcolor{blue}{P} }; & \node (n13) { \textcolor{blue}{P} }; & & \node (n15) { \textcolor{blue}{P} }; & \node (n17) { \textcolor{blue}{P} }; & \node (n25) { \textcolor{blue}{P} }; & \node (n28) { \textcolor{blue}{P} }; & \node (n32) { \textcolor{blue}{P} }; & & \node (n34) { \textcolor{blue}{P} }; & \node (n36) { \textcolor{blue}{P} }; & \node (n23) { \textcolor{blue}{P} }; \\
\node (n7) { \textcolor{red}{Dabei} }; & \node (n20) { \textcolor{red}{werden} }; & \node (n22) { \textcolor{red}{einige} }; & & \node (n11) { \textcolor{red}{durch} }; & \node (n14) { \textcolor{red}{den} }; & & \node (n16) { \textcolor{red}{Rost} }; & \node (n18) { \textcolor{red}{fallen} }; & \node (n26) { \textcolor{red}{sind} }; & \node (n29) { \textcolor{red}{sich} }; & \node (n33) { \textcolor{red}{Minassian} }; & & \node (n35) { \textcolor{red}{und} }; & \node (n37) { \textcolor{red}{Richter} }; & \node (n24) { \textcolor{red}{einig} }; \\
};
\draw (n36) -- +(0, -0.5) -| (n37);
\draw (n34) -- +(0, -0.5) -| (n35);
\draw (n32) -- +(0, -0.5) -| (n33);
\draw (n28) -- +(0, -0.5) -| (n29);
\draw (n25) -- +(0, -0.5) -| (n26);
\draw (n23) -- +(0, -0.5) -| (n24);
\draw (n21) -- +(0, -0.5) -| (n22);
\draw (n19) -- +(0, -0.5) -| (n20);
\draw (n17) -- +(0, -0.5) -| (n18);
\draw (n15) -- +(0, -0.5) -| (n16);
\draw (n13) -- +(0, -0.5) -| (n14);
\draw (n10) -- +(0, -0.5) -| (n11);
\draw (n6) -- +(0, -0.5) -| (n7);
\draw (n31) -- +(0, -0.5) -| (n32);
\draw (n31) -- +(0, -0.5) -| (n34);
\draw (n12) -- +(0, -0.5) -| (n13);
\draw (n12) -- +(0, -0.5) -| (n15);
\draw (n30) -- +(0, -0.5) -| (n31);
\draw (n30) -- +(0, -0.5) -| (n36);
\draw (n9) -- +(0, -0.5) -| (n10);
\draw (n9) -- +(0, -0.5) -| (n12);
\draw (n27) -- +(0, -0.5) -| (n28);
\draw (n27) -- +(0, -0.5) -| (n30);
\draw (n8) -- +(0, -0.5) -| (n9);
\draw (n8) -- +(0, -0.5) -| (n17);
\draw [white, -, line width=6pt] (n5)  +(0, -0.5) -| (n6);
\draw (n5) -- +(0, -0.5) -| (n6);
\draw [white, -, line width=6pt] (n5)  +(0, -0.5) -| (n8);
\draw (n5) -- +(0, -0.5) -| (n8);
\draw [white, -, line width=6pt] (n4)  +(0, -0.5) -| (n5);
\draw (n4) -- +(0, -0.5) -| (n5);
\draw [white, -, line width=6pt] (n4)  +(0, -0.5) -| (n19);
\draw (n4) -- +(0, -0.5) -| (n19);
\draw (n3) -- +(0, -0.5) -| (n4);
\draw (n3) -- +(0, -0.5) -| (n21);
\draw [white, -, line width=6pt] (n2)  +(0, -0.5) -| (n3);
\draw (n2) -- +(0, -0.5) -| (n3);
\draw [white, -, line width=6pt] (n2)  +(0, -0.5) -| (n23);
\draw (n2) -- +(0, -0.5) -| (n23);
\draw [white, -, line width=6pt] (n1)  +(0, -0.5) -| (n2);
\draw (n1) -- +(0, -0.5) -| (n2);
\draw [white, -, line width=6pt] (n1)  +(0, -0.5) -| (n25);
\draw (n1) -- +(0, -0.5) -| (n25);
\draw (n0) -- +(0, -0.5) -| (n1);
\draw (n0) -- +(0, -0.5) -| (n27);
\end{tikzpicture}
}
\caption{}
\end{subfigure}

\begin{subfigure}{.99\linewidth}
{
\begin{tikzpicture}[scale=0.5, align=center,
                                text width=1cm, inner sep=0mm, node distance=1mm]
\scriptsize
\matrix[row sep=0.3cm,column sep=0.1cm] {
& & & & & & \node (n1) { \textcolor{blue}{S} }; & & & & & & & \\
& & & & & \node (n2) { \textcolor{blue}{VP} }; & & & & & & & & \\
& \node (n3) { \textcolor{blue}{NP} }; & & & & & & & & & & & & \\
& & \node (n6) { \textcolor{blue}{PP} }; & & & & & & & & & & & \\
& & & & \node (n11) { \textcolor{blue}{NP} }; & & & & \node (n25) { \textcolor{blue}{NP} }; & & & \node (n16) { \textcolor{blue}{NP} }; & & \\
\node (n4) { \textcolor{blue}{NN} }; & \node (n7) { \textcolor{blue}{APPRART} }; & \node (n9) { \textcolor{blue}{NN} }; & \node (n12) { \textcolor{blue}{ART} }; & & \node (n14) { \textcolor{blue}{NN} }; & \node (n23) { \textcolor{blue}{VMFIN} }; & \node (n26) { \textcolor{blue}{ART} }; & & \node (n28) { \textcolor{blue}{NN} }; & \node (n17) { \textcolor{blue}{ART} }; & & \node (n19) { \textcolor{blue}{NN} }; & \node (n21) { \textcolor{blue}{VVINF} }; \\
\node (n5) { \textcolor{red}{Informationen} }; & \node (n8) { \textcolor{red}{zur} }; & \node (n10) { \textcolor{red}{Geschichte} }; & \node (n13) { \textcolor{red}{des} }; & & \node (n15) { \textcolor{red}{Schlosses} }; & \node (n24) { \textcolor{red}{können} }; & \node (n27) { \textcolor{red}{die} }; & & \node (n29) { \textcolor{red}{Besucher} }; & \node (n18) { \textcolor{red}{einem} }; & & \node (n20) { \textcolor{red}{Schild} }; & \node (n22) { \textcolor{red}{entnehmen} }; \\
};
\draw (n28) -- +(0, -0.5) -| (n29);
\draw (n26) -- +(0, -0.5) -| (n27);
\draw (n23) -- +(0, -0.5) -| (n24);
\draw (n21) -- +(0, -0.5) -| (n22);
\draw (n19) -- +(0, -0.5) -| (n20);
\draw (n17) -- +(0, -0.5) -| (n18);
\draw (n14) -- +(0, -0.5) -| (n15);
\draw (n12) -- +(0, -0.5) -| (n13);
\draw (n9) -- +(0, -0.5) -| (n10);
\draw (n7) -- +(0, -0.5) -| (n8);
\draw (n4) -- +(0, -0.5) -| (n5);
\draw (n25) -- +(0, -0.5) -| (n26);
\draw (n25) -- +(0, -0.5) -| (n28);
\draw (n16) -- +(0, -0.5) -| (n17);
\draw (n16) -- +(0, -0.5) -| (n19);
\draw (n11) -- +(0, -0.5) -| (n12);
\draw (n11) -- +(0, -0.5) -| (n14);
\draw (n6) -- +(0, -0.5) -| (n7);
\draw (n6) -- +(0, -0.5) -| (n9);
\draw (n6) -- +(0, -0.5) -| (n11);
\draw (n3) -- +(0, -0.5) -| (n4);
\draw (n3) -- +(0, -0.5) -| (n6);
\draw [white, -, line width=6pt] (n2)  +(0, -0.5) -| (n3);
\draw (n2) -- +(0, -0.5) -| (n3);
\draw [white, -, line width=6pt] (n2)  +(0, -0.5) -| (n16);
\draw (n2) -- +(0, -0.5) -| (n16);
\draw [white, -, line width=6pt] (n2)  +(0, -0.5) -| (n21);
\draw (n2) -- +(0, -0.5) -| (n21);
\draw (n1) -- +(0, -0.5) -| (n2);
\draw (n1) -- +(0, -0.5) -| (n23);
\draw (n1) -- +(0, -0.5) -| (n25);
\end{tikzpicture}
\\
\begin{tikzpicture}[scale=0.40, align=center,
                                text width=0.95cm, text height=0.2cm, inner sep=0mm, node distance=1mm]
\tiny
\matrix[row sep=0.3cm,column sep=0.04cm] {
& & & & & & & & \node (n0) { \textcolor{blue}{NT} }; & & & & & & & \\
& & & & & & & \node (n1) { \textcolor{blue}{NT} }; & & & & & & & & \\
& & & & & & \node (n2) { \textcolor{blue}{NT} }; & & & & & & & & & \\
& & \node (n3) { \textcolor{blue}{NT} }; & & & & & & & & & & & & & \\
& & & \node (n6) { \textcolor{blue}{NT} }; & & & & & & & & & & & & \\
& & & & \node (n9) { \textcolor{blue}{NT} }; & & & & & & & & & & \node (n17) { \textcolor{blue}{NT} }; & \\
& & & & & \node (n12) { \textcolor{blue}{NT} }; & & & & & \node (n27) { \textcolor{blue}{NT} }; & & & \node (n18) { \textcolor{blue}{NT} }; & & \\
\node (n4) { \textcolor{blue}{P} }; & \node (n7) { \textcolor{blue}{P} }; & \node (n10) { \textcolor{blue}{P} }; & & \node (n13) { \textcolor{blue}{P} }; & & \node (n15) { \textcolor{blue}{P} }; & & \node (n25) { \textcolor{blue}{P} }; & \node (n28) { \textcolor{blue}{P} }; & & \node (n30) { \textcolor{blue}{P} }; & \node (n19) { \textcolor{blue}{P} }; & & \node (n21) { \textcolor{blue}{P} }; & \node (n23) { \textcolor{blue}{P} }; \\
\node (n5) { \textcolor{red}{Informationen} }; & \node (n8) { \textcolor{red}{zur} }; & \node (n11) { \textcolor{red}{Geschichte} }; & & \node (n14) { \textcolor{red}{des} }; & & \node (n16) { \textcolor{red}{Schlosses} }; & & \node (n26) { \textcolor{red}{können} }; & \node (n29) { \textcolor{red}{die} }; & & \node (n31) { \textcolor{red}{Besucher} }; & \node (n20) { \textcolor{red}{einem} }; & & \node (n22) { \textcolor{red}{Schild} }; & \node (n24) { \textcolor{red}{entnehmen} }; \\
};
\draw (n30) -- +(0, -0.5) -| (n31);
\draw (n28) -- +(0, -0.5) -| (n29);
\draw (n25) -- +(0, -0.5) -| (n26);
\draw (n23) -- +(0, -0.5) -| (n24);
\draw (n21) -- +(0, -0.5) -| (n22);
\draw (n19) -- +(0, -0.5) -| (n20);
\draw (n15) -- +(0, -0.5) -| (n16);
\draw (n13) -- +(0, -0.5) -| (n14);
\draw (n10) -- +(0, -0.5) -| (n11);
\draw (n7) -- +(0, -0.5) -| (n8);
\draw (n4) -- +(0, -0.5) -| (n5);
\draw (n27) -- +(0, -0.5) -| (n28);
\draw (n27) -- +(0, -0.5) -| (n30);
\draw (n18) -- +(0, -0.5) -| (n19);
\draw (n18) -- +(0, -0.5) -| (n21);
\draw (n12) -- +(0, -0.5) -| (n13);
\draw (n12) -- +(0, -0.5) -| (n15);
\draw (n17) -- +(0, -0.5) -| (n18);
\draw (n17) -- +(0, -0.5) -| (n23);
\draw (n9) -- +(0, -0.5) -| (n10);
\draw (n9) -- +(0, -0.5) -| (n12);
\draw (n6) -- +(0, -0.5) -| (n7);
\draw (n6) -- +(0, -0.5) -| (n9);
\draw (n3) -- +(0, -0.5) -| (n4);
\draw (n3) -- +(0, -0.5) -| (n6);
\draw [white, -, line width=6pt] (n2)  +(0, -0.5) -| (n3);
\draw (n2) -- +(0, -0.5) -| (n3);
\draw [white, -, line width=6pt] (n2)  +(0, -0.5) -| (n17);
\draw (n2) -- +(0, -0.5) -| (n17);
\draw [white, -, line width=6pt] (n1)  +(0, -0.5) -| (n2);
\draw (n1) -- +(0, -0.5) -| (n2);
\draw [white, -, line width=6pt] (n1)  +(0, -0.5) -| (n25);
\draw (n1) -- +(0, -0.5) -| (n25);
\draw (n0) -- +(0, -0.5) -| (n1);
\draw (n0) -- +(0, -0.5) -| (n27);
\end{tikzpicture}
}
\caption{}
\end{subfigure}

\begin{subfigure}{.99\linewidth}
{
\begin{tikzpicture}[scale=0.30, align=center,
                                text width=1cm, inner sep=0mm, node distance=1mm]
\tiny
\matrix[row sep=0.3cm,column sep=0.06cm] {
& & & & & & & \node (n1) { \textcolor{blue}{S} }; & & & & & & & \\
& & & & & & & & \node (n2) { \textcolor{blue}{VP} }; & & & & & & \\
& & \node (n3) { \textcolor{blue}{NP} }; & & & & & & & & & & & & \\
& & & & \node (n8) { \textcolor{blue}{VP} }; & & & & & & & & & & \\
& & & \node (n9) { \textcolor{blue}{NP} }; & & & & & & \node (n26) { \textcolor{blue}{NP} }; & & & & & \\
\node (n4) { \textcolor{blue}{ART} }; & \node (n6) { \textcolor{blue}{NN} }; & \node (n10) { \textcolor{blue}{PPOSAT} }; & & \node (n12) { \textcolor{blue}{NN} }; & \node (n14) { \textcolor{blue}{VVIZU} }; & \node (n24) { \textcolor{blue}{VAFIN} }; & & \node (n27) { \textcolor{blue}{ART} }; & & \node (n29) { \textcolor{blue}{NN} }; & \node (n16) { \textcolor{blue}{ADV} }; & \node (n18) { \textcolor{blue}{ADJD} }; & \node (n20) { \textcolor{blue}{PTKNEG} }; & \node (n22) { \textcolor{blue}{VVPP} }; \\
\node (n5) { \textcolor{red}{Dem} }; & \node (n7) { \textcolor{red}{Appell} }; & \node (n11) { \textcolor{red}{ihre} }; & & \node (n13) { \textcolor{red}{Waffen} }; & \node (n15) { \textcolor{red}{abzuliefern} }; & \node (n25) { \textcolor{red}{sind} }; & & \node (n28) { \textcolor{red}{die} }; & & \node (n30) { \textcolor{red}{Aufständischen} }; & \node (n17) { \textcolor{red}{bisher} }; & \node (n19) { \textcolor{red}{offenbar} }; & \node (n21) { \textcolor{red}{nicht} }; & \node (n23) { \textcolor{red}{gefolgt} }; \\
};
\draw (n29) -- +(0, -0.5) -| (n30);
\draw (n27) -- +(0, -0.5) -| (n28);
\draw (n24) -- +(0, -0.5) -| (n25);
\draw (n22) -- +(0, -0.5) -| (n23);
\draw (n20) -- +(0, -0.5) -| (n21);
\draw (n18) -- +(0, -0.5) -| (n19);
\draw (n16) -- +(0, -0.5) -| (n17);
\draw (n14) -- +(0, -0.5) -| (n15);
\draw (n12) -- +(0, -0.5) -| (n13);
\draw (n10) -- +(0, -0.5) -| (n11);
\draw (n6) -- +(0, -0.5) -| (n7);
\draw (n4) -- +(0, -0.5) -| (n5);
\draw (n26) -- +(0, -0.5) -| (n27);
\draw (n26) -- +(0, -0.5) -| (n29);
\draw (n9) -- +(0, -0.5) -| (n10);
\draw (n9) -- +(0, -0.5) -| (n12);
\draw (n8) -- +(0, -0.5) -| (n9);
\draw (n8) -- +(0, -0.5) -| (n14);
\draw (n3) -- +(0, -0.5) -| (n4);
\draw (n3) -- +(0, -0.5) -| (n6);
\draw (n3) -- +(0, -0.5) -| (n8);
\draw [white, -, line width=6pt] (n2)  +(0, -0.5) -| (n3);
\draw (n2) -- +(0, -0.5) -| (n3);
\draw [white, -, line width=6pt] (n2)  +(0, -0.5) -| (n16);
\draw (n2) -- +(0, -0.5) -| (n16);
\draw [white, -, line width=6pt] (n2)  +(0, -0.5) -| (n18);
\draw (n2) -- +(0, -0.5) -| (n18);
\draw [white, -, line width=6pt] (n2)  +(0, -0.5) -| (n20);
\draw (n2) -- +(0, -0.5) -| (n20);
\draw [white, -, line width=6pt] (n2)  +(0, -0.5) -| (n22);
\draw (n2) -- +(0, -0.5) -| (n22);
\draw (n1) -- +(0, -0.5) -| (n2);
\draw (n1) -- +(0, -0.5) -| (n24);
\draw (n1) -- +(0, -0.5) -| (n26);
\end{tikzpicture}
\\
\begin{tikzpicture}[scale=0.3, align=center,
                                text width=.88cm, inner sep=0mm, node distance=1mm]
\tiny
\matrix[row sep=0.33cm,column sep=0.06cm] {
& & & & & & & & & \node (n0) { \textcolor{blue}{NT} }; & & & & & & & \\
& & & & & & & & \node (n1) { \textcolor{blue}{NT} }; & & & & & & & & \\
& & & & & & & & & \node (n2) { \textcolor{blue}{NT} }; & & & & & & & \\
& & & \node (n3) { \textcolor{blue}{NT} }; & & & & & & & & & & & \node (n17) { \textcolor{blue}{NT} }; & & \\
& & & & & \node (n9) { \textcolor{blue}{NT} }; & & & & & & & & \node (n18) { \textcolor{blue}{NT} }; & & & \\
& \node (n4) { \textcolor{blue}{NT} }; & & & \node (n10) { \textcolor{blue}{NT} }; & & & & & & \node (n30) { \textcolor{blue}{NT} }; & & & & \node (n21) { \textcolor{blue}{NT} }; & & \\
\node (n5) { \textcolor{blue}{P} }; & & \node (n7) { \textcolor{blue}{P} }; & \node (n11) { \textcolor{blue}{P} }; & & \node (n13) { \textcolor{blue}{P} }; & \node (n15) { \textcolor{blue}{P} }; & \node (n28) { \textcolor{blue}{P} }; & & \node (n31) { \textcolor{blue}{P} }; & & \node (n33) { \textcolor{blue}{P} }; & \node (n19) { \textcolor{blue}{P} }; & \node (n22) { \textcolor{blue}{P} }; & & \node (n24) { \textcolor{blue}{P} }; & \node (n26) { \textcolor{blue}{P} }; \\
\node (n6) { \textcolor{red}{Dem} }; & & \node (n8) { \textcolor{red}{Appell} }; & \node (n12) { \textcolor{red}{ihre} }; & & \node (n14) { \textcolor{red}{Waffen} }; & \node (n16) { \textcolor{red}{abzuliefern} }; & \node (n29) { \textcolor{red}{sind} }; & & \node (n32) { \textcolor{red}{die} }; & & \node (n34) { \textcolor{red}{Aufständischen} }; & \node (n20) { \textcolor{red}{bisher} }; & \node (n23) { \textcolor{red}{offenbar} }; & & \node (n25) { \textcolor{red}{nicht} }; & \node (n27) { \textcolor{red}{gefolgt} }; \\
};
\draw (n33) -- +(0, -0.5) -| (n34);
\draw (n31) -- +(0, -0.5) -| (n32);
\draw (n28) -- +(0, -0.5) -| (n29);
\draw (n26) -- +(0, -0.5) -| (n27);
\draw (n24) -- +(0, -0.5) -| (n25);
\draw (n22) -- +(0, -0.5) -| (n23);
\draw (n19) -- +(0, -0.5) -| (n20);
\draw (n15) -- +(0, -0.5) -| (n16);
\draw (n13) -- +(0, -0.5) -| (n14);
\draw (n11) -- +(0, -0.5) -| (n12);
\draw (n7) -- +(0, -0.5) -| (n8);
\draw (n5) -- +(0, -0.5) -| (n6);
\draw (n30) -- +(0, -0.5) -| (n31);
\draw (n30) -- +(0, -0.5) -| (n33);
\draw (n21) -- +(0, -0.5) -| (n22);
\draw (n21) -- +(0, -0.5) -| (n24);
\draw (n10) -- +(0, -0.5) -| (n11);
\draw (n10) -- +(0, -0.5) -| (n13);
\draw (n4) -- +(0, -0.5) -| (n5);
\draw (n4) -- +(0, -0.5) -| (n7);
\draw (n18) -- +(0, -0.5) -| (n19);
\draw (n18) -- +(0, -0.5) -| (n21);
\draw (n9) -- +(0, -0.5) -| (n10);
\draw (n9) -- +(0, -0.5) -| (n15);
\draw (n3) -- +(0, -0.5) -| (n4);
\draw (n3) -- +(0, -0.5) -| (n9);
\draw (n17) -- +(0, -0.5) -| (n18);
\draw (n17) -- +(0, -0.5) -| (n26);
\draw [white, -, line width=6pt] (n2)  +(0, -0.5) -| (n3);
\draw (n2) -- +(0, -0.5) -| (n3);
\draw [white, -, line width=6pt] (n2)  +(0, -0.5) -| (n17);
\draw (n2) -- +(0, -0.5) -| (n17);
\draw [white, -, line width=6pt] (n1)  +(0, -0.5) -| (n2);
\draw (n1) -- +(0, -0.5) -| (n2);
\draw [white, -, line width=6pt] (n1)  +(0, -0.5) -| (n28);
\draw (n1) -- +(0, -0.5) -| (n28);
\draw (n0) -- +(0, -0.5) -| (n1);
\draw (n0) -- +(0, -0.5) -| (n30);
\end{tikzpicture}
}
\caption{}
\end{subfigure}

\caption{Examples of gold (top) and predicted (bottom) trees in Germain. NT and P denote  predicted nonterminals and preterminals.}
\label{fig:appd-example-german}
\end{figure*}

%% file: figures/appd_tree_dutch.tex
\begin{figure*}[t!]
\begin{subfigure}{.9\linewidth}{
\begin{tikzpicture}[scale=0.4, align=center,
                                text width=0.57cm, inner sep=0mm, node distance=1mm]
\tiny
\matrix[row sep=0.2cm,column sep=0.05cm] {
& & & & & & & & & & & \node (n1) { \textcolor{blue}{SMAIN} }; & & & & & & & & & & & & & \\
& & & & & & & & & & & & \node (n20) { \textcolor{blue}{WHSUB} }; & & & & & & & & & & & & \\
& & & & & & & & & & & & & & & & \node (n23) { \textcolor{blue}{CONJ} }; & & & & & & & & \\
& & & & & & & & & & & & & \node (n24) { \textcolor{blue}{SSUB} }; & & & & & & & & & & & \\
& & & & & & \node (n9) { \textcolor{blue}{PP} }; & & & & & & & \node (n27) { \textcolor{blue}{INF} }; & & & & & & & & & \node (n44) { \textcolor{blue}{SSUB} }; & & \\
& & & & & & & \node (n12) { \textcolor{blue}{NP} }; & & & & & \node (n28) { \textcolor{blue}{NP} }; & & & & & & & & & \node (n45) { \textcolor{blue}{INF} }; & & & \\
& \node (n2) { \textcolor{blue}{NP} }; & & & & & \node (n13) { \textcolor{blue}{NP} }; & & & & & & & & \node (n33) { \textcolor{blue}{PP} }; & & & & & & \node (n46) { \textcolor{blue}{NP} }; & & & & \\
\node (n3) { \textcolor{blue}{n} }; & & \node (n5) { \textcolor{blue}{n} }; & \node (n7) { \textcolor{blue}{ww} }; & \node (n10) { \textcolor{blue}{vz} }; & \node (n14) { \textcolor{blue}{lid} }; & & \node (n16) { \textcolor{blue}{n} }; & \node (n18) { \textcolor{blue}{n} }; & \node (n21) { \textcolor{blue}{bw} }; & \node (n25) { \textcolor{blue}{vnw} }; & \node (n29) { \textcolor{blue}{ww} }; & \node (n31) { \textcolor{blue}{n} }; & \node (n34) { \textcolor{blue}{vz} }; & & \node (n36) { \textcolor{blue}{n} }; & \node (n40) { \textcolor{blue}{ww} }; & \node (n38) { \textcolor{blue}{ww} }; & \node (n42) { \textcolor{blue}{vg} }; & \node (n47) { \textcolor{blue}{adj} }; & & \node (n49) { \textcolor{blue}{n} }; & & \node (n53) { \textcolor{blue}{ww} }; & \node (n51) { \textcolor{blue}{ww} }; \\
\node (n4) { \textcolor{red}{Yens} }; & & \node (n6) { \textcolor{red}{moeder} }; & \node (n8) { \textcolor{red}{leert} }; & \node (n11) { \textcolor{red}{in} }; & \node (n15) { \textcolor{red}{een} }; & & \node (n17) { \textcolor{red}{paar} }; & \node (n19) { \textcolor{red}{gesprekken} }; & \node[yshift=-2mm] (n22) { \textcolor{red}{hoe} }; & \node (n26) { \textcolor{red}{ze} }; & \node (n30) { \textcolor{red}{gewenst} }; & \node[yshift=-2mm] (n32) { \textcolor{red}{gedrag} }; & \node (n35) { \textcolor{red}{van} }; & & \node (n37) { \textcolor{red}{Yen} }; & \node (n41) { \textcolor{red}{kan} }; & \node (n39) { \textcolor{red}{belonen} }; & \node (n43) { \textcolor{red}{en} }; & \node (n48) { \textcolor{red}{ongewenst} }; & & \node (n50) { \textcolor{red}{gedrag} }; & & \node (n54) { \textcolor{red}{kan} }; & \node (n52) { \textcolor{red}{bijsturen} }; \\
};
\draw (n53) -- +(0, -0.5) -| (n54);
\draw (n51) -- +(0, -0.5) -| (n52);
\draw (n49) -- +(0, -0.5) -| (n50);
\draw (n47) -- +(0, -0.5) -| (n48);
\draw (n42) -- +(0, -0.5) -| (n43);
\draw (n40) -- +(0, -0.5) -| (n41);
\draw (n38) -- +(0, -0.5) -| (n39);
\draw (n36) -- +(0, -0.5) -| (n37);
\draw (n34) -- +(0, -0.5) -| (n35);
\draw (n31) -- +(0, -0.5) -| (n32);
\draw (n29) -- +(0, -0.5) -| (n30);
\draw (n25) -- +(0, -0.5) -| (n26);
\draw (n21) -- +(0, -0.5) -| (n22);
\draw (n18) -- +(0, -0.5) -| (n19);
\draw (n16) -- +(0, -0.5) -| (n17);
\draw (n14) -- +(0, -0.5) -| (n15);
\draw (n10) -- +(0, -0.5) -| (n11);
\draw (n7) -- +(0, -0.5) -| (n8);
\draw (n5) -- +(0, -0.5) -| (n6);
\draw (n3) -- +(0, -0.5) -| (n4);
\draw (n46) -- +(0, -0.5) -| (n47);
\draw (n46) -- +(0, -0.5) -| (n49);
\draw (n33) -- +(0, -0.5) -| (n34);
\draw (n33) -- +(0, -0.5) -| (n36);
\draw (n13) -- +(0, -0.5) -| (n14);
\draw (n13) -- +(0, -0.5) -| (n16);
\draw (n2) -- +(0, -0.5) -| (n3);
\draw (n2) -- +(0, -0.5) -| (n5);
\draw [white, -, line width=6pt] (n45)  +(0, -0.5) -| (n46);
\draw (n45) -- +(0, -0.5) -| (n46);
\draw [white, -, line width=6pt] (n45)  +(0, -0.5) -| (n51);
\draw (n45) -- +(0, -0.5) -| (n51);
\draw (n28) -- +(0, -0.5) -| (n29);
\draw (n28) -- +(0, -0.5) -| (n31);
\draw (n28) -- +(0, -0.5) -| (n33);
\draw (n12) -- +(0, -0.5) -| (n13);
\draw (n12) -- +(0, -0.5) -| (n18);
\draw [white, -, line width=6pt] (n27)  +(0, -0.5) -| (n28);
\draw (n27) -- +(0, -0.5) -| (n28);
\draw [white, -, line width=6pt] (n27)  +(0, -0.5) -| (n38);
\draw (n27) -- +(0, -0.5) -| (n38);
\draw (n44) -- +(0, -0.5) -| (n45);
\draw (n44) -- +(0, -0.5) -| (n53);
\draw (n9) -- +(0, -0.5) -| (n10);
\draw (n9) -- +(0, -0.5) -| (n12);
\draw (n24) -- +(0, -0.5) -| (n25);
\draw (n24) -- +(0, -0.5) -| (n27);
\draw (n24) -- +(0, -0.5) -| (n40);
\draw (n23) -- +(0, -0.5) -| (n24);
\draw (n23) -- +(0, -0.5) -| (n42);
\draw (n23) -- +(0, -0.5) -| (n44);
\draw (n20) -- +(0, -0.5) -| (n21);
\draw (n20) -- +(0, -0.5) -| (n23);
\draw (n1) -- +(0, -0.5) -| (n2);
\draw (n1) -- +(0, -0.5) -| (n7);
\draw (n1) -- +(0, -0.5) -| (n9);
\draw (n1) -- +(0, -0.5) -| (n20);
\end{tikzpicture}

\begin{tikzpicture}[scale=0.4, align=center,
                                text width=0.5cm, inner sep=0mm, node distance=0.05mm]
\tiny
\matrix[row sep=0.2cm,column sep=0.05cm] {
& & & & & & & & & & & & \node (n0) { \textcolor{blue}{NT} }; & & & & & & & & & & & & & & & \\
& & & & & & & & & & & & & & & & \node (n9) { \textcolor{blue}{NT} }; & & & & & & & & & & & \\
& & & & & & & & & & & & & & & & & & & \node (n21) { \textcolor{blue}{NT} }; & & & & & & & & \\
& & & & & & & & & & & & & & \node (n22) { \textcolor{blue}{NT} }; & & & & & & & & & & & & & \\
& & & & & & & & & & & & & & & & & \node (n28) { \textcolor{blue}{NT} }; & & & & & & & \node (n45) { \textcolor{blue}{NT} }; & & & \\
& & & & & & \node (n10) { \textcolor{blue}{NT} }; & & & & & & & & & & \node (n29) { \textcolor{blue}{NT} }; & & & & & & & & & \node (n48) { \textcolor{blue}{NT} }; & & \\
& & \node (n1) { \textcolor{blue}{NT} }; & & & & & \node (n13) { \textcolor{blue}{NT} }; & & & & & & & & \node (n30) { \textcolor{blue}{NT} }; & & & & & & & & & \node (n49) { \textcolor{blue}{NT} }; & & & \\
& \node (n2) { \textcolor{blue}{NT} }; & & & & & \node (n14) { \textcolor{blue}{NT} }; & & & & \node (n23) { \textcolor{blue}{NT} }; & & & \node (n31) { \textcolor{blue}{NT} }; & & & & \node (n36) { \textcolor{blue}{NT} }; & & & & & & \node (n50) { \textcolor{blue}{NT} }; & & & & \\
\node (n3) { \textcolor{blue}{P} }; & & \node (n5) { \textcolor{blue}{P} }; & \node (n7) { \textcolor{blue}{P} }; & \node (n11) { \textcolor{blue}{P} }; & \node (n15) { \textcolor{blue}{P} }; & & \node (n17) { \textcolor{blue}{P} }; & \node (n19) { \textcolor{blue}{P} }; & \node (n24) { \textcolor{blue}{P} }; & & \node (n26) { \textcolor{blue}{P} }; & \node (n32) { \textcolor{blue}{P} }; & & \node (n34) { \textcolor{blue}{P} }; & & \node (n37) { \textcolor{blue}{P} }; & & \node (n39) { \textcolor{blue}{P} }; & \node (n43) { \textcolor{blue}{P} }; & \node (n41) { \textcolor{blue}{P} }; & \node (n46) { \textcolor{blue}{P} }; & \node (n51) { \textcolor{blue}{P} }; & & \node (n53) { \textcolor{blue}{P} }; & & \node (n57) { \textcolor{blue}{P} }; & \node (n55) { \textcolor{blue}{P} }; \\
\node (n4) { \textcolor{red}{Yens} }; & & \node (n6) { \textcolor{red}{moeder} }; & \node (n8) { \textcolor{red}{leert} }; & \node (n12) { \textcolor{red}{in} }; & \node (n16) { \textcolor{red}{een} }; & & \node (n18) { \textcolor{red}{paar} }; & \node (n20) { \textcolor{red}{gesprekken} }; & \node[yshift=-2mm] (n25) { \textcolor{red}{hoe} }; & & \node (n27) { \textcolor{red}{ze} }; & \node (n33) { \textcolor{red}{gewenst} }; & & \node (n35) { \textcolor{red}{gedrag} }; & & \node (n38) { \textcolor{red}{van} }; & & \node (n40) { \textcolor{red}{Yen} }; & \node (n44) { \textcolor{red}{kan} }; & \node (n42) { \textcolor{red}{belonen} }; & \node (n47) { \textcolor{red}{en} }; & \node (n52) { \textcolor{red}{ongewenst} }; & & \node (n54) { \textcolor{red}{gedrag} }; & & \node (n58) { \textcolor{red}{kan} }; & \node (n56) { \textcolor{red}{bijsturen} }; \\
};
\draw (n57) -- +(0, -0.5) -| (n58);
\draw (n55) -- +(0, -0.5) -| (n56);
\draw (n53) -- +(0, -0.5) -| (n54);
\draw (n51) -- +(0, -0.5) -| (n52);
\draw (n46) -- +(0, -0.5) -| (n47);
\draw (n43) -- +(0, -0.5) -| (n44);
\draw (n41) -- +(0, -0.5) -| (n42);
\draw (n39) -- +(0, -0.5) -| (n40);
\draw (n37) -- +(0, -0.5) -| (n38);
\draw (n34) -- +(0, -0.5) -| (n35);
\draw (n32) -- +(0, -0.5) -| (n33);
\draw (n26) -- +(0, -0.5) -| (n27);
\draw (n24) -- +(0, -0.5) -| (n25);
\draw (n19) -- +(0, -0.5) -| (n20);
\draw (n17) -- +(0, -0.5) -| (n18);
\draw (n15) -- +(0, -0.5) -| (n16);
\draw (n11) -- +(0, -0.5) -| (n12);
\draw (n7) -- +(0, -0.5) -| (n8);
\draw (n5) -- +(0, -0.5) -| (n6);
\draw (n3) -- +(0, -0.5) -| (n4);
\draw (n50) -- +(0, -0.5) -| (n51);
\draw (n50) -- +(0, -0.5) -| (n53);
\draw (n36) -- +(0, -0.5) -| (n37);
\draw (n36) -- +(0, -0.5) -| (n39);
\draw (n31) -- +(0, -0.5) -| (n32);
\draw (n31) -- +(0, -0.5) -| (n34);
\draw (n23) -- +(0, -0.5) -| (n24);
\draw (n23) -- +(0, -0.5) -| (n26);
\draw (n14) -- +(0, -0.5) -| (n15);
\draw (n14) -- +(0, -0.5) -| (n17);
\draw (n2) -- +(0, -0.5) -| (n3);
\draw (n2) -- +(0, -0.5) -| (n5);
\draw [white, -, line width=6pt] (n49)  +(0, -0.5) -| (n50);
\draw (n49) -- +(0, -0.5) -| (n50);
\draw [white, -, line width=6pt] (n49)  +(0, -0.5) -| (n55);
\draw (n49) -- +(0, -0.5) -| (n55);
\draw (n30) -- +(0, -0.5) -| (n31);
\draw (n30) -- +(0, -0.5) -| (n36);
\draw (n13) -- +(0, -0.5) -| (n14);
\draw (n13) -- +(0, -0.5) -| (n19);
\draw (n1) -- +(0, -0.5) -| (n2);
\draw (n1) -- +(0, -0.5) -| (n7);
\draw [white, -, line width=6pt] (n29)  +(0, -0.5) -| (n30);
\draw (n29) -- +(0, -0.5) -| (n30);
\draw [white, -, line width=6pt] (n29)  +(0, -0.5) -| (n41);
\draw (n29) -- +(0, -0.5) -| (n41);
\draw (n48) -- +(0, -0.5) -| (n49);
\draw (n48) -- +(0, -0.5) -| (n57);
\draw (n10) -- +(0, -0.5) -| (n11);
\draw (n10) -- +(0, -0.5) -| (n13);
\draw (n28) -- +(0, -0.5) -| (n29);
\draw (n28) -- +(0, -0.5) -| (n43);
\draw (n45) -- +(0, -0.5) -| (n46);
\draw (n45) -- +(0, -0.5) -| (n48);
\draw (n22) -- +(0, -0.5) -| (n23);
\draw (n22) -- +(0, -0.5) -| (n28);
\draw (n21) -- +(0, -0.5) -| (n22);
\draw (n21) -- +(0, -0.5) -| (n45);
\draw (n9) -- +(0, -0.5) -| (n10);
\draw (n9) -- +(0, -0.5) -| (n21);
\draw (n0) -- +(0, -0.5) -| (n1);
\draw (n0) -- +(0, -0.5) -| (n9);
\end{tikzpicture}
}
\caption{}
\end{subfigure}

\begin{subfigure}{.99\linewidth}
{
\begin{tikzpicture}[scale=0.4, align=center,
                                text width=0.5cm, inner sep=0mm, node distance=1mm]
\tiny
\matrix[row sep=0.2cm,column sep=0.05cm] {
& & & & & & & & & & & & & \node (n1) { \textcolor{blue}{SMAIN} }; & & & & & & & & & & & & & \\
& & & & & & & & & & & & & & & & & & & \node (n30) { \textcolor{blue}{CONJ} }; & & & & & & & \\
& & \node (n2) { \textcolor{blue}{PP} }; & & & & & & & & & & & & & & & & & & & & \node (n46) { \textcolor{blue}{WHSUB} }; & & & & \\
& & & \node (n5) { \textcolor{blue}{NP} }; & & & & & & & & & & & \node (n31) { \textcolor{blue}{WHSUB} }; & & & & & & & & & & \node (n49) { \textcolor{blue}{SSUB} }; & & \\
& & & & \node (n10) { \textcolor{blue}{PP} }; & & & & & \node (n20) { \textcolor{blue}{NP} }; & & & & & & \node (n34) { \textcolor{blue}{SSUB} }; & & & & & & & & & \node (n52) { \textcolor{blue}{INF} }; & & \\
& & & & & \node (n13) { \textcolor{blue}{NP} }; & & & & & & \node (n25) { \textcolor{blue}{PP} }; & & & & & & \node (n37) { \textcolor{blue}{INF} }; & & & & & & \node (n53) { \textcolor{blue}{NP} }; & & & \\
\node (n3) { \textcolor{blue}{vz} }; & \node (n6) { \textcolor{blue}{lid} }; & \node (n8) { \textcolor{blue}{n} }; & \node (n11) { \textcolor{blue}{vz} }; & \node (n14) { \textcolor{blue}{lid} }; & & \node (n16) { \textcolor{blue}{n} }; & \node (n18) { \textcolor{blue}{ww} }; & \node (n21) { \textcolor{blue}{lid} }; & \node (n23) { \textcolor{blue}{n} }; & \node (n26) { \textcolor{blue}{vz} }; & & \node (n28) { \textcolor{blue}{n} }; & \node (n32) { \textcolor{blue}{bw} }; & \node (n35) { \textcolor{blue}{vnw} }; & \node (n38) { \textcolor{blue}{n} }; & \node (n42) { \textcolor{blue}{ww} }; & & \node (n40) { \textcolor{blue}{ww} }; & \node (n44) { \textcolor{blue}{vg} }; & \node (n47) { \textcolor{blue}{bw} }; & \node (n50) { \textcolor{blue}{vnw} }; & \node (n54) { \textcolor{blue}{adj} }; & & \node (n56) { \textcolor{blue}{n} }; & \node (n60) { \textcolor{blue}{ww} }; & \node (n58) { \textcolor{blue}{ww} }; \\
\node (n4) { \textcolor{red}{Dankzij} }; & \node (n7) { \textcolor{red}{de} }; & \node (n9) { \textcolor{red}{adviezen} }; & \node (n12) { \textcolor{red}{van} }; & \node (n15) { \textcolor{red}{de} }; & & \node (n17) { \textcolor{red}{Vroeghulp} }; & \node[yshift=-2mm] (n19) { \textcolor{red}{weten} }; & \node (n22) { \textcolor{red}{de} }; & \node (n24) { \textcolor{red}{ouders} }; & \node (n27) { \textcolor{red}{van} }; & & \node (n29) { \textcolor{red}{Harm} }; & \node (n33) { \textcolor{red}{hoe} }; & \node (n36) { \textcolor{red}{ze} }; & \node (n39) { \textcolor{red}{Harm} }; & \node (n43) { \textcolor{red}{kunnen} }; & & \node[yshift=-2mm] (n41) { \textcolor{red}{begeleiden} }; & \node (n45) { \textcolor{red}{en} }; & \node (n48) { \textcolor{red}{waarvoor} }; & \node (n51) { \textcolor{red}{ze} }; & \node (n55) { \textcolor{red}{diverse} }; & & \node (n57) { \textcolor{red}{hulpverleners} }; & \node[yshift=-2mm] (n61) { \textcolor{red}{kunnen} }; & \node (n59) { \textcolor{red}{inschakelen} }; \\
};
\draw (n60) -- +(0, -0.5) -| (n61);
\draw (n58) -- +(0, -0.5) -| (n59);
\draw (n56) -- +(0, -0.5) -| (n57);
\draw (n54) -- +(0, -0.5) -| (n55);
\draw (n50) -- +(0, -0.5) -| (n51);
\draw (n47) -- +(0, -0.5) -| (n48);
\draw (n44) -- +(0, -0.5) -| (n45);
\draw (n42) -- +(0, -0.5) -| (n43);
\draw (n40) -- +(0, -0.5) -| (n41);
\draw (n38) -- +(0, -0.5) -| (n39);
\draw (n35) -- +(0, -0.5) -| (n36);
\draw (n32) -- +(0, -0.5) -| (n33);
\draw (n28) -- +(0, -0.5) -| (n29);
\draw (n26) -- +(0, -0.5) -| (n27);
\draw (n23) -- +(0, -0.5) -| (n24);
\draw (n21) -- +(0, -0.5) -| (n22);
\draw (n18) -- +(0, -0.5) -| (n19);
\draw (n16) -- +(0, -0.5) -| (n17);
\draw (n14) -- +(0, -0.5) -| (n15);
\draw (n11) -- +(0, -0.5) -| (n12);
\draw (n8) -- +(0, -0.5) -| (n9);
\draw (n6) -- +(0, -0.5) -| (n7);
\draw (n3) -- +(0, -0.5) -| (n4);
\draw [white, -, line width=6pt] (n37)  +(0, -0.5) -| (n38);
\draw (n37) -- +(0, -0.5) -| (n38);
\draw [white, -, line width=6pt] (n37)  +(0, -0.5) -| (n40);
\draw (n37) -- +(0, -0.5) -| (n40);
\draw (n53) -- +(0, -0.5) -| (n54);
\draw (n53) -- +(0, -0.5) -| (n56);
\draw (n25) -- +(0, -0.5) -| (n26);
\draw (n25) -- +(0, -0.5) -| (n28);
\draw (n13) -- +(0, -0.5) -| (n14);
\draw (n13) -- +(0, -0.5) -| (n16);
\draw [white, -, line width=6pt] (n52)  +(0, -0.5) -| (n53);
\draw (n52) -- +(0, -0.5) -| (n53);
\draw [white, -, line width=6pt] (n52)  +(0, -0.5) -| (n58);
\draw (n52) -- +(0, -0.5) -| (n58);
\draw (n34) -- +(0, -0.5) -| (n35);
\draw (n34) -- +(0, -0.5) -| (n37);
\draw (n34) -- +(0, -0.5) -| (n42);
\draw (n20) -- +(0, -0.5) -| (n21);
\draw (n20) -- +(0, -0.5) -| (n23);
\draw (n20) -- +(0, -0.5) -| (n25);
\draw (n10) -- +(0, -0.5) -| (n11);
\draw (n10) -- +(0, -0.5) -| (n13);
\draw (n49) -- +(0, -0.5) -| (n50);
\draw (n49) -- +(0, -0.5) -| (n52);
\draw (n49) -- +(0, -0.5) -| (n60);
\draw (n31) -- +(0, -0.5) -| (n32);
\draw (n31) -- +(0, -0.5) -| (n34);
\draw (n5) -- +(0, -0.5) -| (n6);
\draw (n5) -- +(0, -0.5) -| (n8);
\draw (n5) -- +(0, -0.5) -| (n10);
\draw (n46) -- +(0, -0.5) -| (n47);
\draw (n46) -- +(0, -0.5) -| (n49);
\draw (n2) -- +(0, -0.5) -| (n3);
\draw (n2) -- +(0, -0.5) -| (n5);
\draw (n30) -- +(0, -0.5) -| (n31);
\draw (n30) -- +(0, -0.5) -| (n44);
\draw (n30) -- +(0, -0.5) -| (n46);
\draw (n1) -- +(0, -0.5) -| (n2);
\draw (n1) -- +(0, -0.5) -| (n18);
\draw (n1) -- +(0, -0.5) -| (n20);
\draw (n1) -- +(0, -0.5) -| (n30);
\end{tikzpicture}

\begin{tikzpicture}[scale=0.38, align=center,
                                text width=0.4cm, inner sep=0mm, node distance=1mm]
\tiny
\matrix[row sep=0.17cm,column sep=0.05cm] {
& & & & & & & & & & & & & & & & \node (n0) { \textcolor{blue}{NT} }; & & & & & & & & & & & & & & \\
& & & & & & & & & & & \node (n1) { \textcolor{blue}{NT} }; & & & & & & & & & & & & & & & & & & & \\
& & & & & & & & & \node (n2) { \textcolor{blue}{NT} }; & & & & & & & & & & & & & & & & & & & & & \\
& & & & & & & \node (n3) { \textcolor{blue}{NT} }; & & & & & & & & & & & & & & & & & & & & & & & \\
& & & & \node (n4) { \textcolor{blue}{NT} }; & & & & & & & & & & & & & & & & & & & & & & \node (n51) { \textcolor{blue}{NT} }; & & & & \\
& & & \node (n5) { \textcolor{blue}{NT} }; & & & & & & & & & & & & & & & \node (n35) { \textcolor{blue}{NT} }; & & & & & & & & & & \node (n54) { \textcolor{blue}{NT} }; & & \\
& & & & \node (n8) { \textcolor{blue}{NT} }; & & & & & & & & & & & & & & & \node (n38) { \textcolor{blue}{NT} }; & & & & & & & & \node (n55) { \textcolor{blue}{NT} }; & & & \\
& & & & & \node (n14) { \textcolor{blue}{NT} }; & & & & & & & \node (n24) { \textcolor{blue}{NT} }; & & & & & & \node (n39) { \textcolor{blue}{NT} }; & & & & & & & & & & \node (n58) { \textcolor{blue}{NT} }; & & \\
& & \node (n9) { \textcolor{blue}{NT} }; & & & & \node (n17) { \textcolor{blue}{NT} }; & & & & \node (n25) { \textcolor{blue}{NT} }; & & & & \node (n30) { \textcolor{blue}{NT} }; & & & & & & & \node (n42) { \textcolor{blue}{NT} }; & & & & & & \node (n59) { \textcolor{blue}{NT} }; & & & \\
\node (n6) { \textcolor{blue}{P} }; & \node (n10) { \textcolor{blue}{P} }; & & \node (n12) { \textcolor{blue}{P} }; & \node (n15) { \textcolor{blue}{P} }; & \node (n18) { \textcolor{blue}{P} }; & & \node (n20) { \textcolor{blue}{P} }; & \node (n22) { \textcolor{blue}{P} }; & \node (n26) { \textcolor{blue}{P} }; & & \node (n28) { \textcolor{blue}{P} }; & & \node (n31) { \textcolor{blue}{P} }; & & \node (n33) { \textcolor{blue}{P} }; & \node (n36) { \textcolor{blue}{P} }; & \node (n40) { \textcolor{blue}{P} }; & \node (n43) { \textcolor{blue}{P} }; & & \node (n47) { \textcolor{blue}{P} }; & & \node (n45) { \textcolor{blue}{P} }; & \node (n49) { \textcolor{blue}{P} }; & \node (n52) { \textcolor{blue}{P} }; & \node (n56) { \textcolor{blue}{P} }; & \node (n60) { \textcolor{blue}{P} }; & & \node (n62) { \textcolor{blue}{P} }; & \node (n66) { \textcolor{blue}{P} }; & \node (n64) { \textcolor{blue}{P} }; \\
\node[yshift=-2mm] (n7) { \textcolor{red}{Dankzij} }; & \node (n11) { \textcolor{red}{de} }; & & \node[yshift=-2mm] (n13) { \textcolor{red}{adviezen} }; & \node (n16) { \textcolor{red}{van} }; & \node (n19) { \textcolor{red}{de} }; & & \node[yshift=-2mm] (n21) { \textcolor{red}{Vroeghulp} }; & \node (n23) { \textcolor{red}{weten} }; & \node (n27) { \textcolor{red}{de} }; & & \node (n29) { \textcolor{red}{ouders} }; & & \node (n32) { \textcolor{red}{van} }; & & \node (n34) { \textcolor{red}{Harm} }; & \node (n37) { \textcolor{red}{hoe} }; & \node (n41) { \textcolor{red}{ze} }; & \node (n44) { \textcolor{red}{Harm} }; & & \node (n48) { \textcolor{red}{kunnen} }; & & \node[yshift=-2mm] (n46) { \textcolor{red}{begeleiden} }; & \node (n50) { \textcolor{red}{en} }; & \node[yshift=-2mm] (n53) { \textcolor{red}{waarvoor} }; & \node (n57) { \textcolor{red}{ze} }; & \node (n61) { \textcolor{red}{diverse} }; & & \node[yshift=-2mm] (n63) { \textcolor{red}{hulpverleners} }; & \node (n67) { \textcolor{red}{kunnen} }; & \node[yshift=-4mm] (n65) { \textcolor{red}{inschakelen} }; \\
};
\draw (n66) -- +(0, -0.5) -| (n67);
\draw (n64) -- +(0, -0.5) -| (n65);
\draw (n62) -- +(0, -0.5) -| (n63);
\draw (n60) -- +(0, -0.5) -| (n61);
\draw (n56) -- +(0, -0.5) -| (n57);
\draw (n52) -- +(0, -0.5) -| (n53);
\draw (n49) -- +(0, -0.5) -| (n50);
\draw (n47) -- +(0, -0.5) -| (n48);
\draw (n45) -- +(0, -0.5) -| (n46);
\draw (n43) -- +(0, -0.5) -| (n44);
\draw (n40) -- +(0, -0.5) -| (n41);
\draw (n36) -- +(0, -0.5) -| (n37);
\draw (n33) -- +(0, -0.5) -| (n34);
\draw (n31) -- +(0, -0.5) -| (n32);
\draw (n28) -- +(0, -0.5) -| (n29);
\draw (n26) -- +(0, -0.5) -| (n27);
\draw (n22) -- +(0, -0.5) -| (n23);
\draw (n20) -- +(0, -0.5) -| (n21);
\draw (n18) -- +(0, -0.5) -| (n19);
\draw (n15) -- +(0, -0.5) -| (n16);
\draw (n12) -- +(0, -0.5) -| (n13);
\draw (n10) -- +(0, -0.5) -| (n11);
\draw (n6) -- +(0, -0.5) -| (n7);
\draw [white, -, line width=6pt] (n42)  +(0, -0.5) -| (n43);
\draw (n42) -- +(0, -0.5) -| (n43);
\draw [white, -, line width=6pt] (n42)  +(0, -0.5) -| (n45);
\draw (n42) -- +(0, -0.5) -| (n45);
\draw (n59) -- +(0, -0.5) -| (n60);
\draw (n59) -- +(0, -0.5) -| (n62);
\draw (n30) -- +(0, -0.5) -| (n31);
\draw (n30) -- +(0, -0.5) -| (n33);
\draw (n25) -- +(0, -0.5) -| (n26);
\draw (n25) -- +(0, -0.5) -| (n28);
\draw (n17) -- +(0, -0.5) -| (n18);
\draw (n17) -- +(0, -0.5) -| (n20);
\draw (n9) -- +(0, -0.5) -| (n10);
\draw (n9) -- +(0, -0.5) -| (n12);
\draw [white, -, line width=6pt] (n58)  +(0, -0.5) -| (n59);
\draw (n58) -- +(0, -0.5) -| (n59);
\draw [white, -, line width=6pt] (n58)  +(0, -0.5) -| (n64);
\draw (n58) -- +(0, -0.5) -| (n64);
\draw [white, -, line width=6pt] (n39)  +(0, -0.5) -| (n40);
\draw (n39) -- +(0, -0.5) -| (n40);
\draw [white, -, line width=6pt] (n39)  +(0, -0.5) -| (n42);
\draw (n39) -- +(0, -0.5) -| (n42);
\draw (n24) -- +(0, -0.5) -| (n25);
\draw (n24) -- +(0, -0.5) -| (n30);
\draw (n14) -- +(0, -0.5) -| (n15);
\draw (n14) -- +(0, -0.5) -| (n17);
\draw [white, -, line width=6pt] (n55)  +(0, -0.5) -| (n56);
\draw (n55) -- +(0, -0.5) -| (n56);
\draw [white, -, line width=6pt] (n55)  +(0, -0.5) -| (n58);
\draw (n55) -- +(0, -0.5) -| (n58);
\draw (n8) -- +(0, -0.5) -| (n9);
\draw (n8) -- +(0, -0.5) -| (n14);
\draw (n38) -- +(0, -0.5) -| (n39);
\draw (n38) -- +(0, -0.5) -| (n47);
\draw (n5) -- +(0, -0.5) -| (n6);
\draw (n5) -- +(0, -0.5) -| (n8);
\draw (n54) -- +(0, -0.5) -| (n55);
\draw (n54) -- +(0, -0.5) -| (n66);
\draw (n35) -- +(0, -0.5) -| (n36);
\draw (n35) -- +(0, -0.5) -| (n38);
\draw (n4) -- +(0, -0.5) -| (n5);
\draw (n4) -- +(0, -0.5) -| (n22);
\draw (n51) -- +(0, -0.5) -| (n52);
\draw (n51) -- +(0, -0.5) -| (n54);
\draw (n3) -- +(0, -0.5) -| (n4);
\draw (n3) -- +(0, -0.5) -| (n24);
\draw (n2) -- +(0, -0.5) -| (n3);
\draw (n2) -- +(0, -0.5) -| (n35);
\draw (n1) -- +(0, -0.5) -| (n2);
\draw (n1) -- +(0, -0.5) -| (n49);
\draw (n0) -- +(0, -0.5) -| (n1);
\draw (n0) -- +(0, -0.5) -| (n51);
\end{tikzpicture}

}
\caption{}
\end{subfigure}

\begin{subfigure}{.99\linewidth}
{
\begin{tikzpicture}[scale=0.4, align=center,
                                text width=0.65cm, inner sep=0mm, node distance=1mm]
\tiny
\matrix[row sep=0.2cm,column sep=0.05cm] {
& & & & & & & & & \node (n1) { \textcolor{blue}{SMAIN} }; & & & & & & & & & & & & \\
& & & & & & & & & & & & & & & \node (n19) { \textcolor{blue}{INF} }; & & & & & & \\
& & \node (n2) { \textcolor{blue}{NP} }; & & & & & & & & & & & \node (n20) { \textcolor{blue}{PPART} }; & & & & & & & & \\
& & & & \node (n9) { \textcolor{blue}{PP} }; & & & & & \node (n21) { \textcolor{blue}{PP} }; & & & & \node (n29) { \textcolor{blue}{PP} }; & & & & \node (n37) { \textcolor{blue}{PP} }; & & & & \\
& & & & & \node (n12) { \textcolor{blue}{NP} }; & & & & & \node (n24) { \textcolor{blue}{NP} }; & & & & \node (n32) { \textcolor{blue}{NP} }; & & & & \node (n40) { \textcolor{blue}{NP} }; & & & \\
\node (n3) { \textcolor{blue}{lid} }; & \node (n5) { \textcolor{blue}{adj} }; & \node (n7) { \textcolor{blue}{n} }; & \node (n10) { \textcolor{blue}{vz} }; & \node (n13) { \textcolor{blue}{lid} }; & & \node (n15) { \textcolor{blue}{n} }; & \node (n17) { \textcolor{blue}{ww} }; & \node (n22) { \textcolor{blue}{vz} }; & \node (n25) { \textcolor{blue}{adj} }; & & \node (n27) { \textcolor{blue}{n} }; & \node (n30) { \textcolor{blue}{vz} }; & \node (n33) { \textcolor{blue}{vnw} }; & & \node (n35) { \textcolor{blue}{n} }; & \node (n38) { \textcolor{blue}{vz} }; & \node (n41) { \textcolor{blue}{adj} }; & & \node (n43) { \textcolor{blue}{n} }; & \node (n47) { \textcolor{blue}{ww} }; & \node (n45) { \textcolor{blue}{ww} }; \\
\node (n4) { \textcolor{red}{Het} }; & \node (n6) { \textcolor{red}{uniform} }; & \node (n8) { \textcolor{red}{deel} }; & \node (n11) { \textcolor{red}{van} }; & \node (n14) { \textcolor{red}{het} }; & & \node (n16) { \textcolor{red}{Basistakenpakket} }; & \node[yshift=-2mm] (n18) { \textcolor{red}{moet} }; & \node (n23) { \textcolor{red}{in} }; & \node (n26) { \textcolor{red}{heel} }; & & \node (n28) { \textcolor{red}{Nederland} }; & \node (n31) { \textcolor{red}{aan} }; & \node (n34) { \textcolor{red}{alle} }; & & \node (n36) { \textcolor{red}{kinderen} }; & \node (n39) { \textcolor{red}{op} }; & \node (n42) { \textcolor{red}{dezelfde} }; & & \node (n44) { \textcolor{red}{wijze} }; & \node (n48) { \textcolor{red}{worden} }; & \node (n46) { \textcolor{red}{aangeboden} }; \\
};
\draw (n47) -- +(0, -0.5) -| (n48);
\draw (n45) -- +(0, -0.5) -| (n46);
\draw (n43) -- +(0, -0.5) -| (n44);
\draw (n41) -- +(0, -0.5) -| (n42);
\draw (n38) -- +(0, -0.5) -| (n39);
\draw (n35) -- +(0, -0.5) -| (n36);
\draw (n33) -- +(0, -0.5) -| (n34);
\draw (n30) -- +(0, -0.5) -| (n31);
\draw (n27) -- +(0, -0.5) -| (n28);
\draw (n25) -- +(0, -0.5) -| (n26);
\draw (n22) -- +(0, -0.5) -| (n23);
\draw (n17) -- +(0, -0.5) -| (n18);
\draw (n15) -- +(0, -0.5) -| (n16);
\draw (n13) -- +(0, -0.5) -| (n14);
\draw (n10) -- +(0, -0.5) -| (n11);
\draw (n7) -- +(0, -0.5) -| (n8);
\draw (n5) -- +(0, -0.5) -| (n6);
\draw (n3) -- +(0, -0.5) -| (n4);
\draw (n40) -- +(0, -0.5) -| (n41);
\draw (n40) -- +(0, -0.5) -| (n43);
\draw (n32) -- +(0, -0.5) -| (n33);
\draw (n32) -- +(0, -0.5) -| (n35);
\draw (n24) -- +(0, -0.5) -| (n25);
\draw (n24) -- +(0, -0.5) -| (n27);
\draw (n12) -- +(0, -0.5) -| (n13);
\draw (n12) -- +(0, -0.5) -| (n15);
\draw (n37) -- +(0, -0.5) -| (n38);
\draw (n37) -- +(0, -0.5) -| (n40);
\draw (n29) -- +(0, -0.5) -| (n30);
\draw (n29) -- +(0, -0.5) -| (n32);
\draw (n21) -- +(0, -0.5) -| (n22);
\draw (n21) -- +(0, -0.5) -| (n24);
\draw (n9) -- +(0, -0.5) -| (n10);
\draw (n9) -- +(0, -0.5) -| (n12);
\draw [white, -, line width=6pt] (n20)  +(0, -0.5) -| (n21);
\draw (n20) -- +(0, -0.5) -| (n21);
\draw [white, -, line width=6pt] (n20)  +(0, -0.5) -| (n29);
\draw (n20) -- +(0, -0.5) -| (n29);
\draw [white, -, line width=6pt] (n20)  +(0, -0.5) -| (n37);
\draw (n20) -- +(0, -0.5) -| (n37);
\draw [white, -, line width=6pt] (n20)  +(0, -0.5) -| (n45);
\draw (n20) -- +(0, -0.5) -| (n45);
\draw (n2) -- +(0, -0.5) -| (n3);
\draw (n2) -- +(0, -0.5) -| (n5);
\draw (n2) -- +(0, -0.5) -| (n7);
\draw (n2) -- +(0, -0.5) -| (n9);
\draw (n19) -- +(0, -0.5) -| (n20);
\draw (n19) -- +(0, -0.5) -| (n47);
\draw (n1) -- +(0, -0.5) -| (n2);
\draw (n1) -- +(0, -0.5) -| (n17);
\draw (n1) -- +(0, -0.5) -| (n19);
\end{tikzpicture}

\begin{tikzpicture}[scale=0.4, align=center,
                                text width=0.6cm, inner sep=0mm, node distance=1mm]
\tiny
\matrix[row sep=0.2cm,column sep=0.05cm] {
& & & & & & & & & & & & \node (n0) { \textcolor{blue}{NT} }; & & & & & & & & & & \\
& & & & & & & & & & \node (n1) { \textcolor{blue}{NT} }; & & & & & & & & & & & & \\
& & & & & & & & & & & & & & \node (n22) { \textcolor{blue}{NT} }; & & & & & & & & \\
& & & & \node (n2) { \textcolor{blue}{NT} }; & & & & & & & & & & & & & \node (n31) { \textcolor{blue}{NT} }; & & & & & \\
& & & \node (n3) { \textcolor{blue}{NT} }; & & & & & & & & & & & & & \node (n32) { \textcolor{blue}{NT} }; & & & & & & \\
& & \node (n4) { \textcolor{blue}{NT} }; & & & \node (n12) { \textcolor{blue}{NT} }; & & & & & \node (n23) { \textcolor{blue}{NT} }; & & & & \node (n33) { \textcolor{blue}{NT} }; & & & & \node (n41) { \textcolor{blue}{NT} }; & & & & \\
& \node (n5) { \textcolor{blue}{NT} }; & & & & & \node (n15) { \textcolor{blue}{NT} }; & & & & & \node (n26) { \textcolor{blue}{NT} }; & & & & \node (n36) { \textcolor{blue}{NT} }; & & & & \node (n44) { \textcolor{blue}{NT} }; & & & \\
\node (n6) { \textcolor{blue}{P} }; & & \node (n8) { \textcolor{blue}{P} }; & \node (n10) { \textcolor{blue}{P} }; & \node (n13) { \textcolor{blue}{P} }; & \node (n16) { \textcolor{blue}{P} }; & & \node (n18) { \textcolor{blue}{P} }; & \node (n20) { \textcolor{blue}{P} }; & \node (n24) { \textcolor{blue}{P} }; & \node (n27) { \textcolor{blue}{P} }; & & \node (n29) { \textcolor{blue}{P} }; & \node (n34) { \textcolor{blue}{P} }; & \node (n37) { \textcolor{blue}{P} }; & & \node (n39) { \textcolor{blue}{P} }; & \node (n42) { \textcolor{blue}{P} }; & \node (n45) { \textcolor{blue}{P} }; & & \node (n47) { \textcolor{blue}{P} }; & \node (n51) { \textcolor{blue}{P} }; & \node (n49) { \textcolor{blue}{P} }; \\
\node (n7) { \textcolor{red}{Het} }; & & \node (n9) { \textcolor{red}{uniform} }; & \node (n11) { \textcolor{red}{deel} }; & \node (n14) { \textcolor{red}{van} }; & \node (n17) { \textcolor{red}{het} }; & & \node[yshift=-2mm] (n19) { \textcolor{red}{Basistakenpakket} }; & \node (n21) { \textcolor{red}{moet} }; & \node (n25) { \textcolor{red}{in} }; & \node (n28) { \textcolor{red}{heel} }; & & \node (n30) { \textcolor{red}{Nederland} }; & \node (n35) { \textcolor{red}{aan} }; & \node (n38) { \textcolor{red}{alle} }; & & \node (n40) { \textcolor{red}{kinderen} }; & \node (n43) { \textcolor{red}{op} }; & \node (n46) { \textcolor{red}{dezelfde} }; & & \node (n48) { \textcolor{red}{wijze} }; & \node (n52) { \textcolor{red}{worden} }; & \node (n50) { \textcolor{red}{aangeboden} }; \\
};
\draw (n51) -- +(0, -0.5) -| (n52);
\draw (n49) -- +(0, -0.5) -| (n50);
\draw (n47) -- +(0, -0.5) -| (n48);
\draw (n45) -- +(0, -0.5) -| (n46);
\draw (n42) -- +(0, -0.5) -| (n43);
\draw (n39) -- +(0, -0.5) -| (n40);
\draw (n37) -- +(0, -0.5) -| (n38);
\draw (n34) -- +(0, -0.5) -| (n35);
\draw (n29) -- +(0, -0.5) -| (n30);
\draw (n27) -- +(0, -0.5) -| (n28);
\draw (n24) -- +(0, -0.5) -| (n25);
\draw (n20) -- +(0, -0.5) -| (n21);
\draw (n18) -- +(0, -0.5) -| (n19);
\draw (n16) -- +(0, -0.5) -| (n17);
\draw (n13) -- +(0, -0.5) -| (n14);
\draw (n10) -- +(0, -0.5) -| (n11);
\draw (n8) -- +(0, -0.5) -| (n9);
\draw (n6) -- +(0, -0.5) -| (n7);
\draw (n44) -- +(0, -0.5) -| (n45);
\draw (n44) -- +(0, -0.5) -| (n47);
\draw (n36) -- +(0, -0.5) -| (n37);
\draw (n36) -- +(0, -0.5) -| (n39);
\draw (n26) -- +(0, -0.5) -| (n27);
\draw (n26) -- +(0, -0.5) -| (n29);
\draw (n15) -- +(0, -0.5) -| (n16);
\draw (n15) -- +(0, -0.5) -| (n18);
\draw (n5) -- +(0, -0.5) -| (n6);
\draw (n5) -- +(0, -0.5) -| (n8);
\draw (n41) -- +(0, -0.5) -| (n42);
\draw (n41) -- +(0, -0.5) -| (n44);
\draw (n33) -- +(0, -0.5) -| (n34);
\draw (n33) -- +(0, -0.5) -| (n36);
\draw (n23) -- +(0, -0.5) -| (n24);
\draw (n23) -- +(0, -0.5) -| (n26);
\draw (n12) -- +(0, -0.5) -| (n13);
\draw (n12) -- +(0, -0.5) -| (n15);
\draw (n4) -- +(0, -0.5) -| (n5);
\draw (n4) -- +(0, -0.5) -| (n10);
\draw (n32) -- +(0, -0.5) -| (n33);
\draw (n32) -- +(0, -0.5) -| (n41);
\draw (n3) -- +(0, -0.5) -| (n4);
\draw (n3) -- +(0, -0.5) -| (n12);
\draw [white, -, line width=6pt] (n31)  +(0, -0.5) -| (n32);
\draw (n31) -- +(0, -0.5) -| (n32);
\draw [white, -, line width=6pt] (n31)  +(0, -0.5) -| (n49);
\draw (n31) -- +(0, -0.5) -| (n49);
\draw (n2) -- +(0, -0.5) -| (n3);
\draw (n2) -- +(0, -0.5) -| (n20);
\draw [white, -, line width=6pt] (n22)  +(0, -0.5) -| (n23);
\draw (n22) -- +(0, -0.5) -| (n23);
\draw [white, -, line width=6pt] (n22)  +(0, -0.5) -| (n31);
\draw (n22) -- +(0, -0.5) -| (n31);
\draw [white, -, line width=6pt] (n1)  +(0, -0.5) -| (n2);
\draw (n1) -- +(0, -0.5) -| (n2);
\draw [white, -, line width=6pt] (n1)  +(0, -0.5) -| (n22);
\draw (n1) -- +(0, -0.5) -| (n22);
\draw (n0) -- +(0, -0.5) -| (n1);
\draw (n0) -- +(0, -0.5) -| (n51);
\end{tikzpicture}

}
\caption{}
\end{subfigure}
\caption{Examples of gold (top) and predicted (bottom) trees in Dutch. NT and P denote  predicted nonterminals and preterminals.}
\label{fig:appd-example-dutch}
\end{figure*}

%% file: acl_latex.bbl
\begin{thebibliography}{79}
\expandafter\ifx\csname natexlab\endcsname\relax\def\natexlab#1{#1}\fi

\bibitem[{Baltin(1983)}]{baltin-1983:extraposition}
Mark~R. Baltin. 1983.
\newblock Extraposition: Bounding versus government-binding.
\newblock \emph{Linguistic Inquiry}, 14(1):155--162.

\bibitem[{Bengio et~al.(2009)Bengio, Louradour, Collobert, and
  Weston}]{Bengio2009CurriculumL}
Yoshua Bengio, J{\'e}r{\^o}me Louradour, Ronan Collobert, and Jason Weston.
  2009.
\newblock Curriculum learning.
\newblock In \emph{ICML '09}.

\bibitem[{Bisk and Hockenmaier(2012)}]{bisk2012ccg}
Yonatan Bisk and Julia Hockenmaier. 2012.
\newblock Simple robust grammar induction with combinatory categorial grammars.
\newblock In \emph{Proceedings of AAAI}.

\bibitem[{Bisk and Hockenmaier(2013)}]{bisk-hockenmaier-2013-hdp}
Yonatan Bisk and Julia Hockenmaier. 2013.
\newblock \href {https://doi.org/10.1162/tacl_a_00211} {An {HDP} model for
  inducing {C}ombinatory {C}ategorial {G}rammars}.
\newblock \emph{Transactions of the Association for Computational Linguistics},
  1:75--88.

\bibitem[{Bouma and van Noord(2017)}]{bouma-van-noord-2017-increasing}
Gosse Bouma and Gertjan van Noord. 2017.
\newblock \href {https://aclanthology.org/W17-0403} {Increasing return on
  annotation investment: The automatic construction of a {U}niversal
  {D}ependency treebank for {D}utch}.
\newblock In \emph{Proceedings of the {N}o{D}a{L}i{D}a 2017 Workshop on
  Universal Dependencies ({UDW} 2017)}, pages 19--26, Gothenburg, Sweden.
  Association for Computational Linguistics.

\bibitem[{Brants et~al.(2001)Brants, Dipper, Hansen, Lezius, and
  Smith}]{Brants2001TheTT}
Sabine Brants, Stefanie Dipper, Silvia Hansen, Wolfgang Lezius, and George
  Smith. 2001.
\newblock The tiger treebank.

\bibitem[{Buhai et~al.(2020)Buhai, Halpern, Kim, Risteski, and
  Sontag}]{DBLP:conf/icml/BuhaiHKRS20}
Rares{-}Darius Buhai, Yoni Halpern, Yoon Kim, Andrej Risteski, and David~A.
  Sontag. 2020.
\newblock \href {http://proceedings.mlr.press/v119/buhai20a.html} {Empirical
  study of the benefits of overparameterization in learning latent variable
  models}.
\newblock In \emph{Proceedings of the 37th International Conference on Machine
  Learning, {ICML} 2020, 13-18 July 2020, Virtual Event}, volume 119 of
  \emph{Proceedings of Machine Learning Research}, pages 1211--1219. {PMLR}.

\bibitem[{Carroll and Charniak(1992)}]{Carroll1992TwoEO}
Glenn Carroll and Eugene Charniak. 1992.
\newblock Two experiments on learning probabilistic dependency grammars from
  corpora.

\bibitem[{Chiang and Riley(2020)}]{DBLP:conf/nips/0001R20}
David Chiang and Darcey Riley. 2020.
\newblock \href
  {https://proceedings.neurips.cc/paper/2020/hash/49ca03822497d26a3943d5084ed59130-Abstract.html}
  {Factor graph grammars}.
\newblock In \emph{Advances in Neural Information Processing Systems 33: Annual
  Conference on Neural Information Processing Systems 2020, NeurIPS 2020,
  December 6-12, 2020, virtual}.

\bibitem[{Chiu and Rush(2020)}]{chiu-rush-2020-scaling}
Justin Chiu and Alexander Rush. 2020.
\newblock \href {https://doi.org/10.18653/v1/2020.emnlp-main.103} {Scaling
  hidden {M}arkov language models}.
\newblock In \emph{Proceedings of the 2020 Conference on Empirical Methods in
  Natural Language Processing (EMNLP)}, pages 1341--1349, Online. Association
  for Computational Linguistics.

\bibitem[{Chiu et~al.(2021)Chiu, Deng, and Rush}]{DBLP:conf/nips/ChiuDR21}
Justin~T. Chiu, Yuntian Deng, and Alexander~M. Rush. 2021.
\newblock \href
  {https://proceedings.neurips.cc/paper/2021/hash/16c0d78ef6a76b5c247113a4c9514059-Abstract.html}
  {Low-rank constraints for fast inference in structured models}.
\newblock In \emph{Advances in Neural Information Processing Systems 34: Annual
  Conference on Neural Information Processing Systems 2021, NeurIPS 2021,
  December 6-14, 2021, virtual}, pages 2887--2898.

\bibitem[{Choi et~al.(2020)Choi, Vergari, and den
  Broeck}]{Choi2020ProbabilisticCA}
YooJung Choi, Antonio Vergari, and Guy~Van den Broeck. 2020.
\newblock Probabilistic circuits: A unifying framework for tractable
  probabilistic models.

\bibitem[{Chomsky and Sch{\"u}tzenberger(1963)}]{chomsky63}
N.~Chomsky and {M. P.} Sch{\"u}tzenberger. 1963.
\newblock \href {https://doi.org/10.1016/S0049-237X(08)72023-8} {The algebraic
  theory of context-free languages}.
\newblock \emph{Studies in Logic and the Foundations of Mathematics},
  35(C):118--161.

\bibitem[{Chomsky(1959)}]{Chomsky59a}
Noam Chomsky. 1959.
\newblock \href
  {http://dblp.uni-trier.de/db/journals/iandc/iandc2.html#Chomsky59a} {On
  certain formal properties of grammars}.
\newblock \emph{Inf. Control.}, 2(2):137--167.

\bibitem[{Cohen et~al.(2013)Cohen, Satta, and
  Collins}]{cohen-etal-2013-approximate}
Shay~B. Cohen, Giorgio Satta, and Michael Collins. 2013.
\newblock \href {https://aclanthology.org/N13-1052} {Approximate {PCFG} parsing
  using tensor decomposition}.
\newblock In \emph{Proceedings of the 2013 Conference of the North {A}merican
  Chapter of the Association for Computational Linguistics: Human Language
  Technologies}, pages 487--496, Atlanta, Georgia. Association for
  Computational Linguistics.

\bibitem[{Corro(2020)}]{corro-2020-span}
Caio Corro. 2020.
\newblock \href {https://doi.org/10.18653/v1/2020.emnlp-main.219} {Span-based
  discontinuous constituency parsing: a family of exact chart-based algorithms
  with time complexities from {O}(n{\^{}}6) down to {O}(n{\^{}}3)}.
\newblock In \emph{Proceedings of the 2020 Conference on Empirical Methods in
  Natural Language Processing (EMNLP)}, pages 2753--2764, Online. Association
  for Computational Linguistics.

\bibitem[{Dubey and Keller(2003)}]{dubey-keller-2003-probabilistic}
Amit Dubey and Frank Keller. 2003.
\newblock \href {https://doi.org/10.3115/1075096.1075109} {Probabilistic
  parsing for {G}erman using sister-head dependencies}.
\newblock In \emph{Proceedings of the 41st Annual Meeting of the Association
  for Computational Linguistics}, pages 96--103, Sapporo, Japan. Association
  for Computational Linguistics.

\bibitem[{Eisner(2016)}]{eisner-2016-inside}
Jason Eisner. 2016.
\newblock \href {https://doi.org/10.18653/v1/W16-5901} {Inside-outside and
  forward-backward algorithms are just backprop (tutorial paper)}.
\newblock In \emph{Proceedings of the Workshop on Structured Prediction for
  {NLP}}, pages 1--17, Austin, TX. Association for Computational Linguistics.

\bibitem[{Evang and Kallmeyer(2011)}]{evang-kallmeyer-2011-plcfrs}
Kilian Evang and Laura Kallmeyer. 2011.
\newblock \href {https://aclanthology.org/W11-2913} {{PLCFRS} parsing of
  {E}nglish discontinuous constituents}.
\newblock In \emph{Proceedings of the 12th International Conference on Parsing
  Technologies}, pages 104--116, Dublin, Ireland. Association for Computational
  Linguistics.

\bibitem[{Fern{\'a}ndez-Gonz{\'a}lez and
  G{\'o}mez-Rodr{\'\i}guez(2020)}]{fernandez2020}
Daniel Fern{\'a}ndez-Gonz{\'a}lez and Carlos G{\'o}mez-Rodr{\'\i}guez. 2020.
\newblock Discontinuous constituent parsing with pointer networks.
\newblock In \emph{Proceedings of AAAI}.

\bibitem[{Fern{\'a}ndez-Gonz{\'a}lez and
  G{\'o}mez-Rodr{\'\i}guez(2021)}]{fernandez-gonzalez-gomez-rodriguez-2021-reducing}
Daniel Fern{\'a}ndez-Gonz{\'a}lez and Carlos G{\'o}mez-Rodr{\'\i}guez. 2021.
\newblock Reducing discontinuous to continuous parsing with pointer network
  reordering.
\newblock In \emph{Proceedings of the 2021 Conference on Empirical Methods in
  Natural Language Processing}, Online and Punta Cana, Dominican Republic.
  Association for Computational Linguistics.

\bibitem[{Fern{\'a}ndez-Gonz{\'a}lez and
  G{\'o}mez-Rodr{\'\i}guez(2023)}]{FERNANDEZGONZALEZ202343}
Daniel Fern{\'a}ndez-Gonz{\'a}lez and Carlos G{\'o}mez-Rodr{\'\i}guez. 2023.
\newblock Discontinuous grammar as a foreign language.
\newblock \emph{Neurocomputing}, 524:43--58.

\bibitem[{Friedman et~al.(2022)Friedman, Wettig, and
  Chen}]{friedman2022finding}
Dan Friedman, Alexander Wettig, and Danqi Chen. 2022.
\newblock {Finding Dataset Shortcuts with Grammar Induction}.
\newblock In \emph{Proceedings of EMNLP}.

\bibitem[{Gazdar(1988)}]{Gazdar1988ApplicabilityOI}
Gerald Gazdar. 1988.
\newblock {Applicability of Indexed Grammars to Natural Languages}.
\newblock In \emph{Natural Language Parsing and Linguistic Theories}, pages
  69--94.

\bibitem[{Gildea(2010)}]{gildea-2010-optimal}
Daniel Gildea. 2010.
\newblock \href {https://aclanthology.org/N10-1118} {Optimal parsing strategies
  for linear context-free rewriting systems}.
\newblock In \emph{Human Language Technologies: The 2010 Annual Conference of
  the North {A}merican Chapter of the Association for Computational
  Linguistics}, pages 769--776, Los Angeles, California. Association for
  Computational Linguistics.

\bibitem[{Goodman(1996)}]{goodman-1996-parsing}
Joshua Goodman. 1996.
\newblock \href {https://doi.org/10.3115/981863.981887} {Parsing algorithms and
  metrics}.
\newblock In \emph{34th Annual Meeting of the Association for Computational
  Linguistics}, pages 177--183, Santa Cruz, California, USA. Association for
  Computational Linguistics.

\bibitem[{Han et~al.(2017)Han, Jiang, and Tu}]{han-etal-2017-dependency}
Wenjuan Han, Yong Jiang, and Kewei Tu. 2017.
\newblock \href {https://doi.org/10.18653/v1/D17-1176} {Dependency grammar
  induction with neural lexicalization and big training data}.
\newblock In \emph{Proceedings of the 2017 Conference on Empirical Methods in
  Natural Language Processing}, pages 1683--1688, Copenhagen, Denmark.
  Association for Computational Linguistics.

\bibitem[{He et~al.(2018)He, Neubig, and
  Berg-Kirkpatrick}]{he-etal-2018-unsupervised}
Junxian He, Graham Neubig, and Taylor Berg-Kirkpatrick. 2018.
\newblock \href {https://doi.org/10.18653/v1/D18-1160} {Unsupervised learning
  of syntactic structure with invertible neural projections}.
\newblock In \emph{Proceedings of the 2018 Conference on Empirical Methods in
  Natural Language Processing}, pages 1292--1302, Brussels, Belgium.
  Association for Computational Linguistics.

\bibitem[{Hong et~al.(2021)Hong, Li, Zhu, and Huang}]{hong2021grammar}
Yining Hong, Qing Li, Song-Chun Zhu, and Siyuan Huang. 2021.
\newblock {VLGrammar: Grounded Grammar Induction of Vision and Language}.
\newblock \emph{arXiv:2103.12975}.

\bibitem[{Icard(2020)}]{Icard2020-ICACGM}
Thomas Icard. 2020.
\newblock Calibrating generative models: The probabilistic
  chomsky-sch\"{u}tzenberger hierarchy.
\newblock \emph{Journal of Mathematical Psychology}, 95.

\bibitem[{Jiang et~al.(2016)Jiang, Han, and Tu}]{jiang-etal-2016-unsupervised}
Yong Jiang, Wenjuan Han, and Kewei Tu. 2016.
\newblock \href {https://doi.org/10.18653/v1/D16-1073} {Unsupervised neural
  dependency parsing}.
\newblock In \emph{Proceedings of the 2016 Conference on Empirical Methods in
  Natural Language Processing}, pages 763--771, Austin, Texas. Association for
  Computational Linguistics.

\bibitem[{Jin et~al.(2019)Jin, Doshi-Velez, Miller, Schwartz, and
  Schuler}]{jin-etal-2019-unsupervised}
Lifeng Jin, Finale Doshi-Velez, Timothy Miller, Lane Schwartz, and William
  Schuler. 2019.
\newblock \href {https://doi.org/10.18653/v1/P19-1234} {Unsupervised learning
  of {PCFG}s with normalizing flow}.
\newblock In \emph{Proceedings of the 57th Annual Meeting of the Association
  for Computational Linguistics}, pages 2442--2452, Florence, Italy.
  Association for Computational Linguistics.

\bibitem[{Jin et~al.(2021)Jin, Oh, and Schuler}]{jin-etal-2021-character-based}
Lifeng Jin, Byung-Doh Oh, and William Schuler. 2021.
\newblock \href {https://doi.org/10.18653/v1/2021.findings-emnlp.371}
  {Character-based {PCFG} induction for modeling the syntactic acquisition of
  morphologically rich languages}.
\newblock In \emph{Findings of the Association for Computational Linguistics:
  EMNLP 2021}, pages 4367--4378, Punta Cana, Dominican Republic. Association
  for Computational Linguistics.

\bibitem[{Jin and Schuler(2020)}]{jin-schuler-2020-grounded}
Lifeng Jin and William Schuler. 2020.
\newblock \href {https://aclanthology.org/2020.aacl-main.42} {Grounded {PCFG}
  induction with images}.
\newblock In \emph{Proceedings of the 1st Conference of the Asia-Pacific
  Chapter of the Association for Computational Linguistics and the 10th
  International Joint Conference on Natural Language Processing}, pages
  396--408, Suzhou, China. Association for Computational Linguistics.

\bibitem[{Johnson et~al.(2007)Johnson, Griffiths, and
  Goldwater}]{johnson-etal-2007-bayesian}
Mark Johnson, Thomas Griffiths, and Sharon Goldwater. 2007.
\newblock \href {https://aclanthology.org/N07-1018} {{B}ayesian inference for
  {PCFG}s via {M}arkov chain {M}onte {C}arlo}.
\newblock In \emph{Human Language Technologies 2007: The Conference of the
  North {A}merican Chapter of the Association for Computational Linguistics;
  Proceedings of the Main Conference}, pages 139--146, Rochester, New York.
  Association for Computational Linguistics.

\bibitem[{Joshi(1975)}]{JOSHI1975136}
Aravind~K. Joshi. 1975.
\newblock Tree adjunct grammars.
\newblock \emph{Journal of Computer and System Sciences}, 10(1):136--163.

\bibitem[{Joshi(1985)}]{joshi1985much}
Aravind~K Joshi. 1985.
\newblock How much context sensitivity is necessary for characterizing
  structural descriptions: Tree adjoining grammars.
\newblock \emph{Natural language parsing: Psychological, computational and
  theoretical perspectives}, pages 206--250.

\bibitem[{Kallmeyer and Maier(2010)}]{kallmeyer-maier-2010-data}
Laura Kallmeyer and Wolfgang Maier. 2010.
\newblock \href {https://aclanthology.org/C10-1061} {Data-driven parsing with
  probabilistic linear context-free rewriting systems}.
\newblock In \emph{Proceedings of the 23rd International Conference on
  Computational Linguistics (Coling 2010)}, pages 537--545, Beijing, China.
  Coling 2010 Organizing Committee.

\bibitem[{Kim(2021)}]{DBLP:conf/nips/Kim21a}
Yoon Kim. 2021.
\newblock \href
  {https://proceedings.neurips.cc/paper/2021/hash/dd17e652cd2a08fdb8bf7f68e2ad3814-Abstract.html}
  {Sequence-to-sequence learning with latent neural grammars}.
\newblock In \emph{Advances in Neural Information Processing Systems 34: Annual
  Conference on Neural Information Processing Systems 2021, NeurIPS 2021,
  December 6-14, 2021, virtual}, pages 26302--26317.

\bibitem[{Kim et~al.(2019)Kim, Dyer, and Rush}]{kim-etal-2019-compound}
Yoon Kim, Chris Dyer, and Alexander Rush. 2019.
\newblock \href {https://doi.org/10.18653/v1/P19-1228} {Compound probabilistic
  context-free grammars for grammar induction}.
\newblock In \emph{Proceedings of the 57th Annual Meeting of the Association
  for Computational Linguistics}, pages 2369--2385, Florence, Italy.
  Association for Computational Linguistics.

\bibitem[{Kingma and Ba(2015)}]{Kingma2015AdamAM}
Diederik~P. Kingma and Jimmy Ba. 2015.
\newblock Adam: A method for stochastic optimization.
\newblock \emph{CoRR}, abs/1412.6980.

\bibitem[{Klein and Manning(2001)}]{DBLP:conf/iwpt/KleinM01}
Dan Klein and Christopher~D. Manning. 2001.
\newblock Parsing and hypergraphs.
\newblock In \emph{Proceedings of the Seventh International Workshop on Parsing
  Technologies (IWPT-2001), 17-19 October 2001, Beijing, China}. Tsinghua
  University Press.

\bibitem[{Lari and Young(1990)}]{lari1990estimation}
Karim Lari and Steve~J Young. 1990.
\newblock The estimation of stochastic context-free grammars using the
  inside-outside algorithm.
\newblock \emph{Computer speech \& language}, 4(1):35--56.

\bibitem[{Levy(2005)}]{levy2005prob}
Roger Levy. 2005.
\newblock \emph{{Probabilistic Models of Word Order and Syntactic
  Discontinuity}}.
\newblock Ph.D. thesis, Stanford University.

\bibitem[{Liang and Klein(2008)}]{liang-klein-2008-analyzing}
Percy Liang and Dan Klein. 2008.
\newblock \href {https://aclanthology.org/P08-1100} {Analyzing the errors of
  unsupervised learning}.
\newblock In \emph{Proceedings of ACL-08: HLT}, pages 879--887, Columbus, Ohio.
  Association for Computational Linguistics.

\bibitem[{Liu et~al.(2022)Liu, Zhang, and den
  Broeck}]{DBLP:journals/corr/abs-2210-04398}
Anji Liu, Honghua Zhang, and Guy~Van den Broeck. 2022.
\newblock \href {https://doi.org/10.48550/arXiv.2210.04398} {Scaling up
  probabilistic circuits by latent variable distillation}.
\newblock \emph{CoRR}, abs/2210.04398.

\bibitem[{Maier(2010)}]{maier-2010-direct}
Wolfgang Maier. 2010.
\newblock \href {https://aclanthology.org/W10-1407} {Direct parsing of
  discontinuous constituents in {G}erman}.
\newblock In \emph{Proceedings of the {NAACL} {HLT} 2010 First Workshop on
  Statistical Parsing of Morphologically-Rich Languages}, pages 58--66, Los
  Angeles, CA, USA. Association for Computational Linguistics.

\bibitem[{Maier(2015)}]{maier-2015-discontinuous}
Wolfgang Maier. 2015.
\newblock \href {https://doi.org/10.3115/v1/P15-1116} {Discontinuous
  incremental shift-reduce parsing}.
\newblock In \emph{Proceedings of the 53rd Annual Meeting of the Association
  for Computational Linguistics and the 7th International Joint Conference on
  Natural Language Processing (Volume 1: Long Papers)}, pages 1202--1212,
  Beijing, China. Association for Computational Linguistics.

\bibitem[{Maier et~al.(2012)Maier, Kaeshammer, and
  Kallmeyer}]{maier-etal-2012-plcfrs}
Wolfgang Maier, Miriam Kaeshammer, and Laura Kallmeyer. 2012.
\newblock \href {https://aclanthology.org/W12-4615} {{PLCFRS} parsing
  revisited: Restricting the fan-out to two}.
\newblock In \emph{Proceedings of the 11th International Workshop on Tree
  Adjoining Grammars and Related Formalisms ({TAG}+11)}, pages 126--134, Paris,
  France.

\bibitem[{Merialdo(1994)}]{merialdo:1994}
Bernard Merialdo. 1994.
\newblock Tagging {English} {T}ext with a {P}robabilistic {M}odel.
\newblock \emph{Computational Linguistics}, 20(2):155--171.

\bibitem[{Pate and Johnson(2016)}]{pate-johnson-2016-grammar}
John~K Pate and Mark Johnson. 2016.
\newblock \href {https://aclanthology.org/C16-1003} {Grammar induction from
  (lots of) words alone}.
\newblock In \emph{Proceedings of {COLING} 2016, the 26th International
  Conference on Computational Linguistics: Technical Papers}, pages 23--32,
  Osaka, Japan. The COLING 2016 Organizing Committee.

\bibitem[{Peharz et~al.(2020)Peharz, Lang, Vergari, Stelzner, Molina, Trapp,
  den Broeck, Kersting, and Ghahramani}]{DBLP:conf/icml/PeharzLVS00BKG20}
Robert Peharz, Steven Lang, Antonio Vergari, Karl Stelzner, Alejandro Molina,
  Martin Trapp, Guy~Van den Broeck, Kristian Kersting, and Zoubin Ghahramani.
  2020.
\newblock \href {http://proceedings.mlr.press/v119/peharz20a.html} {Einsum
  networks: Fast and scalable learning of tractable probabilistic circuits}.
\newblock In \emph{Proceedings of ICML}.

\bibitem[{Pereira and Warren(1983)}]{pereira-warren-1983-parsing}
Fernando C.~N. Pereira and David H.~D. Warren. 1983.
\newblock \href {https://doi.org/10.3115/981311.981338} {Parsing as deduction}.
\newblock In \emph{21st Annual Meeting of the Association for Computational
  Linguistics}, pages 137--144, Cambridge, Massachusetts, USA. Association for
  Computational Linguistics.

\bibitem[{Pollard(1985)}]{pollard85head}
Carl Pollard. 1985.
\newblock \emph{{Generalized phrase structure grammars, head grammars and
  natural language}}.
\newblock Ph.D. thesis, Stanford University.

\bibitem[{Rabanser et~al.(2017)Rabanser, Shchur, and
  G{\"{u}}nnemann}]{DBLP:journals/corr/abs-1711-10781}
Stephan Rabanser, Oleksandr Shchur, and Stephan G{\"{u}}nnemann. 2017.
\newblock \href {http://arxiv.org/abs/1711.10781} {Introduction to tensor
  decompositions and their applications in machine learning}.
\newblock \emph{CoRR}, abs/1711.10781.

\bibitem[{Seddah et~al.(2014)Seddah, K{\"u}bler, and
  Tsarfaty}]{seddah-etal-2014-introducing}
Djam{\'e} Seddah, Sandra K{\"u}bler, and Reut Tsarfaty. 2014.
\newblock \href {https://aclanthology.org/W14-6111} {Introducing the {SPMRL}
  2014 shared task on parsing morphologically-rich languages}.
\newblock In \emph{Proceedings of the First Joint Workshop on Statistical
  Parsing of Morphologically Rich Languages and Syntactic Analysis of
  Non-Canonical Languages}, pages 103--109, Dublin, Ireland. Dublin City
  University.

\bibitem[{Seki et~al.(1991)Seki, Matsumura, Fujii, and Kasami}]{Seki1991OnMC}
Hiroyuki Seki, Takashi Matsumura, Mamoru Fujii, and Tadao Kasami. 1991.
\newblock On multiple context-free grammars.
\newblock \emph{Theor. Comput. Sci.}, 88:191--229.

\bibitem[{Shen et~al.(2018)Shen, Lin, Huang, and Courville}]{shen2018nlm}
Yikang Shen, Zhouhan Lin, Chin-Wei Huang, and Aaron Courville. 2018.
\newblock {N}eural {L}anguage {M}odeling by {J}ointly {L}earning {S}yntax and
  {L}exicon.
\newblock In \emph{Proceedings of ICLR}.

\bibitem[{Shen et~al.(2019)Shen, Tan, Sordoni, and Courville}]{shen2019ordered}
Yikang Shen, Shawn Tan, Alessandro Sordoni, and Aaron Courville. 2019.
\newblock {O}rdered {N}eurons: {I}ntegrating {T}ree {S}tructures into
  {R}ecurrent {N}eural {N}etworks.
\newblock In \emph{Proceedings of ICLR}.

\bibitem[{Shi et~al.(2020)Shi, Livescu, and Gimpel}]{shi-etal-2020-role}
Haoyue Shi, Karen Livescu, and Kevin Gimpel. 2020.
\newblock \href {https://doi.org/10.18653/v1/2020.emnlp-main.614} {On the role
  of supervision in unsupervised constituency parsing}.
\newblock In \emph{Proceedings of the 2020 Conference on Empirical Methods in
  Natural Language Processing (EMNLP)}, pages 7611--7621, Online. Association
  for Computational Linguistics.

\bibitem[{Shieber(1985)}]{shieber:cfl}
Stuart Shieber. 1985.
\newblock Evidence against the context-freeness of natural language.
\newblock \emph{Linguistics and Philosophy}, 8:333--43.

\bibitem[{Skut et~al.(1997)Skut, Krenn, Brants, and
  Uszkoreit}]{skut-etal-1997-annotation}
Wojciech Skut, Brigitte Krenn, Thorsten Brants, and Hans Uszkoreit. 1997.
\newblock \href {https://doi.org/10.3115/974557.974571} {An annotation scheme
  for free word order languages}.
\newblock In \emph{Fifth Conference on Applied Natural Language Processing},
  pages 88--95, Washington, DC, USA. Association for Computational Linguistics.

\bibitem[{Smith and Eisner(2006)}]{smith-eisner-2006-minimum}
David~A. Smith and Jason Eisner. 2006.
\newblock \href {https://aclanthology.org/P06-2101} {Minimum risk annealing for
  training log-linear models}.
\newblock In \emph{Proceedings of the {COLING}/{ACL} 2006 Main Conference
  Poster Sessions}, pages 787--794, Sydney, Australia. Association for
  Computational Linguistics.

\bibitem[{Stanojevi{\'c} and Steedman(2020)}]{stanojevic-steedman-2020-span}
Milo{\v{s}} Stanojevi{\'c} and Mark Steedman. 2020.
\newblock \href {https://doi.org/10.18653/v1/2020.iwpt-1.12} {Span-based
  {LCFRS}-2 parsing}.
\newblock In \emph{Proceedings of the 16th International Conference on Parsing
  Technologies and the IWPT 2020 Shared Task on Parsing into Enhanced Universal
  Dependencies}, pages 111--121, Online. Association for Computational
  Linguistics.

\bibitem[{Steedman(1987)}]{steedman1987ccg}
Mark Steedman. 1987.
\newblock {Combinatory Grammars and Parasitic Gaps}.
\newblock \emph{Natural Language and Linguistic Theory}, 5:403--439.

\bibitem[{Su et~al.(2021)Su, Rijhwani, Zhu, He, Wang, Bisk, and
  Neubig}]{su-etal-2021-dependency}
Ruisi Su, Shruti Rijhwani, Hao Zhu, Junxian He, Xinyu Wang, Yonatan Bisk, and
  Graham Neubig. 2021.
\newblock \href {https://doi.org/10.18653/v1/2021.conll-1.2} {Dependency
  induction through the lens of visual perception}.
\newblock In \emph{Proceedings of the 25th Conference on Computational Natural
  Language Learning}, pages 17--26, Online. Association for Computational
  Linguistics.

\bibitem[{van Cranenburgh et~al.(2016)van Cranenburgh, Scha, and
  Bod}]{vanCranenburgh2016DataOrientedPW}
Andreas van Cranenburgh, Remko J.~H. Scha, and Rens Bod. 2016.
\newblock Data-oriented parsing with discontinuous constituents and function
  tags.
\newblock \emph{J. Lang. Model.}, 4:57--111.

\bibitem[{van~der Beek et~al.(2001)van~der Beek, Bouma, Malouf, and van
  Noord}]{Beek2001TheAD}
Leonoor van~der Beek, Gosse Bouma, Robert Malouf, and Gertjan van Noord. 2001.
\newblock The alpino dependency treebank.
\newblock In \emph{CLIN}.

\bibitem[{van Noord et~al.(2013)van Noord, Bouma, Eynde, de~Kok, van~der Linde,
  Schuurman, Sang, and Vandeghinste}]{DBLP:series/tanlp/NoordBEKLSSV13}
Gertjan van Noord, Gosse Bouma, Frank~Van Eynde, Dani{\"{e}}l de~Kok, Jelmer
  van~der Linde, Ineke Schuurman, Erik Tjong~Kim Sang, and Vincent
  Vandeghinste. 2013.
\newblock \href {https://doi.org/10.1007/978-3-642-30910-6\_9} {Large scale
  syntactic annotation of written dutch: Lassy}.
\newblock In Peter Spyns and Jan Odijk, editors, \emph{Essential Speech and
  Language Technology for Dutch, Results by the STEVIN-programme}, Theory and
  Applications of Natural Language Processing, pages 147--164. Springer.

\bibitem[{Vijay-Shanker et~al.(1987)Vijay-Shanker, Weir, and
  Joshi}]{vijay-shanker-etal-1987-characterizing}
K.~Vijay-Shanker, David~J. Weir, and Aravind~K. Joshi. 1987.
\newblock \href {https://doi.org/10.3115/981175.981190} {Characterizing
  structural descriptions produced by various grammatical formalisms}.
\newblock In \emph{25th Annual Meeting of the Association for Computational
  Linguistics}, pages 104--111, Stanford, California, USA. Association for
  Computational Linguistics.

\bibitem[{Vilares and
  G{\'o}mez-Rodr{\'\i}guez(2020)}]{vilares-gomez-rodriguez-2020-discontinuous}
David Vilares and Carlos G{\'o}mez-Rodr{\'\i}guez. 2020.
\newblock \href {https://doi.org/10.18653/v1/2020.emnlp-main.221}
  {Discontinuous constituent parsing as sequence labeling}.
\newblock In \emph{Proceedings of the 2020 Conference on Empirical Methods in
  Natural Language Processing (EMNLP)}, pages 2771--2785, Online. Association
  for Computational Linguistics.

\bibitem[{Wang et~al.(2022)Wang, Titov, Andreas, and Kim}]{wang2022hier}
Bailin Wang, Ivan Titov, Jacob Andreas, and Yoon Kim. 2022.
\newblock {Hierarchical Phrase-based Sequence-to-Sequence Learning}.
\newblock In \emph{Proceedings of EMNLP}.

\bibitem[{Yang et~al.(2020)Yang, Jiang, Han, and Tu}]{yang-etal-2020-second}
Songlin Yang, Yong Jiang, Wenjuan Han, and Kewei Tu. 2020.
\newblock \href {https://doi.org/10.18653/v1/2020.coling-main.347}
  {Second-order unsupervised neural dependency parsing}.
\newblock In \emph{Proceedings of the 28th International Conference on
  Computational Linguistics}, pages 3911--3924, Barcelona, Spain (Online).
  International Committee on Computational Linguistics.

\bibitem[{Yang et~al.(2022)Yang, Liu, and Tu}]{yang-etal-2022-dynamic}
Songlin Yang, Wei Liu, and Kewei Tu. 2022.
\newblock \href {https://doi.org/10.18653/v1/2022.naacl-main.353} {Dynamic
  programming in rank space: Scaling structured inference with low-rank {HMM}s
  and {PCFG}s}.
\newblock In \emph{Proceedings of the 2022 Conference of the North American
  Chapter of the Association for Computational Linguistics: Human Language
  Technologies}, pages 4797--4809, Seattle, United States. Association for
  Computational Linguistics.

\bibitem[{Yang et~al.(2021{\natexlab{a}})Yang, Zhao, and
  Tu}]{yang-etal-2021-neural}
Songlin Yang, Yanpeng Zhao, and Kewei Tu. 2021{\natexlab{a}}.
\newblock \href {https://doi.org/10.18653/v1/2021.acl-long.209} {Neural
  bi-lexicalized {PCFG} induction}.
\newblock In \emph{Proceedings of the 59th Annual Meeting of the Association
  for Computational Linguistics and the 11th International Joint Conference on
  Natural Language Processing (Volume 1: Long Papers)}, pages 2688--2699,
  Online. Association for Computational Linguistics.

\bibitem[{Yang et~al.(2021{\natexlab{b}})Yang, Zhao, and
  Tu}]{yang-etal-2021-pcfgs}
Songlin Yang, Yanpeng Zhao, and Kewei Tu. 2021{\natexlab{b}}.
\newblock \href {https://doi.org/10.18653/v1/2021.naacl-main.117} {{PCFG}s can
  do better: Inducing probabilistic context-free grammars with many symbols}.
\newblock In \emph{Proceedings of the 2021 Conference of the North American
  Chapter of the Association for Computational Linguistics: Human Language
  Technologies}, pages 1487--1498, Online. Association for Computational
  Linguistics.

\bibitem[{Zhang et~al.(2021)Zhang, Song, Jin, Xu, Yu, and
  Luo}]{zhang-etal-2021-video}
Songyang Zhang, Linfeng Song, Lifeng Jin, Kun Xu, Dong Yu, and Jiebo Luo. 2021.
\newblock \href {https://doi.org/10.18653/v1/2021.naacl-main.119} {Video-aided
  unsupervised grammar induction}.
\newblock In \emph{Proceedings of the 2021 Conference of the North American
  Chapter of the Association for Computational Linguistics: Human Language
  Technologies}, pages 1513--1524, Online. Association for Computational
  Linguistics.

\bibitem[{Zhao and Titov(2020)}]{zhao-titov-2020-visually}
Yanpeng Zhao and Ivan Titov. 2020.
\newblock \href {https://doi.org/10.18653/v1/2020.emnlp-main.354} {Visually
  grounded compound {PCFG}s}.
\newblock In \emph{Proceedings of the 2020 Conference on Empirical Methods in
  Natural Language Processing (EMNLP)}, pages 4369--4379, Online. Association
  for Computational Linguistics.

\bibitem[{Zhu et~al.(2020)Zhu, Bisk, and Neubig}]{zhu-etal-2020-return}
Hao Zhu, Yonatan Bisk, and Graham Neubig. 2020.
\newblock \href {https://doi.org/10.1162/tacl_a_00337} {The return of lexical
  dependencies: Neural lexicalized {PCFG}s}.
\newblock \emph{Transactions of the Association for Computational Linguistics},
  8:647--661.

\end{thebibliography}
